\documentclass{article}

\PassOptionsToPackage{table}{xcolor}
\usepackage{etoolbox}       %
\usepackage[utf8]{inputenc} 

\usepackage{graphicx}  %

\usepackage[parfill]{parskip}
\usepackage{amsmath,amsthm}
\usepackage{mathtools}  %
\usepackage{adjustbox}
\newtoggle{arxiv}
\toggletrue{arxiv}

\newcommand{\vv}{\mathbf{v}}

  \usepackage[parfill]{parskip}
  \usepackage[citestyle=authoryear-comp, uniquename=false, uniquelist=false,sorting=ynt, maxbibnames=99, maxcitenames=1, backend=biber]{biblatex}
  \newcommand{\citep}{\parencite}
  
  \newcommand{\citet}{\textcite}
  \newlength{\defbaselineskip}
  \RequirePackage[T1]{fontenc}
  \RequirePackage[tt=false, type1=true]{libertine}
  \RequirePackage[varqu]{zi4}
  \RequirePackage[libertine]{newtxmath}

\usepackage{bm}

\usepackage[inline]{enumitem}
\usepackage{booktabs}
\usepackage{subcaption}
\usepackage{multirow}
\usepackage{wrapfig}
\usepackage{algorithm,algorithmicx,algpseudocode}
\usepackage{nicematrix}

\usepackage[frozencache,cachedir=.]{minted} %

\usepackage{import}
\usepackage{lettrine}

\newtheorem{dfn}{Definition}

\usepackage{pifont}%
\newcommand{\xmark}{\ding{55}}%
\usepackage[T1]{fontenc}

\usepackage{thm-restate}

\usepackage{url}            
\usepackage{booktabs}       
\usepackage{amsfonts}       
\usepackage{nicefrac}       
\usepackage{microtype}      

\usepackage{wrapfig}

\usepackage{caption}
\usepackage{capt-of}
\usepackage{lipsum}

\usepackage[colorlinks]{hyperref}
\usepackage[capitalise,noabbrev]{cleveref}

\usepackage{threeparttable}

\definecolor{c7}{HTML}{6C7869}
\definecolor{c6}{HTML}{FBE7B4}
\definecolor{c5}{HTML}{9B5353}
\definecolor{c1}{HTML}{586770}
\definecolor{c4}{HTML}{2a4a67}
\definecolor{c3}{HTML}{6d2a58}
\definecolor{c2}{HTML}{34142a}
\hypersetup{colorlinks=true, citecolor=c3, linkcolor=c2, urlcolor=c2}

\definecolor{mydarkred}{HTML}{9B5353}
\definecolor{mydarkpurple}{HTML}{65657E}
\definecolor{mydarkgreen}{HTML}{6C7869}
\definecolor{darkred}{HTML}{a70c0c}
\definecolor{darkblue}{HTML}{4D5B9C}

\definecolor{myblue}{HTML}{FDF5E0} 
\definecolor{mygray}{HTML}{DBE2E9} 
\definecolor{mygreen}{HTML}{E6F3FC}

\definecolor{dark2orange}{rgb}{0.9, 0.4, 0.}
\definecolor{dark2purple}{rgb}{0.4, 0.4, 0.8}

\hypersetup{
    colorlinks=true,
    linkcolor=black,
    urlcolor=c1,
    citecolor=c1,
}

\usepackage{multirow}

\newcommand{\undermath}[2]{\underset{#1}{\underbrace{#2}}}

\newcommand{\R}[0]{\mathbb{R}}

\newcommand{\inner}[2]{\langle #1, #2 \rangle}

\newcommand{\M}[0]{\mathcal{M}}

\newcommand{\head}[1]{\vspace{1.7mm}\noindent{{\textcolor{c4}{\bf #1.}}}}
\newcommand{\headred}[1]{\vspace{1.7mm}\noindent{{\textcolor{c5}{\bf #1.}}}}
\newcommand{\headdot}[1]{\vspace{1.7mm}\noindent{{\textcolor{c4}{\bf #1}}}}

\newcommand{\model}[0]{\textsc{Hope}}

\newcommand{\mb}[1]{\boldsymbol{#1}}

\usepackage{mathtools}

\usepackage{tikz}

\usepackage{authblk}
\usepackage{epigraph}
\usepackage{soul}

\usepackage{dsfont}

\newcommand{\vk}{\boldsymbol{k}}
\newcommand{\vq}{\boldsymbol{q}}
\renewcommand{\vv}{\boldsymbol{v}}
\newcommand{\vu}{\boldsymbol{u}}
\newcommand{\vx}{\boldsymbol{x}}
\newcommand{\vy}{\boldsymbol{y}}

\newcommand{\vg}{\boldsymbol{g}}
\newcommand{\vm}{\boldsymbol{m}}
\newcommand{\vo}{\boldsymbol{O}}
\newcommand{\vp}{\boldsymbol{P}}

\usepackage{tcolorbox}
\tcbuselibrary{breakable}
\usepackage{xcolor}
\usepackage{lettrine,Zallman}
\newtcolorbox{c4box}{boxrule=0.75pt, left=3pt,right=3pt, colback=c4!3!white,colframe=c4!70!white}
\newtcolorbox{c3box}{boxrule=1pt, colback=c3!5!white,colframe=c3!50!white}
\newtcolorbox{c2box}{boxrule=1pt, colback=c2!5!white,colframe=c2!50!white}
\newtcolorbox{c1box}{boxrule=1pt, colback=c1!5!white,colframe=c1!50!white}

\usepackage[framemethod=tikz]{mdframed}
\mdfdefinestyle{mystyle}{%
  rightline=true,
  innerleftmargin=10,
  innerrightmargin=10,
  outerlinewidth=3pt,
  topline=false,
  rightline=true,
  bottomline=false,
  skipabove=\topsep,
  skipbelow=\topsep
}

\newtcolorbox{myboxi}[1][]{
  breakable,
  title=#1,
  colback=c4!5,
  colbacktitle=c4!5,
  coltitle=black,
  fonttitle=\bfseries,
  bottomrule=0pt,
  toprule=0pt,
  leftrule=2pt,
  rightrule=2pt,
  titlerule=0pt,
  arc=0pt,
  outer arc=0pt,
  colframe=c4,
}

\newtcolorbox{myboxismall}[1][]{
  breakable,
  title=#1,
  colback=c5!5,
  colbacktitle=c5!5,
  coltitle=black,
  fonttitle=\bfseries,
  bottomrule=0pt,
  toprule=0pt,
  leftrule=2pt,
  rightrule=2pt,
  titlerule=0pt,
  arc=0pt,
  outer arc=0pt,
  colframe=c5,
}

\newtcolorbox{myboxismallyellow}[1][]{
  breakable,
  title=#1,
  colback=c6!5,
  colbacktitle=c6!5,
  coltitle=black,
  fonttitle=\bfseries,
  bottomrule=0pt,
  toprule=0pt,
  leftrule=2pt,
  rightrule=2pt,
  titlerule=0pt,
  arc=0pt,
  outer arc=0pt,
  colframe=c6,
}

\newtcolorbox{myboxismallgreen}[1][]{
  breakable,
  title=#1,
  colback=c7!5,
  colbacktitle=c7!5,
  coltitle=black,
  fonttitle=\bfseries,
  bottomrule=0pt,
  toprule=0pt,
  leftrule=2pt,
  rightrule=2pt,
  titlerule=0pt,
  arc=0pt,
  outer arc=0pt,
  colframe=c7,
}

\newtcolorbox{myboxnote}[1][]{
  breakable,
  title=#1,
  colback=orange!0,
  colbacktitle=orange!0,
  coltitle=black,
  fonttitle=\bfseries,
  bottomrule=0pt,
  toprule=0pt,
  leftrule=2pt,
  rightrule=2pt,
  titlerule=0pt,
  arc=0pt,
  outer arc=0pt,
  colframe=orange,
}

\newcommand{\nsam}{\texttt{NSAM}}

\newcommand\blfootnote[1]{%
  \begingroup
    \addtocounter{footnote}{3}%
  \renewcommand\thefootnote{}\footnote{#1}%
  \addtocounter{footnote}{-4}%
  \endgroup
}

\usepackage{authblk}
\usepackage{epigraph}

\usepackage{lettrine,Zallman}

\title{\vspace{-5ex} Nested Learning: The Illusion of Deep Learning Architecture}

\author{Ali Behrouz \protect \blfootnote{Correspondence
to: \texttt{\{alibehrouz,~razaviyayn,~mirrokni\}@google.com} \: and \: \texttt{peilin.zhong@columbia.edu}.}}
\author{Meisam Razaviyayn}
\author{Peilin Zhong}
\author{Vahab Mirrokni \protect \blfootnote{A version of this work is published at Neural Information Processing Systems (NeurIPS) 2025.}}

\affil[]{\protect \includegraphics[width=40mm]{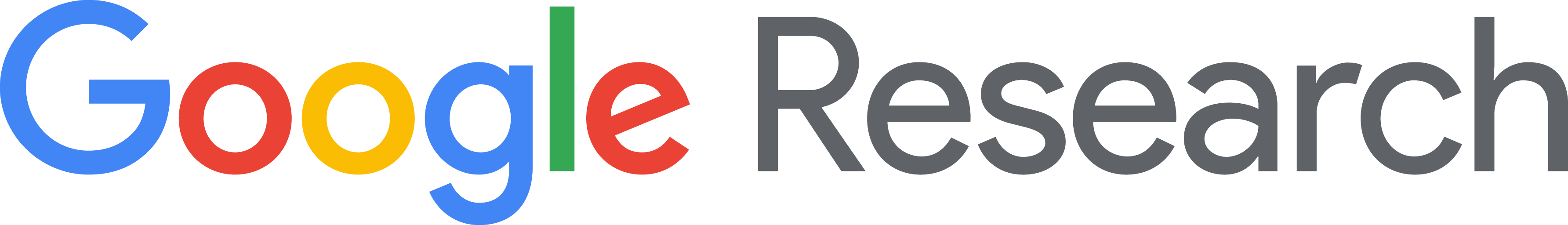}} 
\vspace{-13ex}
\date{}

\begin{document}

\maketitle

\vspace{-3ex}
\begin{abstract}
Over the last decades, developing more powerful neural architectures and simultaneously designing optimization algorithms to effectively train them have been the core of research efforts to enhance the capability of machine learning models. Despite the recent progresses, particularly in developing Language Models (LMs), there are fundamental challenges and unanswered questions about how such models can \emph{continually learn/memorize, self-improve, and find effective solutions}. In this paper, we present a new learning paradigm, called Nested Learning (NL), that coherently represents a machine learning model with a set of nested, multi-level, and/or parallel optimization problems, each of which with its own ``\emph{context flow}''. Through the lenses of NL,  existing deep learning methods learns from data through \emph{compressing} their own context flow, and  \emph{in-context learning} naturally emerges in large models. NL suggests a philosophy  to design more expressive learning algorithms with more ``\emph{levels}'', resulting in higher-order in-context learning and potentially unlocking effective continual learning capabilities. In addition to its neuro-scientific motivation,  we advocate for NL by presenting three core contributions: (1) Expressive Optimizers: We show that  known gradient-based optimizers, such as Adam, SGD with Momentum, etc., are in fact associative memory modules that aim to compress the gradients' information (by gradient descent). Building on this insight, we present other ``more expressive" optimizers with deep memory and/or more powerful learning rules; (2) Self-Modifying Learning Module: Taking advantage of NL's insights on learning algorithms, we present a  sequence model that learns how to modify itself by learning its own update algorithm; and (3) Continuum Memory System: We present a new formulation for memory system that generalizes the traditional viewpoint of ``long-term/short-term memory''. Combining our self-modifying sequence model with the continuum memory system, we present a continual learning module, called \model, showing promising results in language modeling, knowledge incorporation, and few-shot generalization tasks, continual learning, and~long-context~reasoning~tasks.   
\end{abstract}

\vspace{-1ex}
\epigraph{``We cannot solve our problems with the same thinking we used when we created them!"}{--- \textup{Attributed to Albert Einstein}}
\vspace{-2ex}

\vspace{-3ex}
\section{Introduction}\label{sec:intro}
For decades, AI research has focused on designing machine learning algorithms that learn from data~\citep{pitts1943linear, mcculloch1949brain, mcculloch1948statistical, samuel1959some} or experience~\citep{silver2025welcome, sutton1998reinforcement, Robotica_1999}; often by optimizing an objective $\mathcal{L}(\boldsymbol{\theta})$ over parameters  $\boldsymbol{\theta} \in \Theta$ with gradient-based methods. While traditional machine learning techniques required careful engineering and domain expertise to design feature extractors, limiting their ability to directly process and learn from natural data~\citep{lecun2015deep}, deep representation learning offered a fully automated alternative to discover the representations needed for the task. Thereafter, deep learning has been an inseparable part of the large-scale computational models with seminal success in chemistry and biology~\citep{jumper2021highly}, games~\citep{silver2016mastering, silver2018general}, computer vision~\citep{krizhevsky2012imagenet, dosovitskiy2021an}, and multimodal and natural language understanding~\citep{comanici2025gemini, liu2024deepseek, achiam2023gpt}.

\begin{figure*}
    \centering
    \includegraphics[width=\linewidth]{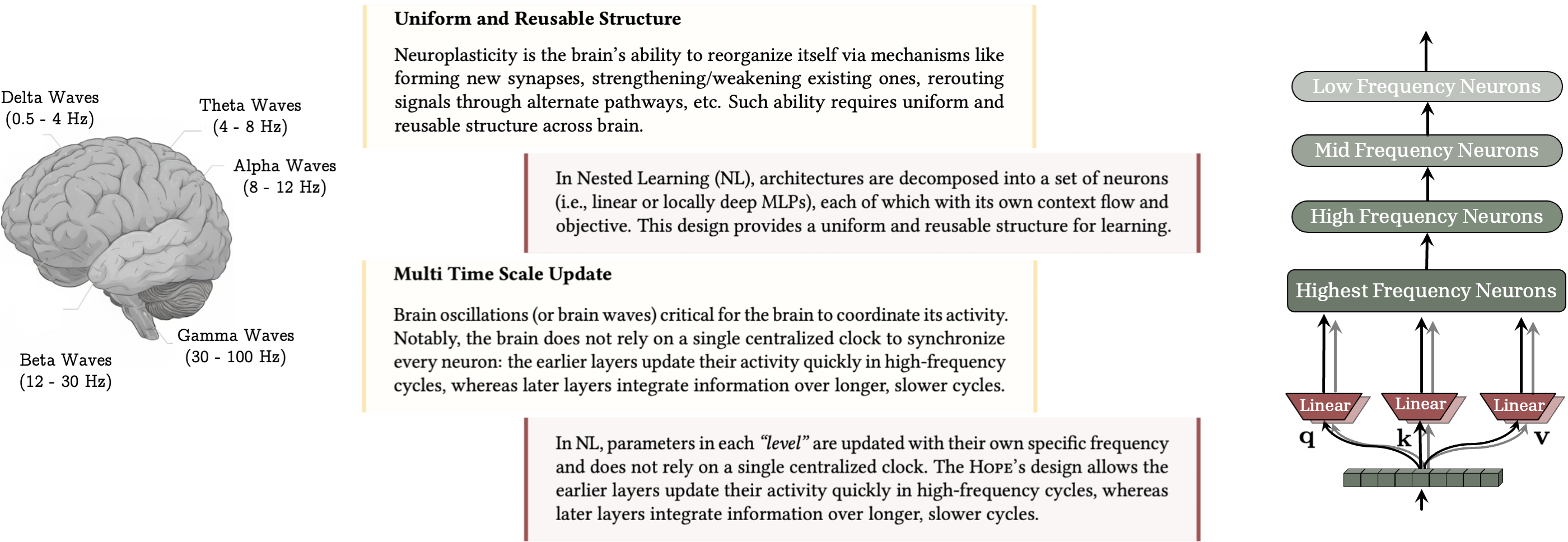}
    \caption{The uniform and reusable structure as well as multi time scale update in the brain are the key components to unlock the continual learning in humans. Nested Learning (NL) allows for multi time-scale update for each component of the brain, while showing that well-known architectures such as Transformers are in fact linear layers with different frequency updates.}
    \label{fig:NestedLearning_teaser}
\end{figure*}

Stacking of multiple layers, as it is done in deep learning models, provides the models with better expressive power in representing complex features, and more internal computations (e.g., \#FLOPS)~\citep{montufar2014number, poole2016exponential, hestness2017deep}, all of which are critical and desirable characteristics for static tasks that require in-distribution predictions over an a-priori fixed set. This deep design, however, is not a universal solution to all the challenges and cannot help the expressive power of the models in multiple aspects, for example: (i) The computational depth of deep models might not change with more layers~\citep{merrill2022saturated, sanford2024transformers}, leaving their ability to implement complex algorithms untouched compared to traditional shallow approaches~\citep{merrill2024the}; (ii) The capacity of some class of parameters might show marginal improvement with increasing the depth/width of the model~\citep{kaplan2020scaling}; (iii) The training process might converge to a suboptimal solution, mainly due to the suboptimal choice of the optimizer or its hyperparameters; and (iv) The model's ability to fast adapt to a new task, continually learn, and/or generalize to out-of-distribution data might not change with stacking more layers and requires more careful designs. 

The core part of the efforts to overcome the above challenges and to enhance the capability of deep learning models concentrate on: (1) developing more expressive class of parameters (i.e., neural architectures)~\citep{LSTM, fukushima1980neocognitron, transformers, krizhevsky2012imagenet, behrouz2024titans}; (2) introducing objectives that can better model the tasks~\citep{rumelhart1986learning, goodfellow2020generative, alshammari2025unifying, kingma2014autoencoding, hjelm2018learning};  (3) designing more efficient/effective optimization algorithms to find better solutions or with more resilience to forgetting~\citep{kingma2014adam, jordanmuon, gupta2018shampoo, farajtabar2020orthogonal}; and (4) scaling the model size to enhance its expressivity, when the ``right'' choice of architecture, objective, and optimization algorithms are made~\citep{hoffmann2022training, brown2020language, kaplan2020scaling}. Collectively, these advancements and new findings on scaling patterns of deep models have established the foundations upon which Large Language Models (LLMs) have been built.

The development of LLMs marks a pivotal milestone in deep learning research: a paradigm shift from task-specific models to more general-purpose systems with various emergent capabilities as a result of scaling the ``right'' architectures~\citep{brown2020language, schaeffer2023emergent}. Despite all their success and remarkable capabilities in diverse sets of tasks~\citep{comanici2025gemini, nijkamp2023codegen, wang2023visionllm}, LLMs are largely static after their initial deployment phase, meaning that they successfully perform tasks learned during pre- or post-training, but are unable to continually acquire new capabilities beyond their immediate context. The only adaptable component of LLMs is their \emph{in-context learning} ability–a (known to be emergent)~characteristic of LLMs that enables fast adaption to the context and so perform zero- or few-shot tasks~\citep{brown2020language}. Beyond in-context learning, recent efforts to overcome the static nature of LLMs either are computationally expensive, require external components, lack generalization, and/or might suffer from catastrophic forgetting~\citep{eyuboglu2025cartridges, yu2025finemedlmo, akyureksurprising}, which has led researchers to question if there is a need to revisit how to design machine learning models and if a new learning paradigm beyond stacking of layers is required to unleash the capabilities of LLMs in continual setups.

\head{Current Models only Experience the Immediate Present}
As an analogy and to better illustrate the static nature of LLMs, we use the example of anterograde amnesia–a neurological condition where a person cannot form new long-term memories after the onset of the disorder, while existing memories remain intact~\citep{scoville1957loss}. This condition limits the person's knowledge and experiences to a short window of present and long past–before the onset of the disorder–which results in continuously experiencing the immediate present as if it were always new. The memory processing system of current LLMs suffer from a similar pattern. Their knowledge is limited to either, the immediate context that fits into their context window, or the knowledge in MLPs that stores long-past, before the onset of \emph{``end of pre-training.''} This analogy, has motivated us to take inspiration from neurophysiology literature and how~brain~consolidate~its~short-term~memories.

\subsection{Human Brain Perspective and Neurophysiological Motivation}\label{sec:brain-perspective}
Human brain is highly efficient and effective when it comes to continual learning, which is often attributed to neuroplasticity—the brain's remarkable capacity to change itself in response to new experiences, memories, learning, and even damage~\citep{pascual2005plastic, johnston2009plasticity}. Recent studies support that the formation of Long-term memory involves at least two distinct but complementary consolidation processes~\citep{Stepwise-consolidation2021, frey1997synaptic, yang2024selection}: (1) A rapid ``online'' consolidation (also known as synaptic consolidation) phase occurs immediately or soon after learning, even during wakefulness. This is when new and initially fragile memory traces are stabilized and begin transferring from short-term to long-term storage; (2) An ``offline'' consolidation (also known as systems consolidation) process repeats the replay of the recently encoded patterns—during sharp‑wave ripples (SWRs) in the hippocampus, coordinated with cortical sleep spindles and slow oscillations—strengthens and reorganizes the memory and supports transfer to cortical sites~\citep{ji2007coordinated, peyrache2009replay, foster2006reverse}.

Coming back to the analogy of anterograde amnesia, evidence indicates that the condition can impact both stages, but especially the online consolidation phase, mainly due to the fact that hippocampus is the gateway for encoding new declarative memories, and so its damage means new information never will be stored in long-term memory. As mentioned above, the design of LLMs, and more specifically Transformer-based backbones, suffers from a similar condition after the pre-training phase. That is, the information provided in the context, never impacts the long-term memory parameters (e.g., feedforward layers), and so the model is not capable of acquiring new knowledge or skill, unless the information is still stored in the short-term memory (e.g., in-context or attention). To this end, although the second stage is equally, or even more, crucial for the consolidation of memories, and its absence can damage the process and might cause loss of memory~\citep{drummond2000altered, yoo2007deficit}, in this work, we focus on the first stage: memory consolidation as an online process. As discussed earlier, the memory processing, its online consolidation, and so continual learning ability of humans are known to be highly relied on the neuroplasticity as well as neural oscillations~\citep{bliss1993synaptic, buzsaki2004neuronal, klinzing2019mechanisms}.

\head{Multi Time scale Processing System} Brain oscillations (also known as brainwaves)–a rhythmic fluctuations in brain activity–is not mere byproducts of brain function but is increasingly understood to play a crucial role in various cognitive functions such as attention, memory, and decision-making, and to be a core mechanism for organizing neural computation, coordinating communication between brain regions, and gating the synaptic plasticity that underlies learning and memory ~\citep{fries2015rhythms, fell2011role, cavanagh2014frontal}. These brainwaves are the results of the brain coordinating its computations in different timescales and frequency updates, where each frequency determines how often groups of brain neurons become active and share updated information.
More specifically, such neural oscillations are typically categorized into distinct frequencies, each of which has been associated with different cognitive functions and, critically, different timescales of information processing: ranging from (1) fast Gamma waves (frequency of $30$-$150$ Hz) that are mainly associated with sensory information to (2) Beta waves (frequency of $13$ - $30$ Hz) that are mainly associated with active thinking~\citep{buzsaki2004neuronal,BuschmanMiller2007Science,Lundqvist2016Neuron}, and (3) slow Delta and Theta waves (frequency of $0.5$ - $8$ Hz), mainly responsible for memory consolidation and learning~\citep{DiekelmannBorn2010NRN,Marshall2006Nature,Ngo2013Neuron,Staresina2015NatNeuro,Heusser2016NatNeuro,Daume2024Nature}. 

In deep learning models, however, the weights of the architectures are fixed at test time and also it is common in pre-training to use the same update rate for all the blocks/layers in the model. Later, in \autoref{sec:revisit}, however, we show that in-context learning provides an extreme case of this design and in fact, Transformer architectures are based on two extreme frequencies of update: i.e., $\infty$ and $0$ for attention and MLP blocks, respectively.

\head{Brain's Uniform and Reusable Structure} As discussed earlier, neuroplasticity is the brain's remarkable capability to change itself in response to new memories, knowledge, and even damage~\citep{pascual2005plastic, johnston2009plasticity}. This characteristic suggests a uniform architecture where neural elements are not rigidly dedicated to one function but are instead reusable, capable of being flexibly redeployed to support different cognitive needs. One real-world example of neural reusability is hemispherectomy–the surgical removal or disabling of one cerebral hemisphere, usually to alleviate severe epilepsy. Amazingly, if this surgery is done in childhood, patients can lead largely normal lives into adulthood with high functioning cognition and intact neural network organization that contains all the same core brain networks present in a typical two-hemisphere brain (networks for language, vision, etc.). This extraordinary outcome provides real-life proof of the brain’s uniform architecture. That is, even half a brain can reallocate resources and reorganize so that the person can function extremely well. Such cases, along with documented instances of individuals living relatively normally with missing pieces of cortex, highlight the brain's uniform and reusable structure.

Furthermore, this interpretation of brain's uniform and reusable structure demonstrate that memory in human brain is not an isolated system in some specific areas, and mainly is distributed across brain. That is, contrary to the traditional models of memory that often implied that different types of memory reside in distinct brain structures (e.g. short-term memory in frontal cortex vs. long-term memory in the hippocampus and cortex), modern research advocate for distributed neural circuits memory processing across multiple regions~\citep{christophel2017distributed, roy2022brain, kitamura2017engrams}.

The modern deep learning architectures in recent years, however, at least on the surface, seem to be heterogeneous and are based on a combination of a subset of self-attention variants~\citep{transformers}, modern recurrent neural networks~\citep{katharopoulos2020transformers, schlag2021linear, peng2025rwkv7, behrouz2024titans}, canon layers~\citep{Allenzhu2025-canon}, global convolutions~\citep{poli2023hyena, hasani2023liquid}, and MLP blocks~\citep{shazeer2020glu}. This raises the question of whether we need a new uniform architecture, or if our beliefs about the heterogeneity of current models need to be revisited.

\begin{figure*}
    \centering
    \includegraphics[width=\linewidth]{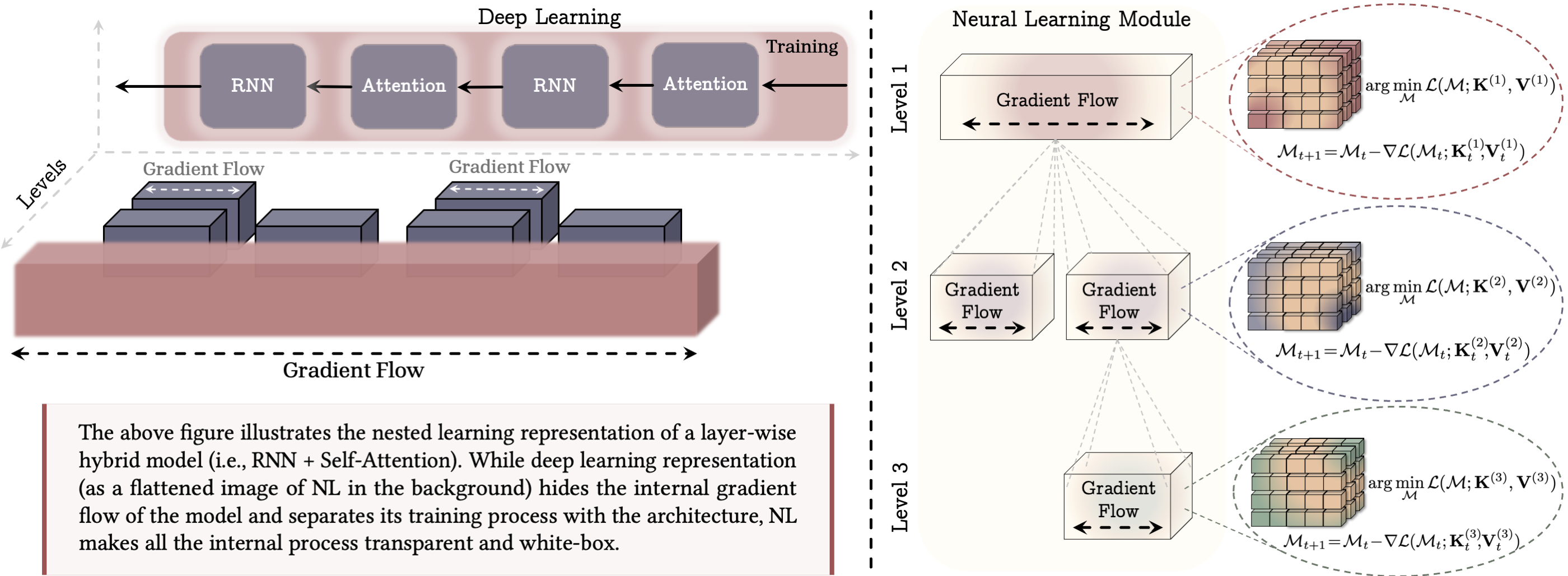}
    \vspace{1ex}
    \caption{Nested Learning Paradigm that represent a machine learning model and its training procedure as a set of nested optimization problems. (\textbf{Left}) An example of Hybrid architecture. While deep learning perspective, as the flattened image of NL, does not provide insight about the depth of computation in the blocks, NL transparently represent all the inner gradient flows. (\textbf{Right}) A Neural Learning Module: A computational model that learns how to compress its own context flow. For example, the first level corresponds to the model's most outer-loop training, often~refer~to~as~``\emph{pre-training}''~step.}
    \label{fig:NestedLearning}
\end{figure*}

\subsection{Contributions and Roadmap}
In this paper, we aim to present a unifying learning paradigm that not only provides new insights about existing algorithms, methods, and architectures, but it also reveals a new dimension to stacking layers in deep learning with enhancement of the computational depth, and continual learning ability of models. After discussing preliminary concepts and backgrounds in \textcolor{black}{\S}\ref{sec:prelim},~we~present:

\head{Nested Learning Paradigm (\textcolor{black}{\S}\ref{sec:nested-learning})} To answer the  questions raised above and to provide new insights on overcoming the design challenges in continual learning, architecture design, and computational depth of modern deep learning models, we present Nested Learning (NL)–a learning paradigm that allows each component of the machine learning model to have its own internal gradient flow on its own context in multiple levels, representing a model and its learning process (i.e., optimization) as an inter-connected system of nested, multi-level, and/or parallel optimization problems. We argue that the optimization process and the learning algorithms/architectures are fundamentally the same concepts but are in different levels of a system with different context (i.e., gradient vs. tokens). Furthermore, they are two inter-connected components where the learning algorithm/architecture generates the context for optimizers (i.e., the gradients), advocating for designing architecture-specific optimizers. We discuss different ways of knowledge transfer between levels, resulting in unifying and generalizing concepts like meta-learning, in-context learning, recurrent neural networks,~hypernetworks,~etc.

\head{Optimizers and Architectures as Learning Module (\textcolor{black}{\S\ref{sec:DeepOptimizers}, \S\ref{sec:arch})}}
Building on the NL's viewpoint, we argue that training \emph{a deep neural network} with backpropagation process and gradient descent is a compression and an optimization problem that aims to train an associative memory to map layers' inputs to their corresponding local error in the prediction. Accordingly, we argue that pre-training is a form of in-context learning, where the context is the entire pre-training data and the layers are compressing the context into their parameters. We demonstrate that such arguments are also valid for other popular gradient-based optimizers and they are associative memories that aim to compress the gradients into their parameters. From NL's terminology, gradient-based optimizers such as gradient descent with momentum, Adam~\citep{kingma2014adam}, and AdaGrad~\citep{duchi2011adaptive} can be decomposed into a two-level nested optimization problems, each of which is optimized with a simple gradient descent. In particular, this viewpoint makes it apparent that for compressing the gradients, in theory, Adam is the optimal associative memory with respect to the element-wise $L_2$~regression~objective.

We revisit previous findings on representing architectures as associative memories~\citep{behrouz2025Miras} and decompose their optimization process into a set of nested optimization problems, all of which optimized with gradient descent. Building on the above findings–i.e., popular gradient-based optimizers and modern architectures are both a set of nested and/or parallel optimization problems–we argue that the combination of these two–i.e., training an architecture with a specific optimizer–can also be represented as a set of nested and/or parallel optimization problem. Therefore, a neural learning module (a joint system of architecture and its training/optimization process) is a uniform model, in which all elements are linear or deep MLPs, while they are optimizing their own internal objective in different~levels~with~different~frequencies. 

Building upon the associative memory perspective of optimizers, we design a set of new learning updates (optimization steps) with more expressive memory structure or memory management in compressing the gradients. In particular, we argue that the choice of optimizer depends on the context of the optimization. A powerful optimizer for compressing the gradients might not be the best choice for compressing the tokens. To this end, we present a new variant of gradient descent, called Delta Gradient Descent (DGD), that its update not only depends on the current input, but also the state of the weight of the neural network, resulting in capturing the dependencies of data samples without i.i.d. assumption. 

\noindent
\textcolor{c5}{\textbf{Main Takeaways and Revisiting Common Terms: Continual and In-context Learning, Pre-Training, and Learning (\textcolor{black}{\S}\ref{sec:revisit})}.}
We discuss the main takeaways of NL about principal concepts and revisit some common terms: (1) We argue that continual learning can be viewed as a learning problem on a sequences of incoming contexts or episodes where different levels are responsible for compressing their own in-context knowledge and transfer it to higher levels. Based on this, we advocate for designing models and pipelines that do not rely on test/training phase, and rather continually manage their knowledge and memory; (2) In-context learning is the characteristic of ``having multiple nested levels''. Accordingly, Transformers in-context learning comes from  being a non-parametric solution to a certain regression objective on tokens, while modern recurrent models uses parametric learning processes in their lower levels; (3) We further revisit other terminologies such as learning/memorization, hybrid architectures, looped architectures, and learned optimizers.

\head{Continuum Memory System, Self-Referential Titans, and Hope (\textcolor{black}{\S}\ref{sec:cmlp}, \textcolor{black}{\S}\ref{sec:hope})}
We generalize the traditional viewpoint of ``long-term/short-term memory'' (LSM) by presenting the Continuum Memory Systems (CMSs) and see memory as a distributed inter-connected system with a spectrum of frequency updates. In this design, higher-frequency neurons are responsible for fast adaption but store memories/knowledge for a short period of time, while lower frequency neurons are responsible for more persistent knowledge. Comparing to LSM, we show that this multi-frequency design results in a loop process for memory of the model, meaning that knowledge can partially be recovered when it is forgotten. While we mainly design this memory system as a replacement of MLP blocks in Transformers, we take advantage of this intuition to design Multi-scale Momentum Muon (M3) optimizer–an optimization algorithm with multiple momentum terms–further supporting the importance of CMSs design in different contexts.

\head{Evaluations (\textcolor{black}{\S}\ref{sec:experiments})}
To support the effectiveness of our proofs-of-concept as well as the importance of nested learning design, we perform experimental evaluation on (1) Continual learning and in-context learning tasks including (i) learning new language, (ii) class incremental learning, and (iii) question/answering on a new corpus; (2) Long context understanding tasks, including needle-in-a-haystack~\citep{hsieh2024ruler} and BABILong~\citep{kuratov2024babilong} benchmarks; (3) Language modeling and common-sense reasoning tasks; (4) In-context recall  and memorization tasks; (5) Language recognition tasks; and (6) comparing different optimizers, including our M3 optimizer. Our results indicate the effectiveness of NL viewpoint in designing models with continual learning ability, having multiple levels of computations, and self-referential~process.

\section{Preliminaries}\label{sec:prelim}
In this section we discuss the notations and  review the background concepts. 

\head{Notations}
We let $x \in \R^{N \times d_{\text{in}}}$ be the input, $\M_t$ represent the state of memory/model $\M$ at time $t$, $\mathbf{K}$ be the keys, $\mathbf{V}$ be the values, and $\mathbf{Q}$ be the query matrices. We use bold lowercase letters with subscript $t$ to refer to the vector corresponds to the input $t$ (i.e., $\vk_t, \vv_t$, and $\vq_t$). We further refer to the distribution of any random variable $\mathcal{T}$ as $p(\mathcal{T})$. Throughout the paper, we use simple MLPs with $\mathcal{L}_{\M} \geq 1$ layers and residual connection as the architecture of the memory module $\M(\cdot)$. When it is needed, we parameterized the memory module with $\boldsymbol{\theta}_{\M} \supseteq \{W_1, W_2, \dots, W_{\mathcal{L}_{\M}} \}$, which at least includes the parameters of linear layers in the MLP. We use superscript with parenthesis to refer to parameters in different \emph{levels} of nested learning (different update frequency): i.e., either by using level index $W^{(\ell)}$ or its corresponding frequency $W^{(f_\ell)}$.

\head{Gradient Descent} Gradient descent is one of the most widely used optimization algorithms for high-dimensional problems, and its variants are standard tools for training large models. Given an objective $\mathcal{L}(\cdot; \cdot)$, the stochastic gradient descent (SGD) with step size $\eta_t > 0$ (a.k.a. learning rate) updates parameters by:
\begin{align}
    W_{t+1} \;=\; W_t - \eta_t \nabla_{W_t} \mathcal{L}(W_t; \boldsymbol{x_t}),
\end{align}
for data sample $\boldsymbol{x_t}$ from training set. The gradient descent formulation admits several equivalent characterizations that are useful in analysis. One of those equivalent formulations  is steepest‑descent in the Euclidean metric, where one step of gradient descent is equivalent to:
\begin{align}
    W_{t+1} \;=\; \arg\min_{W}\;\Big\{\,
\langle \nabla_{W} \mathcal{L}(W_t; \boldsymbol{x_t}),\, W \rangle \;+\; \tfrac{1}{2\eta_t}\|W-W_t\|_2^2
\,\Big\},
\end{align}
which is  minimizing a first‑order Taylor approximation regularized by a quadratic proximal term. The  GD step above is precisely a proximal update on the linearization of $\mathcal{L}(\cdot; \cdot)$ at $W_t$, revealing an implicit bias toward small moves in $L_2$-distance. Accumulating steps (with a constant learning rate $\eta$) yields the \emph{follow‑the‑regularized‑leader} (\textsc{FTRL}) form
\begin{align}
    W_{t+1} \;=\;
\arg\min_{W} \Big\{\;\bigg\langle \sum_{s=1}^{t} \nabla \mathcal{L}(W_s; \boldsymbol{x}_s),\, W \bigg\rangle
\;+\; \frac{1}{2\eta}\,\|W-W_1\|_2^2\Big\},
\end{align}
whose solution is $W_{t+1}=W_1-\eta\sum_{s=1}^{t}\nabla \mathcal{L}(W_s; \boldsymbol{x}_s)$. These two formulations are used in this paper interchangeably. However,   our discussions and formulations are generally valid for many other optimization algorithms as well.

\head{Meta Learning}
Designing an effective machine learning model often requires making decisions about its architecture parameterized by $\boldsymbol{\theta} \in \Theta$, objective $\mathcal{L}(\theta)$, and an optimizer, aiming to iteratively optimize the objective. Meta learning paradigm (or learning to learn)~\citep{schmidhuber1996simple, finn2017model, akyurek2022learning, irie2025metalearning} aim to automate a part of such decisions by modeling it as a \emph{two-level} optimization procedure, in which the outer model aims to learn to set parameters for the inner procedure to maximize the performance across a set of tasks. That is, given an objective parameterized by a parameter $\Phi$: i.e., $\ell(\boldsymbol{\theta}, \mathcal{D}; \Phi)$, one can formalize the outer loop process as optimizing parameter $\Phi$ over a set of tasks:
\begin{align}
    \Phi^{\ast} =
\underset{\Phi}{\arg\min}
\;\;
\mathbb{E}_{\mathcal{T}_i \sim p(\mathcal{T})}
\Biggl[
    \ell(\theta, \mathcal{T}_i; \Phi)
\Biggr],
\end{align}
where $p(\mathcal{T})$ is the distribution of tasks. While initial studies on meta-learning used supervised settings for the outer loop~\citep{schmidhuber1996simple}, recently, a more flexible family of methods that use an unsupervised process for the outer loop has gained popularity~\citep{brown2020language, chen2022meta, qu2025optimizing, finn2017model, akyurek2022learning}. In addition to the growing interest to use meta learning methods for a diverse set of downstream tasks~\citep{finn2017model, chen2022meta, qu2025optimizing, munkhdalai2019metalearned, irie2025metalearning}, in recent years, it also has shown popularity as a paradigm to design powerful sequence models~\citep{sun2024learning, behrouz2025Miras, behrouz2024titans}.

\head{Fast Weight Programmers (FWPs)}
Fast weight programmers (more recently refer to as linear Transformers)~\citep{schmidhuber1992learning, hinton1987using, ba2016using, schlag2021linear}, which also have been motivated from neuroscience perspective~\citep{gershman2025key}, are recurrent neural networks whose memory (or hidden state) is \emph{matrix‑valued}: a time‑varying fast‑weight matrix $\M_t\in\mathbb{R}^{d_{\mathrm{out}}\times d_{\mathrm{key}}}$ that serves as a short‑term memory. A separate “programmer” (slow net) maps each input $\boldsymbol{x}_t\in\mathbb{R}^{d_{\mathrm{in}}}$ to query, key, and value vectors and updates the fast weights online. A basic (Hebbian/outer‑product) FWP—often called \emph{vanilla} FWP—update its parameters with:
\begin{align}
    &\M_t \;=\; \alpha_t \M_{t-1} \;+\; \vv_t \phi(\vk_t)^{\top},
\end{align}
and retrieve from memory with $y_t \;=\; \M_t\,\phi(q_t)$, where $\phi(\cdot)$ is an element-wise feature map (often applied to both keys and queries). Unlike traditional RNNs or early variants of modern RNNs~\citep{LSTM, botev2024recurrentgemma, sun2023retentive} with vector states, the \emph{matrix} $\M_t$ is the recurrent state; it is \emph{written} by rank‑one update and \emph{read} by a matrix–vector multiplication, providing a compact, learnable key–value memory with constant state size across time.

\head{In-context Learning}
The concept of ``in-context learning'' initially defined by \citet{brown2020language} as the ability of a language model to leverage knowledge acquired during pre-training in order to infer and perform a new task solely based on its context (e.g., few examples, or natural language instructions). This broad and general definition, which simply is applicable for any language model with any architectural backbones and/or objective, later was formalized in a way that only described in-context learning for Transformer architectures trained with next token prediction objectives. Accordingly, despite extensive research on the algorithms/problems that a transformer-based model can learn in-context~\citep{akyurek2022learning, akyurek2024context, dherin2025learning, zhang2024trained}, the in-context learning as its general form is relatively underexplored. Throughout this paper, we use the most general definition of ``in-context learning'' and refer to it as the ability of a model to adapt itself to and learn from a given context. Our NL formulation connects ICL with the concept of associative memory, offering a unified explanation for the ICL capabilities of models regardless of their architectural backbone and/or objectives.

\section{Nested Learning}\label{sec:nested-learning}
This section discusses the motivations, formal definitions, and general high-level implications of Nested Learning (NL). We start with a formulation of associative memory and then by using step-by-step examples, we build the intuition behind architecture decomposition and its connection to modeling a neural network as an integrated system of optimization problems. We aim to first show how existing methods and concepts in deep learning fall under the NL paradigm and then we present new formulations that go beyond traditional methods and/or provide insights on how to improve existing algorithms and designs.

\subsection{Associative Memory}\label{sec:associaitve-memory}
Associative memory—the ability to form and retrieve connections between events—is a fundamental mental process and is an inseparable component of human learning~\citep{terry2017learning}. Often in the literature, the concept of memorization and learning are used interchangeably; in neuropsychology literature, however, these two are clearly distinguished. More specifically, following neuropsychology literature~\citep{okano2000learning}, we build our terminology based on the following definition of memory and learning:

\vspace{8pt}
\begin{myboxi}[Learning vs. Memorization:]
\textit{Memory is a neural update caused by an input, and {learning} is the process for acquiring effective and useful memory.}
\end{myboxi}
\vspace{8pt}

In this work, our goal is to first show that all the elements of a computational sequence model, including optimizers and neural networks, are \emph{associative memory systems} that compress their own \emph{context flow}. Broadly speaking, associative memory is an operator that maps a set of keys to a set of values. We follow the general definition of associative memory by \citet{behrouz2025Miras}: 

\begin{dfn}[Associative Memory] \label{def:associative-memory}
    Given a set of keys $\mathcal{K} \subseteq \mathbb{R}^{d_k}$ and values $\mathcal{V} \subseteq \mathbb{R}^{d_v}$, associative memory is an operator $\M(\cdot)$  
    that maps the set of keys $\mathcal{K}$ to values $\mathcal{V}$. To learn such mapping from the data, an objective $\tilde{\mathcal{L}}(\cdot;\cdot)$ measures the quality of the mapping and $\M$ 
    can be computed by:
\begin{align}\label{eq:attentional-bias-loss}
    \M^* = \arg\min_{\M}\quad \tilde{\mathcal{L}}(\M(\mathcal{K}); \mathcal{V}).
\end{align}
\end{dfn}
While the operator itself is a memory and the mapping acts as a memorization process (i.e., memorizing the connections of events in the context), acquiring such effective operator based on the data, is a learning process. Notice that, here, keys and values can be any arbitrary event that memory aims to map them and are not limited to tokens. Later, we will discuss that given a context flow, keys and values might be tokens, gradients, sub-sequences, etc. Furthermore, while the term of associative memory is more common in neuroscience and neuropsychology literature, the above formulation is also closely related to data compression and low-dimensional representation. That is, one can interpret the optimization process in \autoref{eq:attentional-bias-loss} as the training process of a network $\M(.)$ that aims to compress the mappings into its parameters, representing them in a lower dimensional space.

In sequence modeling, where keys and values are input tokens (e.g., tokenized text), the choice of objective and the optimization process for solving \autoref{eq:attentional-bias-loss} can result in distinct sequence modeling architectures~(see \cite{liu2024longhorn} and \cite{behrouz2025Miras}) such as global/local softmax attention~\citep{transformers}, or other modern recurrent models~\citep{katharopoulos2020transformers, sun2023retentive, behrouz2024titans}. This simple formulation of sequence models provides us with better understanding of their internal process and also a tool to simply compare their modeling power based on their objective and optimization process. In the following, using step-by-step examples, we discuss how this formulation can be applied to all components of a neural architecture (including its optimization process in pre-training) and in fact, how a model is an integrated system of multi-level, nested, potentially parallel memories, each of which with its own context flow.

\head{A Simple Example of MLP Training}
We start with a simple example, in which we aim to train a 1-layer MLP (parameterized with $W$) for task $\mathcal{T}$ and on dataset $\mathcal{D}_{\text{train}} = \{x_1, \dots, \boldsymbol{x}_{|\mathcal{D}_{\text{train}}|}\}$ by optimizing the objective $\mathcal{L}(\cdot; \cdot)$ with gradient descent. In this case, the training process objective is to solve the following optimization problem:
\begin{align}\label{eq:example1}
    W^{*} = \arg\min_{W} \:\: \mathcal{L}(W; \mathcal{D}_{\text{train}}), 
\end{align}
whose optimization by (stochastic/online) gradient descent results in a weight update rule:
\begin{align}\label{eq:gradient-dual}
    W_{t+1} = W_{t} - \eta_{t+1} \undermath{\text{Surprise}}{\nabla_{W} \mathcal{L}(W_t; \boldsymbol{x}_{t+1})} = W_{t} - \eta_{t+1} \undermath{\text{Surprise in the Output}}{\nabla_{y_{t+1}} \mathcal{L}(W_t; \boldsymbol{x}_{t+1})} \otimes \boldsymbol{x}_{t+1}, \qquad \text{where}\:\: \boldsymbol{x}_{t+1} \thicksim \mathcal{D}_{\text{train}},
\end{align}
where $y_{t+1}=Wx_{t+1}$ is the output of the model for input $x_{t+1}$ and we used the simplifying notation: $\nabla_{y_{t+1}} \mathcal{L}(W_t; \boldsymbol{x}_{t+1}):= \frac{\partial \mathcal{L}}{\partial y}|_{y = W x_{t+1}}$.  In this case, $\nabla_{W} \mathcal{L}(W_t; \boldsymbol{x}_{t+1})$ is the surprise metric that shows how much the current input is different from previously observed data. Similarly, $\nabla_{y_{t+1}} \mathcal{L}(W_t; \boldsymbol{x}_{t+1})$ is the surprised metric for the output (or more accurately \emph{local surprise signal in representation space} that quantifies the mismatch between the current output and the structure the objective $\mathcal{L}(\cdot; \cdot)$ enforces)–measuring how much surprising the model's prediction is for this input. Given this formulation, one can let the surprise value of output be $u_{t+1} = \nabla_{y_{t+1}} \mathcal{L}(W_t; \boldsymbol{x}_{t+1})$ and reformulate the backpropagation process {as the solution to an optimization problem} on finding an  associative memory that maps input data points $\mathcal{D}_{\text{train}} = \{x_t\}_{t=1}^{|\mathcal{D}_{\text{train}}|}$ to their corresponding $u_{t+1} = \nabla_{y_{t+1}} \mathcal{L}(W_t; \boldsymbol{x}_{t+1})$. That is, we let $\M(\cdot) = W_t \: \cdot$ parametrizes the memory, and use dot-product similarity to measure the quality of $W_t$'s mapping between $x_{t+1}$ and $\nabla_{y_{t+1}} \mathcal{L}(W_t; \boldsymbol{x}_{t+1})$:
\begin{align}
    W_{t+1} = \arg\min_{W} \:\:  \inner{W \boldsymbol{x}_{t+1}}{u_{t+1}} + \frac{1}{2\eta_{t+1}}\:\|W - W_{t} \|^{2}_2 = \arg\min_{W} \:\:  \inner{W \boldsymbol{x}_t}{\nabla_{y_{t+1}} \mathcal{L}(W_t; \boldsymbol{x}_{t+1})} + \frac{1}{2\eta_{t+1}}\:\|W - W_{t} \|^{2}_2.
\end{align}
 Therefore, this formulation translates the training phase of the model as a process of acquiring effective memory that maps data samples to their Local Surprise Signal (LSS) in representation space–measuring how surprising its corresponding output is. This gradient can be viewed as an error in the prediction (with gradient being zero when the loss is minimized). Later in \autoref{sec:DeepOptimizers}, we discuss the backpropagation process as an associative memory in more details, but as a preliminary takeaway from this simple example: 

\vspace{8pt}
 \begin{myboxi}[Training a Linear Layer with Backpropagation as a Surprise-based Memory:]
     \textit{A linear layer trained with backpropagation learns from data by memorizing how surprising their predicted outputs are; i.e., backpropagation can be viewed as an associative memory that maps each data sample to the error of its corresponding prediction. }
 \end{myboxi}
\vspace{8pt}

 Accordingly, in this example, our model has \emph{a single gradient flow} over the data samples, which is only active over dataset $\mathcal{D}_{\text{train}} = \{x_1, \dots, \boldsymbol{x}_{|\mathcal{D}_{\text{train}}|}\}$ and \textit{will be frozen for any other data samples afterwards (i.e., inference or test time). }

In the above example, we can replace the gradient descent algorithm with its  momentum-based variant, resulting in the update rule of:
\begin{align}\label{eq:grad-mlp-example}
    W_{t+1} &= W_t - \vm_{t+1},\\ \label{eq:momentum-mlp-example}
    \vm_{t+1} &= \vm_{t} + \eta_{t+1} \nabla_{W} \mathcal{L}(W_t; \boldsymbol{x}_{t+1}) = \vm_{t} + \eta_{t+1} \nabla_{y_{t+1}} \mathcal{L}(W_t; \boldsymbol{x}_{t+1}) \otimes \boldsymbol{x}_{t+1}.
\end{align}
In \autoref{eq:momentum-mlp-example}, given the previous state of~\autoref{eq:grad-mlp-example} (at time $t$), the value of $\nabla_{W} \mathcal{L}(W_t; \boldsymbol{x}_{t+1})$ or similarly $\nabla_{y_{t+1}} \mathcal{L}(W_t; \boldsymbol{x}_{t+1})$ does not depend on the output of recurrence in \autoref{eq:momentum-mlp-example} and so can be pre-computed beforehand: Leting $u_{t+1} = \nabla_{W} \mathcal{L}(W_t; \boldsymbol{x}_{t+1})$,  \autoref{eq:momentum-mlp-example} can be reformulated as:
\begin{align}\label{eq:reformulate-grad-mlp-example}
    W_{t+1} &= W_t - \vm_{t+1},\\ \label{eq:reformulate-momentum-mlp-example}
    \vm_{t+1} &= \arg\min_{\vm} \:\: - \inner{\vm}{\nabla_{W_t} \mathcal{L}(W_t; \boldsymbol{x}_{t+1})} + \frac{1}{2\eta_{t+1}}\:\| \vm - \vm_t \|^2_2 = \arg\min_{\vm} \:\: - \inner{\vm \: \boldsymbol{x}_{t+1}}{\nabla_{y_{t+1}} \mathcal{L}(W_t; \boldsymbol{x}_{t+1})} +\frac{1}{2\eta_{t+1}} \:\| \vm - \vm_t \|^2_2.
\end{align}
 
 Given this formulation, one can interpret the momentum term as either: (1) a value-less associative memory that compress the gradients into its parameters, or (2) an associative memory that learns how to map data points to their corresponding LSS-value. Interestingly, this formulation reveals that gradient descent with momentum can be viewed as a two-level optimization procedure, where the memory is optimized by simple gradient descent algorithm\footnote{We use the term two- or multi-level to describe the optimization \textit{procedure}. This differs from classical multi-level optimization, where the optimization \textit{problems} are arranged hierarchically.}.

Concluding the above examples, we observed that the training process of a 1-layer MLP with: (1) Gradient descent is a \emph{1-level} associative memory that learns how to map data points to their corresponding LSS-value; and (2) Gradient descent with momentum is a \emph{2-level} associative memory (or optimization process) that the inner-level learns to store gradient values into its parameters, and then the outer-level updates the slow weight (i.e., $W_t$) with the value of the inner-level memory. While these are the most simple examples with respect to both architecture and optimizer algorithms, one might ask if similar conclusion can be made in more complex setups.

\head{An Example of Architectural Decomposition}
In the next example, we replace the MLP module in our previous example with a linear attention~\citep{katharopoulos2020transformers}. That is, we aim to train a 1-layer linear attention for task $\mathcal{T}$ and on a sequence of $\mathcal{D}_{\text{train}} = \{x_1, \dots, \boldsymbol{x}_{|\mathcal{D}_{\text{train}}|}\}$ by optimizing the objective $\mathcal{L}$ with gradient descent. Recalling the unnormalized linear attention formulation: 
\begin{align}\label{eq:linear-attn-projection}
    & \vk_t =  W_{\vk} \boldsymbol{x}_t , \qquad \vv_t =  W_{\vv} \boldsymbol{x}_t, \qquad \vq_t =  W_{\vq}\boldsymbol{x}_t,   \\ \label{eq:linear-attention-recurrence}
    &\M_{t} = \M_{t-1} + \vv_t \vk_t^{\top}, \\ \label{eq:linear-attn-output}
    &y_t = \M_t \vq_t \:.
\end{align}
As discussed in earlier studies~\citep{liu2024longhorn, behrouz2025Miras}, the recurrence in \autoref{eq:linear-attention-recurrence} can be reformulated as the optimization process of a matrix-valued associative memory $\M_t(\cdot)$ to compress the mappings of keys and values into its parameters. Specifically, in Definition~\ref{def:associative-memory}, if we let $\tilde{\mathcal{L}}(\M_{t-1}; \vk_t, \vv_t) := - \inner{\M_{t-1} \vk_t}{\vv_t}$ and aim to optimize the memory with gradient descent, the memory update rule is: (Note that $\nabla \tilde{\mathcal{L}}(\M_{t-1}; \vk_t, \vv_t) = - \vv_t \vk_t^{\top}$ and we let learning rate $\eta_{t} = 1$)    
\begin{align}\label{eq:linear-attention-optimization}
    &\M_{t+1} = \arg \min_{\M} \:\: - \inner{\M \vk_{t+1}}{\vv_{t+1}} + \frac{1}{2}\| \M - \M_{t} \|^2_2 \quad  \\ 
    \Rightarrow \:\: &\M_{t+1} = \M_t - \nabla \tilde{\mathcal{L}}(\M_t; \vk_{t+1}, \vv_{t+1}) = \M_t + \vv_{t+1} \vk_{t+1}^{\top},
\end{align}
which is equivalent to the update rule of an unnormalized linear attention in \autoref{eq:linear-attention-recurrence}. Also, as we observed in the first example, training a linear layer with gradient descent can be viewed as a 1-level optimization of an associative memory (\autoref{eq:gradient-dual}) and so the general training/updating process of projection layers  (i.e., $W_{\vk}, W_{\vv},$ and $W_{\vq}$) is itself an optimization process of associative memory. Therefore, training a linear attention with gradient descent can be seen as a two-level optimization process, where the outer-loop (also known as training process) optimizes the projection layers with gradient descent, while the inner-loop optimizes the inner memory of $\M_t$ with gradient descent.

In the examples we discussed so far, we have two associative memories,  each of which has their own optimization process and gradient flow. That is, in the optimization of the outer-level parameters of $W_{\vk}, W_{\vv},$ and $W_{\vq}$ there is no gradient with respect to parameter $\M(\cdot)$ and so there is no backpropagation through it. Similarly, in the inner-level, there is no backpropagation through projection layers and they are considered frozen. Furthermore, it is notable that in this example, the above formulation is also closely connected to FWPs perspective of linear attentions~\citep{schlag2021linear}, where projections are considered slow weights, and memory update in \autoref{eq:linear-attention-recurrence} is the fast weight update rule.

\head{Architectural Decomposition with More Levels} In both  examples above, we discussed how they can be viewed as a 2-level optimization process (coinciding with their FWPs interpretations). In practice, however, we may need to use more powerful optimization process, and/or more powerful recurrent update rules for memory. As a simple example, assume we use gradient descent with momentum to train a linear attention model. As we saw above,  the linear attention component can be decomposed into two nested optimization processes. Similarly,  the model here can be represented as a 2-level optimization problem, where (1) the inner level optimizes the memory to compress the context using gradient descent (\autoref{eq:linear-attention-optimization}), and (2) the outer level optimizes the projection layers with gradient descent with momentum. Interestingly,  we saw that ``gradient descent with momentum'' algorithm itself can be viewed as a 2-level optimization process where the momentum term itself is an associative memory that compress the past gradients into its parameters.

\begin{figure*}
    \centering
    \includegraphics[width=0.9\linewidth]{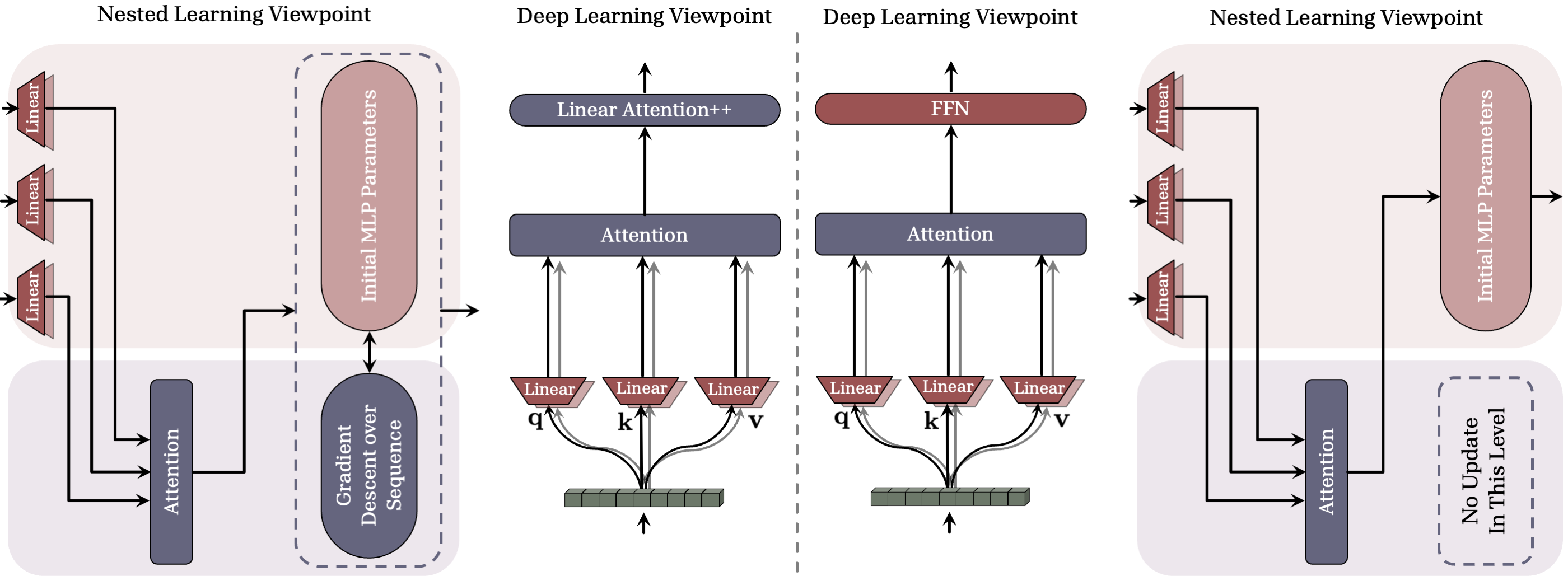}
    \caption{An example of comparing a FFN (e.g., MLP) with linear attention in a Transformer-based backbone, optimizing with gradient descent. The \textcolor{mydarkred}{red components} are blocks in the first level (with frequency 1), while \textcolor{mydarkpurple}{blue components} are blocks in the second level (frequency $L$). Linear attention with learnable initial memory state (referred to as Linear Attention++) is the same as an MLP layer but with in-context learning ability and adaptation to the input sequence.}
    \label{fig:mlp-vs-lin-attention}
\end{figure*}

\subsection{Nested Optimization Processes}\label{sec:nop}
In the previous section, we provided examples to demonstrate how one can decompose a machine learning model into a set of nested or multi-level optimization procedures. Next, we first  present a formalization of nested learning problems and then define Neural Learning Module–an integrated computational system that learns from~data.    

In previous sections,  we decomposed the model into a set of optimization process. However, it is still unclear if we can define a hierarchy (or order) over these processes, and {uniquely represent the model in this format.} Inspired by the hierarchy of brain waves that indicates the information processing frequency rate of each part (discussed in \autoref{sec:intro}), we use the update rate of each optimization process to order the components in multiple levels. To this end, we let the one update step over one data point to be the unit of time, and define the update frequency rate of each component as: 
\begin{dfn}[Update Frequency]\label{dfn:uf}
    For any component of $A$, which can be a parametric component (e.g., learnable weights or momentum term in gradient descent with momentum) or a non-parametric component (e.g., attention block), we define its frequency, denoted by $f_A$, as its number of updates per unit of time.   
\end{dfn}
Given the above update frequency, we can order the components of a machine learning algorithm based on operator $(\cdot \succ \cdot)$. We say $A$ is faster than $B$ and denote $A \succ B$ if: \textbf{(1)}  $f_A > f_B$, or \textbf{(2)} $f_A = f_B$ but the computation of the $B$'s state at time $t$ requires the computation of $A$'s state at time $t$. In this definition, when $A \nsucc B$ and $B \nsucc A$, we let $A \overset{f}{=} B$, which indicates that $A$ and $B$ has the same frequency update, but their computation is independent of each other (Later, we provide an example of this cases in AdamW optimizer).  Based on the above operator, we sort the components into an ordered set of ``\emph{levels}'', where (1) components in the same level have the same frequency update, and (2) the higher the level is, the lower its frequency. Given the above formulations of levels and update frequency, we next formally define nested learning:

\begin{dfn}[Nested System]
    A (ordered) nested system is a system with $K$ (ordered) levels such that each level $k$, $1 \leq k \leq K$, consists of a set of optimization problems $\{ (\mathcal{L}^{(k)}_{i},  \mathcal{C}^{(k)}_i, \boldsymbol{\Theta}^{(k)}_{i}) \}_{i = 1}^{N_k}$, where $\mathcal{L}_{i}(\cdot; \cdot)$ is the optimization objective in the $i$-th problem, $\mathcal{C}_i$ is its context (the data that is optimized on), $\boldsymbol{\Theta}_{i}$ is the feasible set of its parameters, and each parameter is optimized using gradient descent:
    \begin{align}
        {\boldsymbol{\theta}_{i}}^{(k)}_{t+1} = \arg\min_{\boldsymbol{\Phi}^{(k)}_{i}} \: \inner{\boldsymbol{\Phi}^{(k)}_{i} \boldsymbol{x}_{t+1}}{- \nabla \mathcal{L}^{(k)}_i(\boldsymbol{\theta}^{(k)}_{i_t};\boldsymbol{x}_{t+1})} + \frac{1}{2{\eta_{i}}^{(k)}_{t+1}}\:\|\boldsymbol{\Phi}^{(k)}_{i} - {\boldsymbol{\theta}_{i}}^{(k)}_{t} \|^{2}_2 \quad \text{where}\:\:\boldsymbol{x}_{t+1} \thicksim \mathcal{C}^{(k)}_i, \:\:\: \text{and} \:\:\: {\boldsymbol{\Phi}_{i}}^{(k)} \in  \boldsymbol{\Theta}^{(k)}_{i}\!\!.
    \end{align}
\end{dfn} 

Notice that each  optimization process has its own gradient flow, and therefore sometimes  we refer to them as a box of gradient flow corresponding to an optimization problem. Through this paper, we further generalize our definition of nested systems, and allow finding non-parametric solutions for some boxes (i.e., optimization problems).

The above definition provides a general flexible definition for a nested system that does not specify if there is any dependence among different boxes  (i.e., a box can determines the context or the parameter space of another box). In the next sections, we discuss how  knowledge/information can be transferred between different levels or boxes. Throughout this paper, we focus on the Nested Systems of Associative Memories (\nsam) that is a nested system, in which each optimization process is an associative-memory. More formally,

\begin{dfn}[Nested System of Associative Memories] \label{dfn:nsam}
    A nested system of associative memory (\nsam) is a system with $K$ (ordered) levels such that each level $k$, $1 \leq k \leq K$, consists of a set of optimization problems $\{ (\mathcal{L}^{(k)}_{i},  \mathcal{C}^{(k)}_i, \boldsymbol{\Theta}^{(k)}_{i}) \}_{i = 1}^{N_k}$, where $\mathcal{C}_i = \{(\vk^{(i)}_j, \vv^{(i)}_j) \}_{j = 1}^{L_{i}}$ is a set of key-value pairs, $\mathcal{L}_{i}(\cdot; \cdot, \cdot)$  measures the quality of memory learned mappings in the $i$-th problem, $\boldsymbol{\Theta}_{i}$ is the set of feasible memory parameters with each parameter optimized using gradient descent:
    \begin{align}
        {\boldsymbol{\theta}_{i}}^{(k)}_{t+1} = \arg\min_{\boldsymbol{\Phi}^{(k)}_{i}} \:\:  \inner{\boldsymbol{\Phi_{i}^{(k)}} \: \vk^{(i)}_{t+1}}{- \nabla \mathcal{L}^{(k)}_i(\boldsymbol{\theta}^{(k)}_{i_t}; \vk^{(i)}_{t+1}, \vv^{(i)}_{t+1})} + \frac{1}{2{\eta_{i}}^{(k)}_{t+1}}\:\|\boldsymbol{\Phi}^{(k)}_{i} - {\boldsymbol{\theta}_{i}}^{(k)}_{t} \|^{2}_2,
    \end{align}
    where $(\vk^{(i)}_{t+1}, \vv^{(i)}_{t+1}) \thicksim \mathcal{C}^{(k)}_i$ and ${\boldsymbol{\Phi}_{i}}^{(k)} \in  \boldsymbol{\Theta}^{(k)}_{i}$. 
\end{dfn} 

Given a query $\vq$, for each associative memory $\M_{i}^{(k)}$,
we use $\M_{i}^{(k)}(\vq)$ to refer to the forward pass process (i.e., retrieval process) of the memory. While our formulation of \nsam can simply be defined by any optimization process beyond gradient descent, and initially, it may seem that the strict condition of \emph{optimization by gradient descent} can limit the modeling power of the definition, through this paper, we show that modern architectures alongside certain well-known optimization algorithms can be viewed as instances of \nsam. We then build on top of this intuition and discuss how to go further with stacking multiple levels and design models with enhanced continual learning capabilities. 

The new dimension of stacking multiple levels is an important characteristic of nested learning, where the depth of computation can be enhanced with increasing the number of levels. Depending on the design, context, and the type of knowledge transfer, this depth of computation can itself be viewed as different concepts such as: higher-order in-context learning ability, latent computation (e.g., Loop Transformers), multiple memory systems, and more expressive optimizers. Later, we discuss all such implications, but next, we use a simple example that connects MLP layers in Transformer architectures with linear or deep memory blocks.

\head{Example of an MLP Layer vs. Linear Attention}
Let us compare two models: (1) a Transformer architecture, and (2) the same backbone but with replacing the MLP block with a linear attention mechanism (sharing the keys and values from previous layer), in which the initial state of its memory is meta-learned (similar to \citet{behrouz2024titans} or \citet{sun2024learning}). We refer to the second variant as Adaptive Transformer or AdaTransformer. Both models are optimized on Next Token Prediction (NTP) objective with gradient descent. As discussed in the initial examples, and also illustrated in \autoref{fig:mlp-vs-lin-attention}, both models have two levels, and for the sake of clarity, we use \textcolor{mydarkred}{red} (resp. \textcolor{c4}{blue}) to highlight computations/weight in the first level (resp. second level). More formally, let $X = \{x_i\}_{i = 1}^{T}$ be an input sequence of tokens, the output of both blocks are computed as (For the sake of simplicity, we assume $\texttt{MLP}(\cdot) = \cdot \: W_{\texttt{MLP}} \:$, and remove normalizations):

\vspace{-8pt}
\begin{minipage}[t]{0.48\textwidth}
\begin{align} \nonumber
    &\vk_t =\boldsymbol{x}_t \textcolor{mydarkred}{W_{\vk}} , \qquad \vv_t =\boldsymbol{x}_t \textcolor{mydarkred}{W_{\vv}} , \qquad \vq_t =\boldsymbol{x}_t \textcolor{mydarkred}{W_{\vq}}, \\ \nonumber
    &\boldsymbol{y}_\texttt{attn} = \textcolor{c4}{\texttt{Attn}}\left(\vk, \vv, \vq \right),\\ \tag{Transformer Block}
    &\boldsymbol{y}_{\texttt{block}} = \textcolor{mydarkred}{\texttt{MLP}} \left( \boldsymbol{y}_\texttt{attn} \right) = \boldsymbol{y}_\texttt{attn} \: \textcolor{mydarkred}{W_{\texttt{MLP}}},
\end{align}
\end{minipage}
\hspace{3.6ex}~
\textcolor{mydarkpurple}{\vline}~
\hspace{-2.6ex}~
\begin{minipage}[t]{0.48\textwidth}
    \begin{align} \nonumber
    &\vk_t =\boldsymbol{x}_t \textcolor{mydarkred}{W_{\vk}} , \qquad \vv_t =\boldsymbol{x}_t \textcolor{mydarkred}{W_{\vv}} , \qquad \vq_t =\boldsymbol{x}_t \textcolor{mydarkred}{W_{\vq}}, \\ \nonumber
    &\boldsymbol{y}_\texttt{attn} = \textcolor{c4}{\texttt{Attn}}\left(\vk, \vv, \vq \right),\\ \tag{AdaTransformer Block}
    &\boldsymbol{y}_{\texttt{block}} = \boldsymbol{y}_\texttt{attn} \: \textcolor{c4}{W_{\texttt{LinAttn}}}.
\end{align}
\end{minipage}

~

The formulation of both blocks seem to be very similar and the only difference comes from the level of $W_{\texttt{MLP}}$ and $W_{\texttt{LinAttn}}$ weights. That is, while $W_{\texttt{MLP}}$ is in the first level and so is persistent with respect to the context, $W_{\texttt{LinAttn}}$ is adaptive and is updated in-context by ($\M(\cdot)$ is parametrized by ${W_{\texttt{LinAttn}}}$):
\begin{align}
    \textcolor{c4}{\M_{t} = \M_{t-1} + \vv_t\vk^{\top}_t}.
\end{align}
In earlier variants of linear attention, the initial state of $\M(\cdot)$ or equivalently ${W_{\texttt{LinAttn}}}$ is considered as zero matrix,  $\M_0 = \mathbf{0}$. Similar to more advanced design choices~\citep{sun2024learning, behrouz2024titans}, however, this initial state can be meta-learned to adapt fast to a context. In this setting, the initial state of $\textcolor{mydarkred}{\M_0 = W_{\texttt{LinAttn}_{\texttt{init}}}}$ is optimized in the first level with NTP objective, while given the context, $\textcolor{c4}{W_{\texttt{LinAttn}}}$ is optimized in the second level as an associative memory with dot-product objective~\citep{behrouz2025Miras}. 

The above example is also valid when using more advanced and deep MLP blocks in Transformer architecture (such as SwiGLU~\citep{shazeer2020glu}) and compared it with its recurrent memory counterpart~\citep{behrouz2025atlas}. Furthermore, this simple example implies that the current perspective on hybrid architectures as the combination of expressive softmax attention and efficient recurrent models is somewhat misleading and it follows the conventional Transformer backbone design but with additional in-context learning capabilities for MLP blocks. We discuss this further in \autoref{sec:revisit} and \autoref{sec:cmlp}. As a takeaway of the discussion in this subsection about the concept of nested systems and nested learning:

\vspace{8pt}
\begin{myboxi}[Stacking Levels in Nested Learning:]
\textit{\textbf{Nested learning} allows computational models that are composed of multiple (multi-layer) levels to learn from and process data with different levels of abstraction and frequencies of update.}
\end{myboxi}
\vspace{8pt}

As discussed earlier, it is common in the literature to separate architectures from their optimization processes and to treat them as independent design choices, with the aim of combining algorithms that achieve the greatest expressive power in each aspect. In practice, however, a Transformer architecture~\citep{transformers} that is optimized with stochastic gradient descent can learn  a very different solution than the same architecture when  Adam optimizer is used~\citep{kingma2014adam}. Accordingly, when interacting with such machine learning algorithms, we observe that despite the similarity in the architecture axes, the overall trained models show different predictions or generates different outputs. From the NL's viewpoint, however, a machine learning algorithm is represented as an interconnected system of optimization problems and model's actions, predictions, and generation of output depends on this system as a whole, not necessarily each of its sub-components. To this end, we define the term of neural learning module to refer to this representation of a model, where architecture and optimization process jointly determine the model and its outputs. While such joint representation might not seem significant in the current machine learning pipelines, where there is a training and then test phases, it becomes more important in the continual setup we advocate for, where there is no training/test phases (see more discussions in \autoref{sec:hope}).

\head{Neural Learning Modules are Inter-connected Systems}  
Based on the definition of neural learning module, one important question is how the architecture and optimization process are interconnected systems and how they can affect each other. Recall the general formulation for training a neural network: Given a task $\mathcal{T}$, its corresponding data distribution $p(\mathcal{T})$, a model $f(\cdot; \cdot)$ parameterized by $\Phi_{\mathcal{T}}$, and an objective $\mathcal{L}(\cdot; \cdot)$ we aim to learn parameters $\Phi^{*}_{\mathcal{T}}$ such that:
\begin{align}
    \Phi^{*}_{\mathcal{T}} = \arg \min_{\Phi}\: \:\mathbb{E}_{\:\vx, \boldsymbol{y} \sim p(\mathcal{T})} \left[ \mathcal{L}\left(\Phi; \vx, \boldsymbol{y}\right) \right].
\end{align}
In practice, we optimize the above problem based on a given dataset $\mathcal{D}_{\text{train}}$ and an optimization algorithm such as stochastic gradient descent:
\begin{align}\label{eq:general-model-opt}
    \Phi_{t+1} = \Phi_{t} - \eta_{t+1} \nabla_{\Phi_t} \mathcal{L}(\Phi_t; \vx_{t+1}, \boldsymbol{y}_{t+1}), \qquad \text{where}\:\: (\vx_{t+1}, \boldsymbol{y}_{t+1}) \thicksim \mathcal{D}_{\text{train}}.
\end{align}
One interpretation for the optimization process of model $f(\cdot; \cdot)$ in \autoref{eq:general-model-opt} is to see the model as the data generator for the optimization process in \autoref{eq:general-model-opt}. That is, as discussed in the first example in \autoref{sec:associaitve-memory}, and later we will show in \autoref{sec:DeepOptimizers}, the optimization process is an associative memory that aims to compress the patterns between the training data and its gradients (or surprise) and so the dataset for internally training such memory (i.e., the gradients of the model) is generated by the model. Therefore, the type of the model can result in generating dataset (i.e., gradients) with different patterns and distributions over time. The effect of the optimization process and this data generation also fed back at the model itself, where the next state of the parameters in the model are determined by the optimization algorithms. As we will discuss in \autoref{sec:DeepOptimizers}, looking at optimizers as associative memories on the gradients of the model implies that each optimizer has some special traits such as better memory management, higher compression, etc. Therefore, the choice of such algorithms requires understanding the generated gradients and also changes of the model in the~parameter~space.

\subsection{Knowledge Transfer Between Levels}\label{sec:knowledge-transfer}
So far, we mainly focused on the concept of nested learning and how optimization problems are located in different levels. It is, however, still unclear that how nested optimization problems (in different levels) can affect each other, or in general how they can contribute to the output of the system, and so be interconnected. In this section, we discuss several potential knowledge transfer methods between components in different levels. For the sake of clarity, we discuss the knowledge transfer between two levels and blocks, i.e., $\mathcal{B}^{(0)} = (\mathcal{L}^{(0)}, \mathcal{C}^{(0)}, \boldsymbol{\Theta}^{(0)})$ and $\mathcal{B}^{(1)} = (\mathcal{L}^{(1)}, \mathcal{C}^{(1)}, \boldsymbol{\Theta}^{(1)})$ with corresponding memories $\M^{(0)}(\cdot)$ and $\M^{(1)}(\cdot)$, respectively:

\head{Direct Connection of Levels (Parametric)} 
The first type of knowledge transfer is to directly incorporate the weights in different levels or blocks. To this end, the forward pass or the retrieval process from the lower-frequency (i.e., higher-level) memory system is also conditioned on the parameters of the higher-frequency (i.e., lower-level) memory:
\begin{align}\label{eq:KT-1}
    \M^{(0)}(\cdot) := \M^{(0)}(\cdot \:;\: \boldsymbol{\Theta}^{(1)}).
\end{align}
In a more specific formulation, and as a special variant of the above formulation, we can condition the output of $\M^{(0)}(\cdot)$ based on the output (or forward pass) of the higher-frequency memory:
\begin{align}
    \M^{(0)}(\cdot) := \M^{(0)}(\cdot \:;\: \M^{(1)}(\cdot)),
\end{align}
where we slightly abused notation by hiding the dependence on the second argument. As an example of this type of knowledge transfer, in linear Transformer (or FWP)~\citep{katharopoulos2020transformers, schlag2021linear}, where the initial memory state is zero, the stored knowledge of a lower level (fast weight) directly affect the output of the model in another level. That is, one can re-write the forward pass (memory retrieval) as:
\vspace{-6pt}
\begin{align}
    \boldsymbol{y}_t = \textcolor{c4}{\M_{t}} \vq_t = \overset{\underset{\text{\small \textcolor{c4}{higher-frequency} memory}}{\text{\small Forward pass of the}}}{\overbrace{\textcolor{c4}{\M_{t}} \undermath{ \underset{\text{\small \textcolor{mydarkred}{lower-frequency} memory}}{ \text{\small Forward pass of the}} }{\left[ \boldsymbol{x}_t \textcolor{mydarkred}{W_q}\right]}{}}}
\end{align}
\vspace{-5pt}
\head{Direct Connection of Levels (Non-Parametric)} 
Another form of direct connection between levels is a non-parametric variant of the above formulation, where the block $\mathcal{B}^{(1)}$ is optimized by finding non-parametric solution. Therefore, the forward pass of the low-frequency memory is conditioned on the context and output of higher-frequency memory:  
\begin{align}
    \M^{(0)}(\cdot) := \M^{(0)}(\cdot \:;\: \mathcal{C}^{(1)}), \qquad \text{or similarly, } \qquad \M^{(0)}(\cdot) := \M^{(0)}(\cdot \:;\: \M^{(1)}(\cdot; \mathcal{C}^{(1)})).
\end{align}
As an example of this variant, one can refer to Transformers and softmax attention module~\citep{transformers}. There is an important characteristic for both above variants: there is no backpropagation through any state of the blocks in two different levels and knowledge transfers through direct conditioning the output of one level on the other's output/parameters. Therefore, in this process the state of each block is treated as hyperparameter for the other.

\head{Knowledge Transfer via Backpropagation}
Another form of knowledge transfer is through backpropagation, where there is a gradient flow between blocks in different levels. The forward pass of this design is the same as the forward pass discussed above. The backward pass, however, is the main difference, where in the above two cases the state of each associative memory is considered as hyperparameters of the other but here both states are optimized in the same gradient flow. Therefore, for a simple case of two blocks in two levels, we have:
\begin{align} \tag{Forward Pass}
    &\:\:\:\:\M^{(0)}(\cdot) := \M^{(0)}(\cdot \:;\: \M^{(1)}(\cdot)) \\ \nonumber
    &\begin{cases}
    \:\boldsymbol{\Theta}^{(1)}_{t+1} = \boldsymbol{\Theta}^{(1)}_{t} - \eta^{(1)}_{t+1} \: \boldsymbol{\delta}_{1} \: \hat{\boldsymbol{x}}_{t+1}^{\top}\:,  \\ \tag{Backward Pass}
    \:\boldsymbol{\Theta}^{(0)}_{t+1} = \boldsymbol{\Theta}^{(0)}_{t} - \eta^{(0)}_{t+1} \: \boldsymbol{\delta}_{0} \: \boldsymbol{x}_{t+1}^{\top}\:,
    \end{cases}
\end{align}
where $\boldsymbol{\delta}^{(0)}_{t+1} = \boldsymbol{J}_{\phi^{(0)}} \left( \boldsymbol{x}_{t+1} \right)^{\top} \left( \boldsymbol{\Theta}^{(1)}_{t+1} \boldsymbol{\delta}^{(1)}_{t+1} \right)$, $\hat{\boldsymbol{x}}_{t+1} = \phi^{(0)}(\M^{(0)}(\boldsymbol{x}_t))$, $\phi^{(0)}(\cdot)$ is non-linearity, and $\boldsymbol{J}_{\phi^{(0)}}(\cdot)$ is the Jacobian. In this design, two blocks are in the same flow of gradient, but are updated based on different frequencies. We provide an example this design in \autoref{sec:cmlp}, when we discuss continuum memory systems.

\head{Knowledge Transfer via Initialization}
Model Agnostic Meta-Learning (MAML)~\citep{finn2017model}, is one of the most popular form of meta-learning (or learning to learn) that aims to learn a global initial point for a model so it could learn fast a new task. From the nested learning perspective, there are two nested optimization processes, in which the inner problem iterates over its own context and based on its internal objective, the higher-level problem measures its own learned weight as the initial point of the inner problem. More formally, we have
\begin{align}
    \boldsymbol{\Theta}^{(1)}_0 =
\underset{\Phi}{\arg\min}
\;\;
\mathbb{E}_{\mathcal{C} \sim \mathcal{C}^{(0)}}
\Biggl[
    \boldsymbol{\ell}(\mathcal{M}^{(1)} \left(\cdot; \Phi\right), \mathcal{C})
\Biggr],
\end{align}
where the higher level block learns the best initial value over all possible contexts the lower level problem might have. As discussed earlier, any MAML-based learning model is an instance of this case, but as a more specific example, we refer to the example of (MLP Layer vs. Linear Attention), which is discussed in \autoref{sec:nop} and \autoref{fig:mlp-vs-lin-attention}.

\head{Connections with Generation}
One of the most common form of knowledge transfer is through generating weights or context. That is, one lower-frequency (resp. higher-frequency) block generates the weight of a higher-frequency (resp. lower-frequency) block. More formally,

\vspace{-9pt}
\begin{minipage}[t]{0.49\textwidth}
\begin{align}\tag{Weight Generation}
    \hspace*{-2ex}\begin{cases}
        \boldsymbol{\Theta}^{(1)} \!= \boldsymbol{g}\left( \M^{(0)}; \mathcal{L}^{(0)}\!\!, \mathcal{C}^{(0)}\!, \boldsymbol{\Theta}^{(0)} \right), & \!\!\!\!\text{or \:} \hspace*{-1.5ex}  \\ 
    \boldsymbol{\Theta}^{(0)} \!= \boldsymbol{g}\left( \M^{(1)}; \mathcal{L}^{(1)}\!\!, \mathcal{C}^{(1)}\!, \boldsymbol{\Theta}^{(1)} \right),
    \end{cases}
\end{align}
\end{minipage}
\hspace{0.0ex}~
\textcolor{mydarkpurple}{\vline}~
\hspace{-5.5ex}~
\begin{minipage}[t]{0.53\textwidth}
    \begin{align}\tag{Context Generation}
    \begin{cases}
        \mathcal{C}^{(1)} \!= \boldsymbol{g}\left( \M^{(0)}; \mathcal{L}^{(0)}\!\!, \mathcal{C}^{(0)}\!, \boldsymbol{\Theta}^{(0)} \right), & \!\!\!\! \text{or} \hspace*{-3ex} \\ 
    \mathcal{C}^{(0)} \!= \boldsymbol{g}\left( \M^{(1)}; \mathcal{L}^{(1)}\!\!, \mathcal{C}^{(1)}\!, \boldsymbol{\Theta}^{(1)} \right).
    \end{cases}
\end{align}
\end{minipage}
~

There are two important examples of the above form for knowledge transfer: (1) Hypernetworks: where the weights of a targeted neural network is generated by another (generator) network. (2) Optimization process: where the architecture generates the input for the optimizer. That is, the context (or input data) of an optimizer is the gradients that are generated by the architecture. For more discussion on this topic, see \autoref{sec:DeepOptimizers}. Note that this example is not necessarily about ``\emph{learned optimizers}'' and it is valid for the commonly used optimization process and algorithms such as gradient descent, Adam~\citep{kingma2014adam}, AdaGrad~\citep{duchi2011adaptive}, etc.

\head{A Note on Designing Neural Learning Modules} 
In the above, we discussed only some examples of possible knowledge transfer methods and also the potential connections of different levels. The formulation of NL and neural learning module, however, is general and so is not limited to the above specific set of methods. Accordingly, to design a neural learning module from a nested learning perspective, there are two important steps and design choices:

\begin{myboxi}[Designing Neural Learning Modules:]
    \textit{There are two high-level design choices in developing a neural learning module: (1) The design of optimization problems and their frequency (i.e., designing components in \nsam); (2) The design of knowledge transfer between levels.}
\end{myboxi}
It is notable that with different choices of knowledge transfer, some learning paradigms can be seen as a part of a neural learning model: E.g., (1) Meta learning, when two blocks in two levels transfer their knowledge with one level meta-learns the other; more specifically, (2) Model Agnostic Meta Learning (MAML)~\citep{finn2017model}, when knowledge transfer is though learning the initialization; (3) Hypernetworks, when one higher-frequency block generates the weights for the other lower-frequency block; (4) Learned optimizers, when the knowledge transfer is through data generation (i.e., one high-frequency block generates the gradient for the other lower-frequency block).

\section{Optimizers as Learning Modules}\label{sec:DeepOptimizers}
In this section, we start with viewing backpropagation process and optimizing a neural network from the associative memory and data compression perspective. Then, we discuss how variants such as momentum-based optimizers are instances of nested associative memory systems. Finally, we discuss alternative methods leading to deep optimizers with higher expressive power from the associative memory perspective.

\subsection{Backpropagation as an Associative Memory}\label{sec:backprop-memory-referential}
Updating the weights of a neural network through backpropagation~\citep{linnainmaa1970representation, rumelhart1986learning} has been the critical component of training large-scale deep neural networks. Intuitively, in this optimization process, first, the error of the model's output with respect to target is calculated, and then each layer is updated based on its contribution to this error. This section aims to explain this process through the lens of associative memory and discuss how it fits within the nested learning paradigm. For the sake of clarity and simplicity, we assume a deep MLP model, but all the derived formulations in the following can simply be adapted to other architectures as well. Given an MLP with $L$ layers parameterized with $\{W_{\ell} \cdot + \: \boldsymbol{b}_{\ell}\}_{\ell = 1}^{L}$, the required gradients in backpropagation are computed as:
\begin{align}\label{eq:backprop-delta-jacobian}
     \frac{\partial \mathcal{L}}{\partial W_{\ell}} = \boldsymbol{\delta}_{\ell} \: \hat{\boldsymbol{x}}_{\ell - 1}^{\top}, \qquad \text{and} \qquad \boldsymbol{\delta}_{\ell} = \undermath{\text{local output surprise for layer $\ell$}}{\boldsymbol{J}_{\phi_\ell} \left( \boldsymbol{z}_{\ell} \right)^{\top} \left( W_{\ell+1}^{\top} \boldsymbol{\delta}_{\ell + 1} \right)},
\end{align}
where $\boldsymbol{z}_{\ell} = W_{\ell} \:\hat{\boldsymbol{x}}_{\ell - 1} + \boldsymbol{b}_{\ell}$ is pre-activation, and so  $\hat{\boldsymbol{x}}_{\ell} = \phi_{\ell}\left( \boldsymbol{z}_{\ell}  \right)$ is the output of $\ell$-th layer, $\phi_{\ell}(\cdot)$ is its non-linearity, and $\boldsymbol{J}_{\phi_\ell}(\cdot)$ is the Jacobian. Therefore, the update of the $\ell$-th layer with gradient descent is computed as:
\begin{align}\label{eq:general-backprop}
    {W_{\ell}}_{_{t+1}} = {W_{\ell}}_{_t} - \eta_{\ell_{t+1}} \: \boldsymbol{\delta}_{\ell} \: \hat{\boldsymbol{x}}_{\ell - 1}^{\top}\:.
\end{align}
Here, $\hat{\boldsymbol{x}}_{\ell - 1}$ is the input of the layer and $\boldsymbol{\delta}_\ell$ measures the local error signal for layer $\ell$ or equivalently is a metric that measures the surprise of layer $\ell$'s output given its input. Similar to our example in \autoref{sec:associaitve-memory}, we can write~\autoref{eq:general-backprop}~as:
\begin{align}\label{eq:backprop-simple}
    {W_{\ell}}_{_{t+1}} = \arg \min_{W} \: \inner{W \hat{\boldsymbol{x}}_{\ell - 1}}{ \boldsymbol{\delta}_{\ell}} + \frac{1}{2 \eta_{\ell_{t+1}}}\| W - {W_{\ell}}_{_{t}} \|_F^{2}, 
\end{align}
which is an associative memory module that aim to map the input of each layer $\hat{\boldsymbol{x}}_{\ell - 1}$ to its local error signal, $\boldsymbol{\delta}_{\ell}$ (see Definition~\ref{def:associative-memory}). That is, the above formulation implies that \textit{the training process of a neural network with gradient descent and backpropagation can be viewed as a compression process}, in which each layer stores the mappings between its input and the corresponding local error signal. Later in \S\ref{sec:L2-backprop}, we discuss how this viewpoint helps with designing more expressive learning rules~for~backpropagation.  

\vspace{8pt}
 \begin{myboxi}[Training a Deep Neural Network with Backpropagation as a Surprise-based Memory:]
     \textit{A neural network trained with backpropagation learns from data by memorizing how surprising their predicted outputs are; i.e., backpropagation is an associative memory that maps each data point~to~the~error~in~its~corresponding~prediction. }
 \end{myboxi}

\head{Backpropagation $\neq$ Linear Attention} A common misinterpretation for \autoref{eq:general-backprop} is to assume $\boldsymbol{\delta}_{\ell}$ is a pre-computed term and so backpropagation (at least on a linear layer) recovers Hebbian-rule, resulting in the equivalency of the optimization process and performing linear attention on gradients. Our formulation, however, shows that the update rule in backpropagation is a self-referential process~\citep{schmidhuber1993self}, where the values of the associative memory is generated by itself, making it a more complex associative memory than a simple linear attention on gradients (see \autoref{sec:L2-backprop}).

\subsection{Momentum-based Optimizers as Associative Memories}\label{sec:momentum-based}
Momentum-based optimizers are the major components of modern machine learning models' training~\citep{kingma2014adam, jordanmuon, duchi2011adaptive}. To explain momentum-based optimizers as associative memories, let us start from a simple gradient descent algorithm:
\begin{align}
    W_{t+1} = W_{t} - \eta_t \nabla_{W_{t}} \mathcal{L} (W_t; \boldsymbol{x}_{t+1}),
\end{align}
which updates the current state of the weights based on the momentary gradient (surprise). This update rule does not incorporate the previous tokens and also the loss landscape that have been traversed so far, resulting in slower (or less robust) convergence in many scenarios. To fix this, momentum-based gradient descent methods incorporate an Exponential Moving Averages (EMAs) of past gradients:
\begin{align}\nonumber
    &{W_{\ell}}_{_{t+1}} = {W_{\ell}}_{_{t}} + {\boldsymbol{m}_{\ell}}_{_{t+1}}\\ \label{eq:momentum1}
    &{\boldsymbol{m}_{\ell}}_{_{t+1}} = \alpha_{\ell, t+1} {\boldsymbol{m}_{\ell}}_{_{t}} - \eta_{\ell, t+1} \nabla_{{W_{\ell}}_{_{t}}} \mathcal{L}\left({W_{\ell}}_{_{t}}; \boldsymbol{x}_{t+1}\right) = \alpha_{\ell, t+1} {\boldsymbol{m}_{\ell}}_{_{t}} - \eta_{\ell, t+1} \: \boldsymbol{\delta}_{\ell} \: \hat{\boldsymbol{x}}_{\ell - 1}^{\top}, 
\end{align}
where matrix (or vector) $\boldsymbol{m}_t$ is the momentum at state $t$ and $\alpha_t$ and $\eta_t$ are (adaptive) learning and momentum rates, respectively, and $\boldsymbol{\delta}_{\ell}$ and $\hat{\boldsymbol{x}}_{\ell - 1}$ are defined the same as in \autoref{eq:backprop-delta-jacobian}. Similar to \autoref{eq:backprop-simple} and one of the examples in \autoref{sec:associaitve-memory}, {assuming $\alpha_{t+1} =1$}, the momentum term can be viewed as the result of optimizing the following objective with gradient descent:
\begin{align}\label{eq:momentum-map1}
    \min_{\boldsymbol{m}}\:\: \inner{\boldsymbol{m} \: \hat{\boldsymbol{x}}_{\ell - 1}}{\boldsymbol{\delta}_{\ell}}. 
\end{align}
The case of $\alpha_{t+1} \neq 1$ is equivalent to GD on the above minimization plus an $\ell_2$-regularization on the momentum term. Thus, momentum can indeed be viewed as an associative memory module that learns how to compress the past gradients of the objective into its parameters. Contrary to \autoref{eq:backprop-simple}, which was a simple 1-level associative memory and the update was directly applied to the memory, here the state of the momentum determines the update for the weights. In other words, it is a 2-level optimization procedure, in which the inner-loop learns the momentum and the outer-loop uses the state of the momentum to update the weights.

From this perspective, we can generalize the definition of momentum from EMAs to any arbitrary associative memory module that aims to compress the past gradients or maps the input of each token to its corresponding local error. This generalized momentum can be expressed as:
\begin{align}
    &{W_{\ell}}_{_{t+1}} = {W_{\ell}}_{_{t}} + {\boldsymbol{m}_{\ell}}_{_{t+1}}, \\
\end{align}
where $\boldsymbol{m}_{\ell}$ is the solution of the following associative memory, optimized by gradient descent:
\begin{align}
    \min_{\boldsymbol{m}}\:\: \tilde{\mathcal{L}}\left(\boldsymbol{m};  \:\hat{\boldsymbol{x}}_{\ell - 1}, - \boldsymbol{\delta}_{\ell} \right).
\end{align}
Here, the objective $\tilde{\mathcal{L}}(\cdot)$ is different from the original objective of the problem at hand, and $\tilde{\mathcal{L}}(\cdot)$ is the objective that defines the momentum and measures the quality of its mappings. In fact, the momentum term in this formulation aims to adapt in-context (recall that the context of the momentum is the gradients) to the local error rates based on the input of the layer. Most popular optimizers are formulated as element-wise update rule (for computational efficiency reasons) and so in \autoref{app:adam}, we first explore the element-wise associative memory formulation of momentum and connect it to popular optimizers such as Adam~\citep{kingma2014adam}. Showing that Adam can be viewed as the optimal associative memory to the $L_2$-regression objective that aims to predict the variance of gradients, we discuss other similar algorithms such as RMSProp~\citep{hinton2012neural}, SignSGD and its momentum-based variants~\citep{bernstein2018signsgd}, NAdam~\citep{dozat2016incorporating}, AMSGrad~\citep{reddi2016stochastic}, RAdam~\citep{Liu2020On}, and Lion~\citep{chen2023symbolic} are also instances of an associative memory that aims to compress the gradients. We then go beyond element-wise formulation and show that AdaGrad~\citep{duchi2011adaptive} is also an associative memory module. Due to the connection of AdaGrad with optimizers such as Shampoo~\citep{gupta2018shampoo} and Soap~\citep{vyas2025soap}–i.e., as the approximation of the preconditioning term–we then conclude that all these optimizers can be re-formulated as associative memory. Next, we discuss another class of optimizers based on preconditioning and reformulate them from NL's perspective in more details:

\head{Preconditioning and Approximation of Hessian} 
Another class of algorithms is preconditioning algorithms where the idea is to approximate Hessian inverse to mimic the behavior of Newton's algorithm. 
Formally, gradient descent~with~preconditioning~is~defined~as:
\begin{align}\label{eq:preconditioner-main}
    {W_{\ell}}_{_{t+1}} = {W_{\ell}}_{_{t}} - \eta_{t+1} \: \boldsymbol{P}^{\:-1}_{t+1} \:{\boldsymbol{g}_{\ell_{t+1}}},
\end{align}
where \emph{preconditioner} $\boldsymbol{P}_{t+1}$ is often a positive-definite matrix. A critical interpretation of preconditioner is their role in performing gradient descent in a  transformed coordinate system, which can be viewed as a mapping from gradients to that system of interest. Accordingly, we reformulate and interpret the preconditioner in \autoref{eq:preconditioner-main} as an associative memory that maps the set of gradients (or a function of gradients denoted as $\boldsymbol{g}$) to the system of our choice, denoted as $\hat{\boldsymbol{g}}$:
\begin{align}\label{eq:preconditioner-memory}
    {W_{\ell}}_{_{t+1}} = {W_{\ell}}_{_{t}} - \eta_{t+1} \: \boldsymbol{P}_{t+1}^{\:-1} \!\left({\boldsymbol{g}_{\ell_{t+1}}}\right),
\end{align}
where internally (in a nested level), $\boldsymbol{P}_{t+1}$ learns how to perform this mapping using an objective:
\begin{align}\label{eq:preconditioner-optimization}
\min_{\boldsymbol{P}} \quad   \tilde{\mathcal{L}} \left( \boldsymbol{P}\left( \hat{\boldsymbol{g}} \right); \boldsymbol{g} \right).
\end{align}
Given this viewpoint, the main question is about finding the best coordinate system that can empower the compression process. The most simple variant is an identity mapping, where we preserve the metric system and use $\boldsymbol{P}$ to map $\boldsymbol{g}$ (i.e., gradients in this case) to itself, resulting in preconditioning terms in Adam~\citep{kingma2014adam} and AdaGrad~\citep{duchi2011adaptive}, as discussed in \autoref{app:adam}. These results, along with the representation of Adam and its variants as associative memories, show that not only momentum-based optimizers are associative memories, but they also can be decomposed into a set of nested learning problems, each of which optimized with gradient descent. In a more general form, however, one can use more nested levels and optimize the inner problems in \autoref{eq:preconditioner-optimization} with gradient descent, resulting in:
\begin{align}
    \boldsymbol{P}_{t+1} = \boldsymbol{P}_{t+1} - \zeta_{t+1} \nabla_{\boldsymbol{P}_{t}} \tilde{\mathcal{L}} \left(\boldsymbol{P}_{t}; {\boldsymbol{g}}_{t+1} , \hat{\boldsymbol{g}}_{t+1}  \right).
\end{align}
In the NL framework, to design an effective preconditioning, one needs to find the right choice of  $\hat{\boldsymbol{g}}$ and  $\tilde{\mathcal{L}}.$
This viewpoint  can also lead to other classes of algorithms with gradient/momentum orthogonalization: e.g., Muon and its variants~\citep{jordanmuon, cesista2025spectralclipping, ManifoldMuon2025Keigwin}. Recalling Muon optimizer~\citep{jordanmuon}:
\begin{align}\nonumber
    &{W_{\ell}}_{_{t+1}} = {W_{\ell}}_{_{t}} + \texttt{NewtonSchulz}_k\left({\boldsymbol{m}_{\ell}}_{_{t+1}}\right)\\ \label{eq:muon1}
    &{\boldsymbol{m}_{\ell}}_{_{t+1}} = \alpha_{\ell, t+1} {\boldsymbol{m}_{\ell}}_{_{t}} - \eta_{\ell, t+1} \nabla_{{W_{\ell}}_{_{t}}} \mathcal{L}\left({W_{\ell}}_{_{t}}; \boldsymbol{x}_{t+1}\right),
\end{align}
where $\texttt{NewtonSchulz}_k(\cdot)$ performs $k$ steps of Newton-Schulz orthogonalization process. From the above discussion about the general formulation of preconditioning, one can see $\texttt{NewtonSchulz}_k(\cdot)$ operator as a mapping from gradients of momentum term to a proper metric system. The choice of proper coordinate system in Muon is to orthogonalize the gradients and so we aim to find a mapping $\boldsymbol{P}\left(\cdot \right)$ by minimizing a loss function $\min_{\boldsymbol{P}}\tilde{\mathcal{L}} (\boldsymbol{P}; \boldsymbol{O}, \vm)$
where objective $\tilde{\mathcal{L}}(\cdot; \cdot, \cdot)$ measures the quality of mapping from $\vo$ to either $\vm$ or $\vg$ by $\vp(\cdot)$. A critical challenge in this process is that the parameter $\vo$ itself is not given and so the mapping requires learning both the mapping and the proper orthogonal space. A simple formulation measuring orthogonalization, can be achieved by defining the objective as:
\begin{align}\label{eq:muon-motivation}
    \tilde{\mathcal{L}}(\vp(\vg); \vg) =  \: \| \vp(\vg)^{\top} \vp(\vg) - \boldsymbol{I} \|^2_{\boldsymbol{F}}, 
\end{align}
where $\vp(\vg)$ is the orthogonal space that we aim to directly learn from gradients. This objective ensures that the gradients (or momentum) and their mapping are relatively close while the mapping is to an orthogonal space.   Optimizing the above objective to find $\vo = \vp(\vg)$ with one step of gradient descent results in: 
\begin{align}\label{eq:muon2}
    \vo_{i+1} = \vo_{i} - \zeta_{i+1} \nabla_{\vo_{i}} \:\tilde{\mathcal{L}}\left(\vo_{}i; \vg_{t}\right) = \vo_{i} - \zeta_{i+1} \left( \vo_{i} - \vg_{t} +  2 \vo_i \left(\vo_i^{\top} \vo_i - \boldsymbol{I}\right) \right),
\end{align}
which recovers the 3-degree polynomial (initial value $\vo_{0} = \vg_t$). In a summary, the higher-frequency level learns the orthogonal mapping and then the lower-frequency process use the learned mapping to optimize the weights.  Later, in \autoref{sec:more-expressive-momentum}, we discuss a more general viewpoint that considers $\texttt{NewtonSchulz}_k\left(\cdot \right)$ as a polynomial mapping to enhance the capacity of the memory.

\subsection{Long Context in Optimizers: An Example of Continual Learning with~Orthogonal~Tasks} \label{sec:Optimizers-continual-learning}
When removing the boundary between train and test time, and  moving towards models that can continually learn for a long period of time, the role of (online) optimizers becomes more prominent: mainly due to the need for finding effective ``solutions" rather than converging faster. Finding an effective solution, however, requires global understanding of the objective to avoid ``local minima" as well as moving toward directions that might cause (catastrophic) forgetting of long past learned tasks. From associative memory perspective and as discussed earlier, the momentum term is expected to be a memory of past gradients, helping the optimization process to have a more global view of the loss landscape. The current design of momentum, however, acts as a simple low-pass filter that smoothifies the gradient updates and thus has limited capacity with only incorporating information from recent past.

To better illustrate this limitation, let $\beta > 0$ be the momentum's decay term, and so the contribution of $i$-th gradient before to the current state of the momentum can be calculated as $\beta^{i} \left(1 - \beta\right)$. Considering the cumulative sum of gradients' contributions to the current state of the momentum term (i.e., $\boldsymbol{S}_t = \sum_{i = 0}^{t}  \beta^{i} (1-\beta)$) and the commonly used value of $\beta = 0.9$ in optimization setups, the last 6 gradients (resp. 43 gradients) are responsible for at least 50\% (resp. 99\%) of the cumulative contribution, i.e., $\boldsymbol{S}_t$. This indicates that gradients and generally the global information beyond only past 43 steps contribute less than 1\%, limiting the understanding of the  objective landscape and the ability to find effective solutions. This simple example shows that the current design is limited even in incorporating the long past information, let alone its ability to properly retrieve information needed~for~the~current~state~of~the~momentum.  

Coming back to the setup of continual learning and considering one of its simple variants with orthogonal tasks. We let $\{(\mathcal{T}_i, \mathcal{D}_i, \mathcal{L}_{i})\}_{i = 1}^{n}$ be the set of tasks, their corresponding data, and their objectives such that for task $\mathcal{T}_i$:
\begin{align}
    \mathcal{L}_i(W) = \mathbb{E}_{(\vx, \vy)\thicksim \mathcal{D}_{i}}\left[ (W^{\top} \vx - \vy)^2 \right],
\end{align}
and gradients live in orthogonal directions of $\{\vu_i\}_{i = 1}^{n}$. Given the task $\mathcal{T}_t$ (for large enough $t > 1$), when optimizing the problem with gradient descent with momentum, and after many steps on task $t$, the gradients now point along $\vu_t$. Accordingly, the momentum term is gradually shifted and now is approximately in $\vu_t$ direction. This can lead to gradual forgetting about past gradients, which in turn may cause catastrophic forgetting in the model. That is, the optimization process can move the weights in a direction that damages the performance on previous tasks, mainly because the optimizer has no memory of the old gradient subspace that it should avoid. This failure is not about the capacity of the model but the memory management of the optimization process, failing at finding an effective solution. 

\begin{myboxi}[Long Context Understanding in Optimizers]
    \textit{Moving from static models to neural learning modules that can continually learn from data/experience, the optimization process itself can benefit from long-term compression/understanding of gradient subspace to find effective solutions over diverse sets of tasks and for a long period of time.}.
\end{myboxi}

Motivated by  this observation, we next discuss more expressive variants of momentum that are capable of better memory management and higher memory capacity:

\subsection{More Expressive Designs for Momentum as an Associative Memory}\label{sec:more-expressive-momentum}
So far we discussed that (1) momentum term can be viewed as an associative memory that aims to compress the (past) gradients into its parameters; and (2) for developing models that can continually learn for a long period of time and on diverse sets of tasks, the optimization process need proper information about long past and the global properties of loss landscape. Next, we discuss how  Nested Learning and associative memory viewpoint could result in designing optimizers with diverse memory management/structure:

\head{Extension: More Expressive Association}
As discussed earlier, vanilla momentum term can be viewed as a value-less associative memory. 
To allow more expressive associative memory  and following the original definition of associative memory (i.e., mapping keys to values), we {let value parameter $\vv_i = \vp_i$} and so the momentum aims to minimize:
\begin{align}
    \min_{\vm}\;\;\inner{\vm \: \nabla \mathcal{L}(W_i;\boldsymbol{x}_i)^{\top}}{\vp_i},
\end{align}
or equivalently, minimizes $\langle \vm, \vp_i  \nabla \mathcal{L}(W_i;\boldsymbol{x}_i)\rangle$.  Using gradient descent for updating the momentum
results in the update rule of:
\begin{align}\nonumber
    &W_{i+1} = W_i + \mathbf{m}_{i+1}\\ \label{eq:momentum-p}
    &\mathbf{m}_{i+1} =   \mathbf{m}_{i} - \eta_i \vp_i  \nabla \mathcal{L}\left(W_{i};\boldsymbol{x}_i\right),
\end{align}
where we also included a forget gate $\alpha_{i+1}$ for fogetting the past.
This update rule can be viewed as preconditioning the momentum GD. Here,  the momentum term with preconditioning is interpreted as an associative memory that learns how to compress the mappings between $\vp_i$ and the gradient term $\nabla \mathcal{L}(W_i;\boldsymbol{x}_i)$. 
While any reasonable choice (e.g., random features) of preconditioning may also be used to improve the expressivity of the initial version of GD with momentum. 
Note that our discussion on preconditioning in \autoref{eq:preconditioner-memory} is different from here and indicates that one can learn the preconditioner as a proper mapping, while the above formulation shows when using preconditioning, momentum term (as a memory for gradients) aims to map~them~to~their~mapping~function. 

\head{Extension: More Expressive Objectives}
Revisiting the formulation of momentum: for a given gradients $\nabla \mathcal{L}\left(W_{i};\boldsymbol{x}_i\right)$, the momentum mapping is based on dot-product similarity as the internal objective (as discussed above) and so its update rule is a Hebbian-rule~\citep{hebb2005organization}. From an associative memory perspective, this update rule has a limited capacity
\citep{storkey1997increasing} and makes the update rule of momentum independent of its current state. Thus it limits its ability to track/compress the  loss landscape information.
A natural extension is to replace the internal objective with $L_2$-regression loss (for measuring the corresponding key-value mapping fitness) and minimize the loss function~$\|\vm  \nabla \mathcal{L}(W_i;\boldsymbol{x}_i)^{\top} - \vp_i\|^2_2$, resulting in the update rule of:
\begin{align}
    &W_{i+1} = W_i + \mathbf{m}_{i+1},\\ \label{eq:momentum-delta}
    &\mathbf{m}_{i+1} =\mathbf{m}_{i} \;  \left(\alpha_{i+1} -  \nabla \mathcal{L}\left(W_{i};\boldsymbol{x}_i\right)^{\top} \nabla \mathcal{L}\left(W_{i};\boldsymbol{x}_i\right) \right)  - \eta_t \vp_i  \nabla \mathcal{L}\left(W_{i};\boldsymbol{x}_i\right), 
\end{align}
This update is based on delta-rule~\citep{prados1989neural} and so it allows the memory (momentum) to better manage its limited capacity (i.e., $\mathcal{O}(N)$) and better memorize the series of past gradients. 
For example, we may learn to forget some of the past gradients during the optimization process (similar to what happens when we move from linear attention to delta rule in associate memory).
We refer to the variants of this type of momentum term as Delta Momentum variants.

\head{Extension: More Expressive Memory}
{Viewing momentum as a compressor or a memory that store past gradients into its elements (parameters), its capacity not only depends on its update rule (similar to the above), but it also requires more expressive structure that allows for larger capacity. The current formulation is based on a linear layer (i.e., matrix-valued) to compress the past gradient values, but this linear nature may limit the capability to only learn linear mappings of past gradients. To increase the learning capacity of this module, one can use more complex mappings  such as replacing a linear matrix-valued memory for momentum with an MLP. This design allows momentum to memorize more gradients and so provides better information for the optimization process. We extend the formulation in \autoref{eq:momentum1} as:
\begin{equation}\label{eq:deep-momentum}
    W_{i+1} = W_i + \mathbf{m}_{i+1}\left(\mb{u}_i\right), \quad \textrm{and}\quad 
    \mathbf{m}_{i+1} = \alpha_{i+1} \mathbf{m}_{i} - \eta_t \nabla \mathcal{L}^{(2)}(\vm_i; \mb{u}_i, \mathds{1}), 
\end{equation}
where $\mb{u}_i = \nabla \mathcal{L}\left(W_{i};\boldsymbol{x}_i\right)$ and $\nabla \mathcal{L}^{(2)}(\cdot)$ is the internal objective of momentum (e.g., dot product similarity $\inner{\vm(\mb{u}_i^{\top})}{\mathds{1}}$). We refer to this variant as Deep Momentum Gradient Descent {(DMGD)}. While it is clear from this example, it is worth emphasizing that the internal loss function and the model need to be designed carefully to obtain an effective momentum module.

\head{Extension: Memory with Higher-order Feature Maps} One of the commonly used techniques to enhance the capacity of a memory is to use higher-order feature maps on the keys~\citep{polysketchformer-kacham2024, katharopoulos2020transformers}. Using this technique on the momentum term, one can obtain: 
\begin{equation}
    W_{i+1} = W_i + \mathbf{m}_{i+1}\quad \textrm{and}\quad
    \mathbf{m}_{i+1} = \alpha_{i+1} \mathbf{m}_{i} - \eta_t \vp_i  \phi(\nabla \mathcal{L}\left(W_{i};\boldsymbol{x}_i\right)), 
\end{equation}
where $\phi(\cdot)$ is a higher-order feature mapping (that may be learned through its internal objective).

\head{Extension: Nonlinear Outputs}
Building upon the associative memory perspective of the momentum, one common technique to enhance the representation power of  memory module is to use non-linearity on top of its output~\citep{behrouz2024titans, sun2024learning}. That is, we re-formulate \autoref{eq:deep-momentum} as:
\begin{equation}\label{eq:deep-momentum2}
    W_{i+1} = W_i + \sigma\left(\mathbf{m}_{i+1}\left(\mb{u}_i\right)\right), \quad \textrm{and}\quad 
    \mathbf{m}_{i+1} = \alpha_{i+1} \mathbf{m}_{i} - \eta_t \nabla \mathcal{L}^{(2)}(\vm_i; \mb{u}_i, \mathbf{I}),
\end{equation}
where $\sigma(\cdot)$ is an arbitrary non-linearity. As an example, we let $\sigma(\cdot) = \texttt{NewtonSchulz}(\cdot)$, where $\texttt{Newton-Schulz}(\cdot)$ is the iterative  Newton-Schulz method~\citep{higham2008functions}, and $\vm(\cdot)$ be a linear layer; resulting in Muon~\citep{jordanmuon}.

\begin{wrapfigure}{r}{0.33\textwidth}
    \centering
    \vspace{-2ex}
    \includegraphics[width=0.85\linewidth]{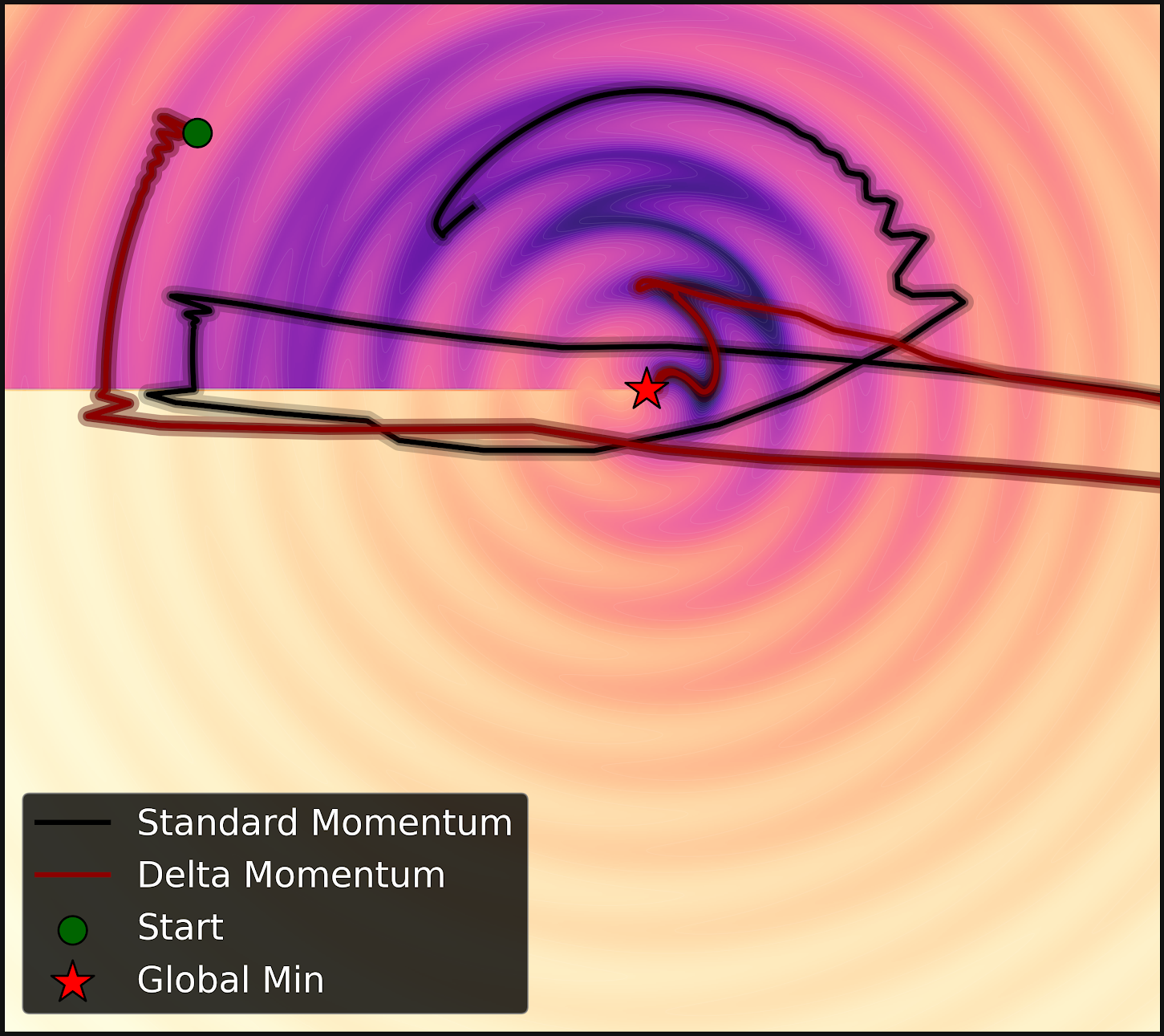}
    \vspace{-4pt}
    \caption{Optimization of function $\boldsymbol{\psi}(r, \theta)$ with standard momentum and our delta momentum. }
    \label{fig:optimizer-state}
    \vspace{-1ex}
\end{wrapfigure}

\head{A Toy Example for Long Context in Optimizer}
In \autoref{sec:Optimizers-continual-learning}, we discussed that in complex setups, including continual learning with orthogonal tasks, we may need more complex momentum terms, either with higher capacity or better memory management. To better illustrate the potential gains of other momentum memory designs, we use a toy example of a time-varying curvature. Since the standard momentum acts as a low-pass filter, if the  landscape changes at a high frequency, then standard momentum which aims to use the weighted average of past gradients, will be under the influence of irrelevant gradient terms, delaying the convergence. As an illustrative example, consider: 
\begin{align}
    \boldsymbol{\psi}(r, \theta) = r^2 + k \times \left( r - \theta + \alpha \sin{\left(\omega r\right)} \right)^{2},
\end{align}
and aims to optimize it using a standard momentum and our delta momentum. We start the optimization process from point $(r_0, \theta_0) = (-3.5, 2)$ and continue until one of the algorithms converge to the optimal solution. The result is visualized in \autoref{fig:optimizer-state}. The delta momentum finds the solution faster, mainly due to its gradient-dependent weight decay that helps the momentum term to decay or stop when it is needed.

\vspace{2ex}
\subsection{Going Beyond Simple Gradient Descent and Momentum} \label{sec:L2-backprop}
Coming back to the discussion in \autoref{sec:associaitve-memory} about the pre-training process and backpropagation being a form of associative memory, in this section, we aim to take advantage of NL's viewpoint and present a more general form for gradient descent. As observed in \autoref{sec:backprop-memory-referential}, backpropagation with gradient descent is an associative memory that aims to map the input data to the surprised caused by its predicted output $\nabla_{y_t} \mathcal{L}(W_t;\boldsymbol{x}_t)$:
\begin{align}\label{eq:gradient-dual-2}
    W_{t+1} = W_{t} - \eta_{t+1} \nabla_{W} \mathcal{L}(W_t;\boldsymbol{x}_t) = W_{t} - \eta_{t+1} \nabla_{y} \mathcal{L}(W_t;\boldsymbol{x}_t) \otimes\boldsymbol{x}_t, \qquad \text{where}\:\:\boldsymbol{x}_t \thicksim \mathcal{D}_{\text{train}},
\end{align}
which from the associative memory perspective and proximal gradient viewpoint is equivalent to: 
\begin{align}
    W_{t+1} = \arg\min_{W} \:\: \inner{W\boldsymbol{x}_t}{\nabla_{y_t} \mathcal{L}(W_t;\boldsymbol{x}_t)} + \frac{1}{2\eta_t} \: \|W - W_{t}\|^{2}_2.
\end{align}
This step aims at learning the negative of the gradient direction. The main drawback of the dot-product similarity as the inner objective is that its corresponding update rule and so learning algorithm treats each data sample (gradients) independent of the state, meaning that the state of the weights and so previous gradients do not affect the update term to the current state. While this design can be effective for nested problems with independent elements in their context (e.g., i.i.d. samples for training), it can be  restrictive for context with highly dependent elements (e.g., tokens in a sequence).  Defining $\mathbf{u}_t = -\nabla_{y_t} \mathcal{L}(W_t;\boldsymbol{x}_t)$, one can extend this process to more expressive objectives such as $\mathcal{L}_2$ regression loss:
\begin{align}
    W_{t+1} = \arg\min_{W} \:\: \frac{1}{2}\|W\boldsymbol{x}_t - \mathbf{u}_t \|^{2}_2 +  \frac{1}{2\eta_t}\|W - W_{t}\|^{2}_2.
\end{align}
For the cases that $\vx_t$ is normalized (e.g.  in normalized memory systems or in neural networks with normalization layers, $\|\vx_t\|_2 = \lambda$), and by defining $\eta_t' = \frac{\eta_t}{1+\eta_t}$, we can use Sherman-Morrison lemma to get (see \autoref{app:DGD} for the details):
\begin{align} \nonumber
    W_{t+1} &= W_{t} \left( \mathbf{I} -\eta'_t\boldsymbol{x}_t\boldsymbol{x}_t^{\top}\right) - \eta_{t}' \nabla_{W_t} \mathcal{L}(W_t;\boldsymbol{x}_t) \\ \label{eq:DGD}
    &= W_{t} \left( \mathbf{I} -\eta'_t\boldsymbol{x}_t \boldsymbol{x}_t^{\top}\right) - \eta_{t}' \nabla_{y_t} \mathcal{L}(W_t; \boldsymbol{x}_t) \otimes \boldsymbol{x}_t, \qquad \text{where}\:\: \boldsymbol{x}_t \thicksim \mathcal{D}_{\text{train}}.
\end{align}
This new algorithm, which based on Delta rule~\citep{prados1989neural} we refer to as Delta Gradient Descent (DGD), updates the weights not only with respect to the current elements, but it also incorporates the previous state of weights, resulting in an adaptive decay term based on the current data sample. Next, we discuss a generalized viewpoint about the process of backpropagation with gradient descent, which later will help us to formulate Generalized Gradient Descent family of learning rules:

\vspace{8pt}
 \begin{myboxi}[Training a Deep Neural Network with Backpropagation is a Self-Referential Process:]
     \textit{As discussed earlier in \autoref{sec:backprop-memory-referential}, one common misinterpretation for gradient descent is to view it as a form of linear recurrence (e.g., linear attention). In a conventional linear recurrence, however, keys and values are independent of the state of the memory, and so allows for parallelization of the formulation. The values in ``gradient descent as associative memory'' viewpoint are a function of the state of the memory, and so it is a self-referential model~\citep{schmidhuber1993self} that controls its own learning process by generating its own values. More formally, one can reformulate the process as: }
\begin{align}
    &W_{t+1} =  W_{t} + \eta_{t+1} \vv_t \otimes \boldsymbol{x}_t,\\ \nonumber
    &\vv_t = \boldsymbol{f}_{W_t}\left(\vx_t \right) = - \nabla_{y_t} \mathcal{L}(W_t;\boldsymbol{x}_t),
\end{align}
\textit{which means that at each step $\vv_t$ is generated by memory $W_t$ and $\vx_t$ as its input.}
 \end{myboxi}

Based on the above interpretation, one can define backpropagation with gradient descent in a general form as \emph{any} self-referential model that aims to compress training samples as keys and map them to \emph{self-generated} values to better control its own learning process. Based on this definition, our above formulation of \autoref{app:DGD}, is only a simple instance that uses $\mathcal{L}_2$ regression loss; in general, however, one can define Generalized Gradient Descent (GGD) as: 

\begin{dfn}[Generalized Gradient Descent (GGD) Learning Rule]
Generalized Gradient Descent (GGD) learning rule is a self-referential associative memory that aims to compress data samples and map them to a set of self-generated keys:
    \begin{align}
        W_{t+1} = \arg\min_{W} \:\: \tilde{\mathcal{L}} \left(\vx_t, \vu_t \right) +  \texttt{Ret}\left(W, \{W_i\}_{i = t - c + 1}^{t} \right),
    \end{align}
    where $\vu_t$ is a {self-generated} value:
    \begin{align}
        \vu_t = \boldsymbol{f}_{W_t}\left( \vx_t \right),
    \end{align}
    for some function $\boldsymbol{f}_{W_t}(\cdot)$ parameterized by $W_t$. Here,  $\tilde{\mathcal{L}} \left(\cdot\right)$ measures the quality of the mapping, and $\:\texttt{Ret}(\cdot)$ ensures that the solution for the new instance is not far away from the current state. 
\end{dfn}
Similarly, this formulation can be adapted for the momentum term, resulting in Generalized Momentum (GM). However, it is notable that the momentum itself, is a conventional associative memory and its keys and values are given, or more specifically are generated by a lower-frequency level. In \autoref{sec:momentum-based}, we explored a special case of this formulation, where $\mathcal{L}(\cdot)$ is $L_2$ regression loss.

\head{A Note on Optimizers in Continual Learning Setup} 
As discussed above, optimizers themselves are learning modules or associative memories that aim to compress the gradients into their parameters. These parameters are not necessarily trainable in the conventional terminology but indeed momentum-based optimizers store the knowledge about the loss landscape, helping them to better update the weights. When the ``end of pretraining'' happens for a neural learning module, the knowledge stored about the distribution of gradients/data, which are stored in the momentum term(s) is removed from the model and so continuing the training without recovering the momentum states can affect the model's ability to learn new capabilities. When the model is in continual learning setup, the knowledge about data is stored in the conventional parameters (optimized with backpropagation), but the knowledge about how the model optimizes itself and about the objective space are optimized in the lower-frequency levels of optimization (e.g., momentum terms).

\section{Existing Architectures as Neural Learning Modules}\label{sec:arch}
Modern sequence models such as Transformers~\citep{transformers} and recurrent models~\citep{katharopoulos2020transformers, schlag2021linear, sun2024learning, behrouz2024titans} are the backbones of recent advances in language models. Recently, the equivalency of such models with associative memories that aim to learn a mapping from keys to values from data have been studied in different settings and objectives~\citep{liu2024longhorn, sun2024learning, wang2025test, behrouz2025Miras}. Particularly, we focus on the general framework of Miras~\citep{behrouz2025Miras}, which defines associative memory as Definition~\ref{def:associative-memory} and optimizes the internal objective (called ``attentional bias'') with a choice of optimization algorithm on an arbitrary class of functions (i.e., memory architecture). While this formulation alone indicates that the well-known architectures are instances of nested systems of associative memory (\nsam), next, we review this equivalency for some learning rules and architectures. 

From now on, we assume that keys $\{\vk_i\}_{i=1}^{L}$, values $\{\vv_i\}_{i=1}^{L}$, and queries $\{\vq_i\}_{i=1}^{L}$ are given: they often are defined as the projections of the input, i.e., 
\begin{align}
    \vk_t =\boldsymbol{x}_t \textcolor{mydarkred}{W_{\vk}} , \qquad \vv_t =\boldsymbol{x}_t \textcolor{mydarkred}{W_{\vv}} , \qquad \vq_t =\boldsymbol{x}_t \textcolor{mydarkred}{W_{\vq}}.
\end{align}
In this design, since projection parameters (i.e., $\textcolor{mydarkred}{W_{\vk}}$, $\textcolor{mydarkred}{W_{\vv}}$, and $\textcolor{mydarkred}{W_{\vq}}$) are optimized in a lower frequency level, the sequence model component (e.g., self-attention) has a higher frequency and so the learning process of the associative memory happens in a lower level. Accordingly, for the sake of clarity, we only discuss the higher frequency level (i.e., the internal learning process of the associative memory).

\head{Softmax Attention} From the associative memory viewpoint: given keys $\{\vk_i\}_{i=1}^{L}$, values $\{\vv_i\}_{i=1}^{L}$, and queries $\{\vq_i\}_{i=1}^{L}$, \texttt{Softmax} attention block~\citep{bahdanau2014neural, transformers} can be reformulated as a non-parametric  solution to the $\ell_2(\cdot)$ regression objective with  Nadaraya-Watson estimators~\citep{zhang2022analysis, fan2018local}:
\begin{align}\label{eq:attention-nadaraya}
    \M^* = \arg\min_{\M} \sum_{i = 1}^{L} \mathbf{s}(\vk_i, \vq) \|\vv_i - \M \|^2_2
    = \sum_{i = 1}^{L} \frac{\mathbf{s}(\vk_i, \vq)}{\sum_{j = 1}^{L} \mathbf{s}(\vk_j, \vq)} \vv_i,
\end{align}
where $L$ is the sequence length~\citep{sun2024learning}. This formulation optimizes the memory $\M(\cdot)$ with respect to the entire context; however, one design choice can be to limit the optimization process to the past $c$ tokens, resulting in:
\begin{align}
    \M^* = \arg\min_{\M} \sum_{i = t - c + 1}^{t} \mathbf{s}(\vk_i, \vq_i) \|\vv_i - \M \|^2_2  = \sum_{i = t-c + 1}^{t} \frac{\mathbf{s}(\vk_i, \vq)}{\sum_{j = t - c + 1}^{t} \mathbf{s}(\vk_j, \vq)} \vv_i, 
\end{align}
which is equivalent to the sliding window attention (SWA). Therefore, attention and its more expressive variants~\citep{wang2025test} also are instances of Definition~\ref{def:associative-memory}, when instead of gradient descent or other parametric methods, we find the optimal non-parametric solution to the mapping. 

\head{RNNs with Hebbian Rule} The first generation of modern recurrent architectures (e.g., Linear attention~\citep{katharopoulos2020transformers}, RetNet~\citep{sun2023retentive}, RWKV~\citep{peng2023rwkv}, lightening attention~\citep{li2025minimax}) are based on Hebbian-like learning rules~\citep{hebb2005organization}. For this class of models, the inner objective to measure the quality of mapping between keys and values is the dot-product similarity. That is, given a matrix-valued memory $\M \in \R^{d \times n}$, keys and values $\vk, \vv \in \R^{d}$, objective $\tilde{\mathcal{L}}(\M; \vk_t, \vv_t) := -2 \inner{\M \vk_t}{\vv_t} $, and a kernel $\phi(\cdot)$, we optimize the equivalent associative memory optimization problem (see Definition~\ref{def:associative-memory}) with gradient descent and weight decay, resulting in:
\begin{align}\label{eq:hebb-rule}
    \M_t = \alpha_t \M_{t-1} - \eta \: \undermath{-\vv_t \phi\left(\vk_t^{\top} \right)}{\nabla_{\M_{t-1}} \tilde{\mathcal{L}}\left( \M_{t-1}; \phi(\vk_t), \vv_t \right)} \ \: = \alpha_t \M_{t-1} + \eta_t \: \vv_t \phi \left(\vk_t^{\top} \right),
\end{align}
which recovers the original linear attention recurrence~\citep{katharopoulos2020transformers}. Given different settings for $\alpha_t$ (i.e., either is $1$, learnable, channel-wise, and/or input-dependent) and also $\phi(\cdot)$ (i.e., identity, polynomial kernels, etc.), the above recurrence recovers different variants of linear attention with Hebbian rule~\citep{katharopoulos2020transformers, polysketchformer-kacham2024, arora2024simple, sun2023retentive, peng2024eagle, beck2024xlstm}. Therefore, the variants of linear attention with Hebbian rule can be reformulated as the process of an optimization problem, in which the memory aims to learn the mapping between keys and values based on dot-product similarity objective, with gradient descent.

\head{RNNs with Delta Rule} To improve the memory management and to enhance the memory capacity of the above group, several studies suggest replacing Hebbian rule with Delta rule as the learning algorithm in recurrent neural networks~\citep{schlag2021linear}, resulting in models such as DeltaNet~\citep{schlag2021linear}, Longhorn~\citep{liu2024longhorn}, and RWKV7~\citep{peng2025rwkv7}. When letting $\M \in \R^{d \times n}$, delta rule is equivalent to optimizing MSE objective $\tilde{\mathcal{L}}_t = \|\M_t \vk_t - \vv_t\|^2_2$ with $\text{Ret}_t(\M, \M_{t-1}) = \| \M_t - \M_{t-1} \|_F^2$ as local retention, and stochastic gradient descent as the optimizer:
\begin{align}
    \M_t = \M_{t-1} - \eta_t \:  \undermath{\left(\M_{t-1}\vk_t - \vv_t\right) \vk_t^{\top}}{\nabla_{\M_{t-1}} \: \tilde{\mathcal{L}}\left( \M_{t-1}; \phi\left(\vk_t\right), \vv_t \right)} = \left( \mathbf{I} - \eta_t \vk_t \vk_t^{\top}\right) \M_{t-1} + \eta_t \:\vv_t \vk_t^{\top}.
\end{align}
Using other forms of retention gates (e.g., $\text{Ret}_t(\M, \M_{t-1}) = \| \M_t - \alpha_t \:\M_{t-1} \|_F^2$), optimization algorithms with weight decay (e.g., regularizing with $\|\M_{t}\|_q^q$ for a given $q > 0$), multiple steps of gradient descent, and/or different formulations of learnable parameters such as $\eta_t$ and $\alpha_t$ can result in diverse variants of delta rule~\citep{peng2025rwkv7, sun2024learning, hu2025improving, wang2025test, siems2025deltaproduct, irie2021going, liu2024longhorn, behrouz2025Miras}. Therefore, Delta rule and its variants are all instances of an optimization problem, in which the model aims to learn a mapping between keys and values based on the $L_2$-regression objective.

\head{Beyond Conventional Learning Rules: Omega, Oja’s, and Non-Euclidean Learning Rules}
More recently, there have been growing interests in designing architectures from the associative memory perspective (see Definition~\ref{def:associative-memory}) and use more complex internal objectives, and/or optimization algorithms, resulting in learning algorithms beyond Delta and Hebbian rules~\citep{behrouz2025Miras, irie2022neural, behrouz2025atlas, von2023uncovering, zhang2025higher}. More specifically, to enhance the stability of Hebbian rule (discussed in \autoref{eq:hebb-rule}), \citet{irie2022neural} introduced OjaNet based on Oja's rule~\citep{oja1982simplified} with the following recurrence:
\begin{align}
    \M_t = \alpha_t \M_{t-1} + \eta_t \: \vv_t  \left(\phi(\vk_t)^{\top}  - \M_{t-1}^{\top} \vv_t \right).
\end{align}
In the associative memory formulation (as in Definition~\ref{def:associative-memory}), this recurrence can simply be reformulated as one step of gradient descent as:
\begin{align}
    \M_t = \M_{t-1} - \eta_t \: \undermath{\M_{t-1}^{\top} \vv_t - \vv_t \phi(\vk_t)^{\top}}{\nabla_{\M_{t-1}} \: \tilde{\mathcal{L}}\left( \M_{t-1}; \phi\left(\vk_t\right), \vv_t \right)},
\end{align}
where $\tilde{\mathcal{L}}(\M; \vk_t, \vv_t) = -2 \inner{\M \vk_t}{\vv_t} + \|\M^{\top} \: \vv_t\|^2_2$ and $\phi(\cdot)$ is a kernel~\citep{irie2022neural, irie2025fast}. Although this design enhances the Hebbian learning rule by enforcing a unit-norm constraint for the single-neuron, it has been reported to empirically underperform models based on Delta learning rule~\citep{irie2022neural}. To further enhance the Delta rule through the design of more expressive objectives, recently, \citet{behrouz2025Miras} suggested going beyond Euclidean spaces and use $L_p = \|\cdot\|_p^{p}$ norm for the internal regression objective, showing better empirical performance and robustness in long context tasks compared to Delta rule and its variants.

While the majority of learning rules are online update mechanisms–meaning that at each state, the models only need to keep the memory and the \emph{current} (batch of) input–Omega rule~\citep{behrouz2025atlas} suggest an update rule based on a set of past (batches of) inputs (or all inputs). More specifically, given a memory $\M$ with an arbitrary structure, keys and values $\vk, \vv \in \R^{d}$, an arbitrary objective $\tilde{\mathcal{L}}(\M; \vk_t, \vv_t)$, and a kernel $\phi(\cdot)$, Omega rule is defined as:
\begin{align}
    \M_{t} = \alpha_t \: \M_{t-1} - \sum_{i = t-c+1}^{t} \gamma_{t, i} \:\:\tilde{\mathcal{L}}(\M_t; \phi(\vk_i), \vv_i),
\end{align}
where $c \geq 1$ is the local window of cached inputs. Note that in the special case of $\gamma_{t, i} = 1$ and $c$ equal to the entire context length, the optimal solution of the above design collapses into an online case, where the update rule only depends on the current state and the current input~\citep{von2023uncovering}. For further discussion with more details about representing architectures as associative memories and so an optimization problem, we refer the reader to \citet{behrouz2025Miras}.

\head{A Note on Gating in Modern Sequence Models}
One of the recent architectural changes in modern language models is the gating of a linear layer's output with the output of the sequence model. Despite significant improvement resulted by this method, it is still unclear that how it enhances the performance. 
As we discussed in \autoref{fig:mlp-vs-lin-attention} and its corresponding example, the main difference between feedforward network and modern recurrent memory modules (e.g., linear attention~\citep{katharopoulos2020transformers} or deep memory modules~\citep{behrouz2024titans}) when their initial state of the memory is meta-learned, is the second level in memory modules that perform in-context learning and adapt its state with the context. From this viewpoint, when the initial value of the memory is not meta-learned, it only relies on the in-context adaption of the memory and so there is no persistent memory system that stores the knowledge of pre-training in this block. Therefore, when the initial value of memory is not meta-learned, which is common in earlier variants of linear transformers, the gating of linear attention acts as a persistent memory and the initialization of the memory module.

\subsection{Revisiting the Human Brain Perspective of Nested Learning}\label{sec:revisit-brain-perspective}
In \autoref{sec:brain-perspective}, we discussed how structure in human brain is uniform and reusable, and if we need a new architecture in deep learning, or if our beliefs about the heterogeneity of current models need to be revisited. In the previous sections, we observed that both optimization process of neural networks as well as neural architectures can be formulated as a set of nested and/or parallel optimization problems, in which the memory structure is a feedforward layer (e.g., either Deep MLPs, linear layers, etc.) and the objective is optimized with gradient descent or Newton's methods.  

From this perspective, modern architectures, are a set of artificial neurons (i.e., linear or deep feedforward networks), and each group of neurons has their own internal objective and so update mechanism. To this end, as a simple example, let us recall the AdaTransformer in \autoref{fig:mlp-vs-lin-attention}: Given $X = \{\vx_i\}_{i = 1}^{T}$ as the input sequence, the output of the block is computed as (For the sake of simplicity, we assume $\texttt{MLP}(\cdot) = \cdot \: W_{\texttt{MLP}} \:$, and remove normalizations): 
\begin{align} \nonumber
    &\vk_t =\boldsymbol{x}_t \textcolor{mydarkred}{W_{\vk}} , \qquad \vv_t =\boldsymbol{x}_t \textcolor{mydarkred}{W_{\vv}} , \qquad \vq_t =\boldsymbol{x}_t \textcolor{mydarkred}{W_{\vq}}, \\ \nonumber
    &\boldsymbol{y}_{\texttt{attn}_{, t}} = \textcolor{c4}{\texttt{Attn}}\left(\vk_t, \vv_t, \vq_t \right),\\ \nonumber
    &\boldsymbol{y}_{\texttt{block}_{, t}} = \boldsymbol{y}_{\texttt{attn}_{, t}} \: \textcolor{c4}{W_{\texttt{LinAttn}_{, t}}}, \:\:\:\quad\qquad \text{where} \\
    & \textcolor{c4}{W_{\texttt{LinAttn}_{, t}}} = \textcolor{c4}{W_{\texttt{LinAttn}_{, t-1}}} + \textcolor{c4}{\vv_t\vk_{t}^{\top}},
\end{align}
The lowest frequency level is responsible for the optimization of  $\textcolor{mydarkred}{W_{\vk}}$, $\textcolor{mydarkred}{W_{\vv}}$, and $\textcolor{mydarkred}{W_{\vq}}$, all of which are feedforward networks and so are uniform. The attention itself is also the non-parametric matrix-valued solution of a regression objective, again verifying that the structure is a matrix of artificial neurons (i.e., parameters). Finally, the Linear Attention++ component is equivalent to the optimizing the dot-product similarity of the mappings over the linear class of functions. Therefore, all the parameters are matrix-valued or deep feedforward layers, meaning that the only difference in the components of an architecture is their level, objective, and/or learning update rule.

~
~
~
\begin{myboxi}[Modern Deep Learning Models Have Uniform and Reusable Structure]
    Neural learning modules consist of a set of feedforward networks, each of which are optimized in different levels and time scales. The heterogeneity we observe in deep learning architectures, however, is due to the lack of view to this new NL's axis, resulting in observing only the solution of optimization problems and so causing the illusion of deep learning architectures.  
\end{myboxi}

\section{Takeaways and Revisiting Common Terms}\label{sec:revisit}
In the previous sections, we discussed the concept of nested learning and how existing well-known components of neural networks such as popular optimizers and architectures fall under the NL paradigm. In this section, we discuss the takeaways, connection of different concepts, and the implications of NL perspective on common terms. 

\headred{Memory and Learning} For a long period of time, in machine learning models, memory have been treated as a separate block with a \emph{clear} distinction between its parameters and other components. Such designs often assume a short and/or long-term memory blocks, where short-term memory is responsible for the local context, while long-term memory is the storage for the persistent knowledge in models. In human brain, however, memory is considered as a distributed interconnected system without a clear known components that are independently responsible for short or long-term memory. In NL, we build upon a common terminology for memory and learning in neuropsychology literature, indicating that: Memory is a neural update caused by an input and so learning is the process of acquiring useful memory~\citep{okano2000learning}. From this viewpoint, any update by gradient descent (or any other optimization algorithms) in any levels of neural learning module is considered as a form of memory. Interestingly, our findings in \autoref{sec:backprop-memory-referential} on gradient descent being (self-referential) associative memory is aligned with this terminology. Furthermore, based on this terminology, in continuum memory system, the neural updates are applied at different frequencies and so memories are stored with different time scales, resulting in more robust memory management with respect to catastrophic forgetting.

\begin{myboxismall} [Memory and Learning from Nested Learning Perspective:] 
    Memory is not an isolated system and is distributed throughout the parameters. Particularly, any update that is caused by the input is a stored memory in the neural network, and the process of effectively store, encode, and in general acquire such memories is referred to as learning process. 
\end{myboxismall}

\headred{A General Note on Parameters of a Model}
The parameters of a model are one of its critical components that shape the units for knowledge storage, internal computations, and adaptability. 
Over the past decades, only (a subset of) the parameters in the architecture of machine learning models have been referred to as the \emph{learnable} entities, mainly due to the fact that they are the only components that we have been directly and intentionally optimizing over the training data. From the NL viewpoint, such parameters are placed in the lowest frequency level (the highest level) and are updated for every (batch of) samples. They are, however, are not the only parameters that contributes to the model internal computation, knowledge storage, and adaptability. As discussed earlier in \autoref{sec:momentum-based}, momentum is an example of such cases, where its parameters are updated over time (by gradient descent) and stores the knowledge about the loss landscape of the model so far. Such information are critical when the model is continually learn, mainly due to the fact that to find an effective solution, the optimizer needs to have more information about the global properties of the loss landscape. Another example of such cases is the memory (or hidden state) of recurrent neural networks. Although those parameters are not directly optimized in the lowest frequency level, they store important knowledge about the current context. With the change of the context, the compressed knowledge in these parameters are removed due to the lack of~knowledge~transfer~between~these~levels.

\begin{myboxismall} [Models Have More Parameters Than We Knew:] 
    \textit{The parameters of a neural learning module are not limited to those optimized in the pre-training level; all parameters that appear in the NL representation of the model contribute to its performance and expressivity.}
\end{myboxismall}

\headred{More Computations per Neuron} One common misinterpretation about the concept of NL is to restrict the model design and stacking multiple levels to CMS case. In general, stacking levels can help the model to enhance the depth of computations and also perform more internal computations per each parameter in the lowest frequency level. One example of such designs is Muon optimizer and $\texttt{NewtonSchulz}_k(\cdot)$ operation (see \autoref{eq:muon1} - \ref{eq:muon2}), which we showed that is equivalent to an internal optimization process. In this design, per each step of momentum update, we need $k$-steps of internal process to learn how to map gradients to an orthogonal space.

\headred{In-Context Learning}
Throughout this paper, we use the most general definition of ``in-context learning'' and refer to it as the ability of a model to adapt to and learn from a given context. Following the definition of \nsam, each block or level has its own context flow and so any neural update or adaptation to that context is considered as a form of in-context learning. Due to the popularity of Transformers as well as being the first model that in-context learning is studied for, the concept of in-context learning sometimes is referred to as conditioning the output on the entire context. Considering the general term of in-context learning, this formulation is only one of the instances of in-context learning, which we referred to as \emph{non-parametric in-context learning}. In general, however, the memory in recurrent models is performing in-context learning, in which the output is conditioned on the \emph{compressed context}. Therefore, from NL perspective, all the levels are performing in-context learning but on their own context flow with their own learning update and optimization process. 

Building on this definition, in-context learning is a model's capability  that is transparent from its NL representation, and per se it is not an emergent characteristic but a direct consequence of having multiple levels in the NL representation of the neural learning module. Although this might seem contradictory with previous claims about ICL being an emergent characteristic~\citep{brown2020language, singh2023transient}, it is notable that the good performance of the model in ICL tasks also requires a powerful low-frequency level, enabling the high frequency level to adapt fast. When the model is not well-trained, the higher-frequency level is on its own to learn from the context. This setup might result in a poor performance, mainly due to the fact that there might be not enough data in context to allow the high-frequency parameters to converge.

\headred{(Test-Time) Learning/Memorization}
Recently, the concept of test time training~\citep{sun2024learning, wang2025test} or test time memorization~\citep{behrouz2025Miras} has gained popularity as a backbone framework to design powerful sequence models. In these frameworks, given the context, a new component/block aims to compress the context into its parameters using a learning rule and objective function. In this formulation when the context is removed the acquired in-context knowledge diminishes along with it. As also discussed in the previous part, this update mechanism and learning process is indeed an instance of \emph{``parametric in-context learning''}:

\begin{myboxismall}[Test Time Training/Memorization are Instances of In-Context Learning]
    \textit{The concepts commonly referred to as test-time training and test-time memorization are in fact instances of parametric in-context learning, where the acquired in-context knowledge does not persist once the current context is removed}. 
\end{myboxismall}

Notably, when moving toward continual learning setup, there is no test-time or training-time and so it can be misleading to refer to (parametric) in-context learning as test-time learning/memorization.

\headred{Pre-training and Test Time}
From the nested learning perspective, the lowest frequency level (i.e., the highest level) corresponds to a learning phase that is often referred to as pre-training. Accordingly, pre-training is one of the levels and so has its own context flow (i.e., pre-training dataset), objective (e.g., next token prediction), and optimization process (e.g., AdamW). Accordingly, one can interpret the pre-training as one of the possible instances of in-context learning, where the context is the entire pre-training data.

\begin{myboxismall} [Pre-training is In-Context Learning with Ultra-Large Context Length:] 
    \textit{From NL's viewpoint pre-training is only one of the possible instances of in-context learning, where the context is the entire pre-training data. The distinction of training and test time in models is the results of disconnecting the knowledge transfer process from the highest frequency level (e.g., the context of Transformers) to the low frequency levels (i.e., pre-training)}.
\end{myboxismall}

This formulation and viewpoint is specifically important when we shift from pre-training paradigm to models that are capable of continually interact and learn from data/world (e.g., \citet{sutton2025oak}):

\headred{Continual Learning} 
From NL's perspective each phase of training for a model is defined as one of the low-frequency levels, which by design, we might want to stop the data processing in one level (e.g., ``End of Pre-training''), or continue it without any knowledge transfer to other levels (e.g., conventional formulation of in-context learning in Transformers). Accordingly, any machine learning model, no matter if it is during its pre-training or at test time, is performing continual learning as given a data sample, it requires performing internal computations to provide the output. However, the knowledge from that learning might not last or transfer to more persistent levels, mainly due to the lack of knowledge transfer between levels.

\begin{myboxismall}[No Training or Test Time in Neural Learning Modules:]
    \textit{For a neural learning module, there is no boarder and clear distinction between training and test time. The model only experiences two different states: when it receives information as input, or when it is an isolated learning system}.
\end{myboxismall}

We also revisit some of the common terms in the architectural design in the following:

\headred{Existing Architectural Backbones and Hybrid Models}
As discussed earlier in \autoref{sec:revisit-brain-perspective}, from the NL's perspective all modern architectures are uniform and in fact are feedforward layers (linear or non-linear MLP blocks) that are trained based on their own context flow and optimization problem. When viewing models from deep learning perspective, we see the final solution of such optimization problem (i.e., we see attention rather than a non-parametric solution to a regression loss), which results in the illusion of having distinct and non-uniform architectures.

\begin{myboxismall}[Recurrent Models are Replacing MLP Blocks:]
    \textit{From NL's viewpoint, (deep or linear memory) recurrent models are MLP blocks that a new level is added to their internal computation. Accordingly, existing hybrid architectures can be seen as conventional Transformer models, when we added a new level of computation to some of the MLP blocks}.
\end{myboxismall}

Another important aspect of understanding the existing architectures during their so-called ``pre-training'' phase is to understand how they transfer their knowledge from one level to another. 

\begin{myboxismall}[Knowledge Transfer from In-Context Learning]
    \textit{Although both modern deep and linear recurrent models are unified using associative memory perspective, there is still an important difference between the existing instances: While deep memory modules such as Titans, Atlas, Miras, and TTT take advantage of knowledge transfer from their high-frequency level to their lower-frequency level through meta-learning the initial state of memory, most linear memory recurrent models have no knowledge transfer process between their levels}.
\end{myboxismall}

\headred{Neural Learning Module as an Inter-Connected System}
One of the critical messages in NL is the fact that neural learning modules are inter-connected systems, meaning that the design of each component can significantly affect the design of other parts. This fact motivates follow up and future studies to better understand how one can properly design a neural learning module with all components work together in a harmony. As an example of such, the context of optimizers (i.e., gradients) is generated by the architecture component. Therefore, different architectures might show different characteristics in the generated gradient patterns and so one optimizer might not be the best option for all the architectures~\citep{zhang2024transformers}.

\begin{myboxismall}[Architectures Generates the Context for Optimizers]
    \textit{Neural learning modules are inter-connected systems, where architecture generates the context for optimizers (i.e., gradients). Therefore, the proper memory management of gradients (i.e., optimization algorithm) relies on the choice of architectures. In future, when viewing models as a neural learning modules, we need to design architecture specific optimizers so this inter-connected system works perfectly in harmony}.
\end{myboxismall}

\headred{Optimizers vs. Learned Optimizers}
Finally, we want to emphasize that our formulation of momentum, gradient descent, and/or other gradient-based optimizers show that they are associative memory modules aiming to compress the data and gradients into their parameters. Such update and compression process is based on gradient descent and so has a very similar nature to the learning process of learned optimizers. From the NL's viewpoint, both vanilla optimizers as well as learned optimizers are instances of the same concept but with different frequency and context flow: Although the parameters of learned optimizers are located in the lowest frequency level (i.e., to train and be optimized along with other parameters in the pre-training), the parameters of vanilla optimizers are located in their own level and so has their own gradient flow.

\section{Continuum Multi-Timescale Memory System}\label{sec:cmlp}  
Existing architectural backbones consist of (1) a \emph{working memory} module (e.g., attention), which is responsible to actively fuse the information across sequence length, and (2) a feed-forward layer (e.g., MLP) that fuse information across features and acts as the persistent memory or knowledge storage of pre-training phase. From the NL perspective, pre-training is the phase that the most outer level of the learning module is updated over its \emph{limited} context flow. Accordingly, in the continual setup, such pre-training phase is also rarely updated over time, and so its corresponding knowledge storage needs to rarely be updated over time. Given this intuition, we extend the traditional view-point of long-term/short-term memory system and suggest a knowledge storage feed-forward for each level (frequency domain).

\subsection{Continuum Memory System (CMS)}
Given the definition of frequency (Definition~\ref{dfn:uf}), Continuum Memory System (CMS) is formalized as a chain of MLP blocks $\texttt{MLP}^{(f_1)}(\cdot), \dots, \texttt{MLP}^{(f_k)}(\cdot)$, each of which associated with a chunk size of $C^{(\ell)} := \frac{\max_{i} C^{(i)}}{f_{\ell}}$ such that given input $x = \{\vx_1, \dots, \boldsymbol{x}_T\}$ the output of the chain is calculated as (we disregard normalizations for the sake of clarity):
\begin{align}\label{eq:output-cms}
    \vy_t = \texttt{MLP}^{(f_k)}(\texttt{MLP}^{(f_{k-1})}(\cdots \texttt{MLP}^{(f_1)}(\vx_t))), 
\end{align}
where the parameters of $\ell$-th MLP block, i.e., $\boldsymbol{\theta}^{(f_{\ell})}$, are updated every $C^{(\ell)}$ steps:
\begin{align}\label{eq:c-mlp}
    \boldsymbol{\theta}^{(f_{\ell})}_{i+1} =  \boldsymbol{\theta}^{(f_{\ell})}_i - \begin{cases} \sum_{t = i - C^{(\ell)}}^{i}\eta^{(\ell)}_t f(\boldsymbol{\theta}^{(f_{\ell})}_{t}; \boldsymbol{x}_{t}) & \text{if \:\:} i \equiv 0 \:\: (\texttt{mod} \: C^{(\ell)}), \\ 0 & \text{otherwise}.
    \end{cases}
\end{align}
 Here $\eta^{(\ell)}_t$ are learning rates corresponds to $\boldsymbol{\theta}^{(f_{\ell})}$, and $f(\cdot)$ is the error component of an arbitrary optimizer (e.g., $\nabla \mathcal{L}(\boldsymbol{\theta}^{(f_{\ell})}_{t}; \boldsymbol{x}_t)$ in gradient descent). The conventional Transformer block~\citep{transformers} is a special instance of this formulation, where $k = 1$ and frequency of update is zero. Note that the objective of $\mathcal{L}(\cdot)$ is the objective of choice for the task at hand, e.g., for language modeling it is next token prediction objective. It is notable that \autoref{eq:c-mlp} provides an important interpretation: parameters $\boldsymbol{\theta}^{(f_{\ell})}_{t}$ are responsible for compressing their own context into the their parameters and so they are a representative of abstract knowledge of their context.

As discussed in \autoref{sec:knowledge-transfer}, different levels might have different process of knowledge transfer. Accordingly, while the above formulation suggests a spectrum of memory systems in different levels and so with different frequencies, their connections can vary based on the design. In the following, we discuss some potential variants:

\head{Nested Continuum Memory Systems}
The first variant is a fully nested continuum memory system, in which the initial state of the MLP block in level $s+1$ is meta-learned in level $s$. This design allows for higher-order in-context learning ability, where each of the levels has its own context flow and re-initialized after the end of the context. More specifically, given an arbitrary $1 \leq s \leq k$, 
\begin{align}
    \boldsymbol{\theta}^{(f_{s+1})}_0 =
\underset{\Phi}{\arg\min}
\;\;
\mathbb{E}_{\mathcal{T} \sim \mathcal{C}^{(s)}}
\Biggl[
    \ell(\Theta, \mathcal{T}; \Phi)
\Biggr],
\end{align}
where $\mathcal{C}^{(s)}$ is the context length of the MLP block in $s$-th level. Following this design, at the end of the optimization process of each block (i.e., after $\lceil C^{(s)}/C^{(s+1)} \rceil$ steps.) the value of the memory will be re-initialized to $ \boldsymbol{\theta}^{(f_{s+1})}_0$. Note that the update mechanism of each block in its own level remain unchanged (i.e., \autoref{eq:c-mlp}).

\head{Sequential Continuum Memory Systems} 
In the second variant, the MLP blocks are located sequentially (i.e., the output of the MLP block in level $s$ is the input for the MLP block in level $s+1$) and also the initial state of MLP blocks are all connected through backpropagation in the lowest frequency level. Given an arbitrary $1 \leq s \leq k$, 
\begin{align}\label{eq:sequential-cms}
    \boldsymbol{\theta}^{(f_{s})}_0 =
\underset{\Phi}{\arg\min}
\;\;
\mathbb{E}_{\mathcal{T} \sim \mathcal{C}^{(1)}}
\Biggl[
    \ell(\Theta, \mathcal{T}; \Phi)
\Biggr],
\end{align}
where $\mathcal{C}^{(1)}$ is the context length of the MLP block in the lowest frequency level. Since the initial state of all memories are meta-learned in the lowest frequency, the most persistent knowledge of all components is the compression of the same context flow.

\head{Independent (Head-wise) Continuum Memory Systems}
In this variant, we keep the knowledge transfer process in \autoref{eq:sequential-cms}, but change the output computation in \autoref{eq:output-cms}. While the previous formulation designs the memory system as a sequence of blocks, and so making their input/out dependent to each other, this variant uses independent blocks with different context length and then combine them using an aggregation process:
\begin{align}\label{eq:independent-variant}
    \vy_t = \texttt{Agg} \left(\texttt{MLP}^{(f_k)}(\vx_t), \texttt{MLP}^{(f_{k-1})}(\vx_t), \cdots, \texttt{MLP}^{(f_1)}(\vx_t) \right).
\end{align}
The above $\texttt{Agg} \left(\cdot\right)$ is an arbitrary function that aggregates all the inputs to compute the output. For example, one straightforward and simple design choice is to use a learnable weighted sum of the input.

\head{CMS Design Helps with Continual Learning} 
Based on the design of CMS, a fair question is to ask: Why and how CMS can help with longer context length and generally continual learning. Here, we provide a simple answer to this question: Viewing MLP blocks in CMS as the storage of model's knowledge catastrophic forgetting can happen when we update a block and as its result, the old knowledge stored in its parameters are forgotten. In CMS design, however, when updating an arbitrary block of $\texttt{MLP}^{(f_s)}(\cdot)$ for some $1 \leq s \leq k$, the potentially forgotten knowledge from $\texttt{MLP}^{(f_s)}(\cdot)$ is still stored in other components such as $\texttt{MLP}^{(f_{s'})}(\cdot)$, where $s' < s$. Also, in this case (i.e., the knowledge is already forgotten from $\texttt{MLP}^{(f_s)}(\cdot)$ but it is still in $\texttt{MLP}^{(f_{s'})}(\cdot)$ for $s' < s$) the knowledge transfer through backpropagation (for their initial state) can circle back the knowledge to $\texttt{MLP}^{(f_s)}(\cdot)$, resulting in a loop through time dimension, and so hardly forgetting important knowledge.

\headdot{Is CMS Efficient Enough?}\label{sec:training-cmlp}
A common concern when updating the parameters of a model in a continual manner is its efficiency. Therefore, a fair question is to ask if CMS causes significant computational overhead for the model. To answer this question, let us recall from \autoref{sec:arch} that modern recurrent neural networks are also continually updating a subset of their parameters (i.e., their memory state). These parameter updates, however, take advantage of sequence parallelization as well as updating only a small number of parameters. To this end, for CMS, we highlight two points: 
\begin{itemize}
    \item In the CMS design, at each time, updates are restricted to blocks approaching their scheduled update time (based on their frequency). As a simple example, consider a Transformers but with replacing its MLP blocks with CMS (later in \autoref{sec:hope}, refer to this variant as \model-Attention). Let the model have $L_{\text{layer}}$ layers, 4 levels of MLP blocks in CMS with highest frequency of $\hat{f}$, and hidden dimension of $d_{\text{in}}$. On average, the update cost is for $\mathcal{O}\left(\frac{1}{\hat{f}} \times \frac{L_{\text{layer}}}{5} \times d_{\text{in}}^2 \right)$ of parameters, which consists of only a small number of parameters at each time. 
    \item The update mechanism of \autoref{eq:c-mlp}, not only helps with the enhancing the persistent memory of the model, but it also unlocks the sequence parallelization for higher frequency levels. More specifically, for input $\vx_i$ when $i \nequiv 0 \:\: (\texttt{mod} \: C^{(\ell)})$ there is no sequential process inside the chunk and so all the computations for tokens correspond to different values of $i \nequiv 0 \:\: (\texttt{mod} \: C^{(\ell)})$ can be done in parallel. The details of such training algorithm is the same as the training procedure in \citet{sun2024learning, behrouz2024titans}.
\end{itemize}
Therefore, in summary, CMS can be fast in practice, mainly due to the fact that it updates only small number of parameters at each time, and also its design unlocks sequence parallelization.

\begin{wrapfigure}{r}{0.45\textwidth}
\vspace*{-11ex}
    \begin{minipage}{0.45\textwidth}
        \begin{algorithm}[H]
            \caption{Multi-scale Momentum Muon (M3) }\label{alg:M3}
            \begin{algorithmic}[1]
            \Require Initial weights $\boldsymbol{\Theta}_0$, objective $\mathcal{L}(\cdot)$, learning rate $\eta > 0$, Newton–Schulz steps $T$, momentum factor $1 > \beta_1, \beta_2, \beta_3, \alpha \geq 0$, $\epsilon>0$, frequency $f$; 
            \State Initialize momentums: $\boldsymbol{M}^{(1)}_{0}, \boldsymbol{M}^{(2)}_{0}  \leftarrow \mathbf{0}$, $\boldsymbol{V}_{0} \leftarrow \mathbf{0}$;
            \For{lower-frequency iteration $k =0, 1, 2, \dots$}
                \State Slow Memory: $\boldsymbol{M}^{(2)}_{t} = \boldsymbol{M}^{(2)}_{t-1} + \beta_3 \sum_{i = (k-1)f}^{k\:f} \boldsymbol{g}_i$; 
                \State $\boldsymbol{O}^{(2)}_t \leftarrow \texttt{Newton–Schulz}_{T}\left( \boldsymbol{M}^{(2)}_{t}\right)$;
                \For{$t = k\: f +  1, k \:f + 2, \dots, (k+1)\: f$}
                    \State Compute Gradient: $\boldsymbol{g}_t = \nabla_{\boldsymbol{\Theta}_t} \mathcal{L}(\boldsymbol{\Theta}_t)$;
                    \State First Momentum: $\boldsymbol{M}^{(1)}_{t} = \boldsymbol{M}^{(1)}_{t-1} + \beta_1 \boldsymbol{g}_t$;
                    \State Second Momentum: $\boldsymbol{V}_{t} = \boldsymbol{V}_{t-1} + \beta_2 \boldsymbol{g}^{2}_t$;
                    \State $\boldsymbol{O}^{(1)}_t \leftarrow \texttt{Newton–Schulz}_{T}\left( \boldsymbol{M}^{(1)}_{t}\right)$;
                    \State $\boldsymbol{\Theta}_t \leftarrow \boldsymbol{\Theta}_{t-1} - \eta \: \frac{\boldsymbol{O}^{(1)}_t + \:\alpha \:\boldsymbol{O}^{(2)}_t}{\sqrt{ \boldsymbol{V}_{t}+ \epsilon}}$;
                \EndFor
            \EndFor
            \end{algorithmic}
        \end{algorithm}
    \end{minipage}
    \vspace{-4ex}
\end{wrapfigure}

\subsection{Continuum Memory System In Optimizers}
As a proof of concept and to support the effectiveness of CMS in different context flows, in this section we present Multi-scale Momentum/Memory Muon (M3) optimizer. Particularly, we aim to use NL's associative memory viewpoint to design an optimizer that not only compress the recent gradients effectively, but it also has a capability of incorporating the information about long past gradients. In \autoref{app:adam} (\autoref{eq:adam-objective2}) we discuss that how Adam optimizer is an instance of associative memory, in which the gradients are mapped to their variance until that point. Following the discussion about the need of long-context capability of optimizers in \autoref{sec:Optimizers-continual-learning}, we first replace the simple associative memory formulation of ${H}$ term in \autoref{eq:solution2} with our CMS (independent variant, \autoref{eq:independent-variant}) with a two-level memory system, which we refer to the memories as $\boldsymbol{M}^{(1)}$ and $\boldsymbol{M}^{(2)}$:
\begin{align}\nonumber
    &\boldsymbol{M}^{(1)}_{t} = \boldsymbol{M}^{(1)}_{t-1} + \beta_1 \vg_t,   \\ 
    &\boldsymbol{M}^{(2)}_{t} = \boldsymbol{M}^{(2)}_{t} - \beta_2 \begin{cases}
        &\sum_{i = t-\hat{C}}^{t}  \vg_i  \quad \quad \text{if} \:\:\:\:\:t \equiv 0 \:\:\: (\text{mod} \:\: \hat{C})\\
        &0 \qquad \quad \:\:\qquad \text{otherwise,}
    \end{cases}
\end{align}
where $\hat{C}$ is the chunk size that we update the lower-frequency momentum term. Finally, to aggregate the momentum terms (the choice of $\texttt{Agg}(\cdot)$ in \autoref{eq:independent-variant}), we use a simple weighted summation with the use of parameter $\alpha > 0$ as the coefficient of $\boldsymbol{M}^{(2)}_{t}$. Following our discussion on the importance of $\texttt{Newton-Schulz}_T(\cdot)$ to map the gradients to a proper metric space (see \autoref{sec:momentum-based} and \autoref{eq:muon-motivation}), following Muon~\citep{jordanmuon}, we use $\texttt{Newton-Schulz}_T(\cdot)$ on the output of the momentum terms, before aggregating them with weighted sum. This helps the associative memory to better manage its capacity by updating its parameters in a proper direction. The pseudocode for M3 is in \autoref{alg:M3}. In summary, one can say that M3 is the combination of Adam~\citep{kingma2014adam}, Muon~\citep{jordanmuon}, and our Continuum Memory System.

Notably, this optimizer is designed as a proof-of-concept to support the design of CMS. The M3 optimizer per se, however, might suffer from computational overhead and so face challenges when scaling to larger networks (see Figure~\ref{fig:optimizer-efficiency}). Furthermore, it is notable that the main point of M3 design based on CMS is to delay the update of the memory to gain longer-context. Similar studies such as the work of \citet{pagliardini2025the, ueaj2025multiscalemuon}, while using multiple momentum terms, approach this long-context momentum for optimizers using a controlled learning rate for the additional momentum terms.

\subsection{Ad-hoc Level Stacking: Initializing CMS with Pre-Trained Models}\label{sec:pre-trained-mlp} 
In our discussion on \autoref{fig:mlp-vs-lin-attention}, we observed that the initial state of the memory modules are optimized in lower-frequency levels and so one can interpret them as MLP blocks in the vanilla Transformer architectures~\citep{transformers}. Therefore, a natural question is if we can leverage pre-trained models to initialize CMS blocks. One of the important advantages of NL is its flexibility to view and modify parameters in different levels. That is, since each level has its own context flow and optimization process, one can simply initialize the parameters in each level independently so that it helps the model to adapt faster to the levels' context flow. To this end, in this section, we suggest initializing the parameters in a level with a model's pre-trained weights. More formally, given a CMS with $\{\texttt{MLP}^{(f_i)}(\cdot)\}_{i = 1}^{k}$, and a set of pre-trained MLP blocks $\{\texttt{MLP}_{\text{pre-trained}_i}(\cdot)\}_{i=1}^{k}$ we use \autoref{eq:c-mlp} to update $\{\texttt{MLP}^{(f_i)}(\cdot)\}_{i = 1}^{k}$ in different levels; we, however, use the trained parameters of $\{\texttt{MLP}_{\text{pre-trained}_i}(\cdot)\}_{i=1}^{k}$ as the initial state of CMS blocks: $\texttt{MLP}^{(f_i)}_0(\cdot) = \texttt{MLP}_{\text{pre-trained}_i}(\cdot)$.

\headdot{Why This Initialization Should Work?} In NL, when there is a knowledge transfer process between two levels, the higher frequency level can take advantage of the knowledge stored in the lower frequency level and so adapt faster to its own context flow. The internal learning rate in the higher-frequency level, however, can control the capacity of the model for adaptability. That is, consider the above case, where all the blocks are initialized with pre-trained MLP blocks, setting $\eta_t^{(\ell)} \rightarrow 0$ keeps the updated memory blocks close to their initial states, resulting in directly using of pre-trained blocks, without adaption. Later in \autoref{sec:experiments}, we use this method to adapt pre-trained Transformer architectures to the \model's setup.

\begin{figure*}
    \centering
    \includegraphics[width=0.9\linewidth]{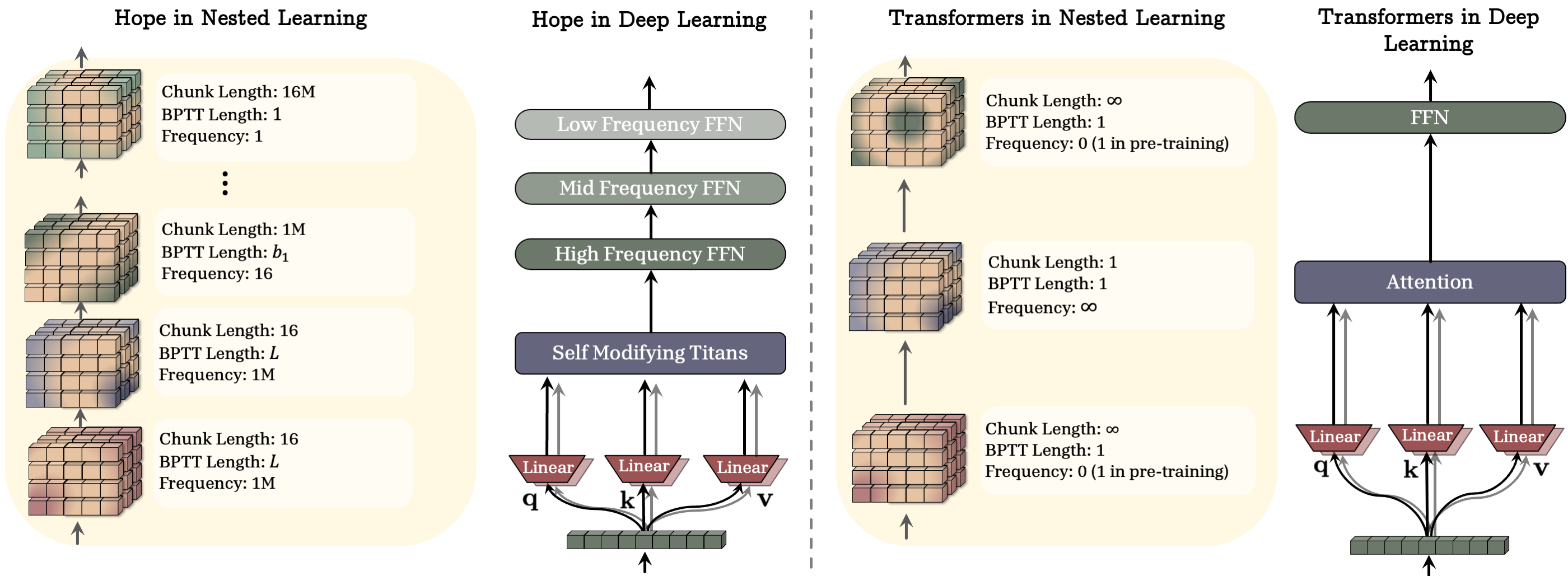}
    \caption{A comparison of Hope architectural backbone with Transformers (Normalization and potential data-dependent components are removed for the sake of clarity).}
    \label{fig:Hope-vs-Transformer}
\end{figure*}

\section{\model: A Self-Referential Learning Module with Continuum Memory}\label{sec:hope}
As we discussed earlier in \autoref{sec:revisit-brain-perspective}, architectures in nested learning are uniform, i.e., a set of feedforward neural network blocks, each of which with its own context, update frequency, and internal objective. Sequence models–a common term to refer to blocks often with the highest update frequency that fuse information across tokens in the input sequence–are critical components for memory management and in-context learning ability of models.  Following our earlier discussion in \autoref{sec:arch}, modern sequence models can be seen as associative memories and so are nested optimization problems. From this perspective, global softmax attention or its more expressive higher-order variants are perfect memories (forcing to cache all past tokens) with frequency update of infinity as they are non-parametric solutions for optimizing (local) $L_2$-regression objective with Nadaraya-Watson estimators~\citep{fan2018local, zhang2022analysis} (see \autoref{eq:attention-nadaraya}). Therefore, parametric solutions (e.g., modern RNNs) for the similar objectives and when the parameter search space are the same (i.e., matrix-valued memory) are not expected to outperform softmax attention when the model size and data scales. To this end, and to design powerful sequence models, we need to understand where Transformers are limited and how one can overcome such limitations. 

From the nested learning perspective, Transformers are two-level components, where projections and MLP blocks are optimized in the first level and the second level is responsible for in-context learning with finding the non-parametric solution and so conditioning the output on the context. This design, however, has limited computational depth as also stated in recent studies on state-tracking and similar computational capabilities of models~\citep{merrill2024the, sanford2024transformers, grazzi2025unlocking}. Furthermore, Transformers' parameters are static throughout their context, meaning that their found solution to map tokens in the context (since it is non-parametric solution) remains the same and so they lack the ability to modify themselves (at least in-context). More specifically, the initial linear blocks, $W_{\vk}, W_{\vv},$ and $W_{\vq}$, that projects input data to keys, values, and queries, are fixed after the pre-training stage (i.e., are in the first level) and so the Transformer's ability to contextualize and map tokens is bounded by the knowledge stored in these blocks. For example, given a 1-layer Transformer, the projection of each token is a function of the token itself and its position; therefore, as an example, it can miss the diverse possible encodings of words whose meaning depend on the context, rather than the word itself. Although with increasing the depth of the model this issue might fade in later layers, we should not rely on the depth to compensate the models ability as it still is a bottleneck to unleash the capability of the model in earlier layers.

To overcome the above challenge, recently, the use of short convolutions and canon layers~\citep{Allenzhu2025-canon} have became a de facto component in modern models. Despite their success in mixing local tokens, still the models are fundamentally limited to adapt to the context and capture the global information beyond the local mixing. In the next part, we discuss a fundamental solution by presenting self-referential Titans that allows all the components to perform in-context learning, and adapt and modify themselves:

\subsection{Deep Self-Referential Titans}\label{sec:SRT}
A general formulation for the associative memory-based blocks is to project the data into keys, values, and queries and learns how to map keys to values and how to retrieve from the mapping based on queries. More formally, for a parametric associative memory, let $\boldsymbol{x}_t \in \mathbb{R}^{d}$ for $t = 1, \dots, L$ be the input, we have:
\begin{align}
    &\vk_t = \boldsymbol{x}_t \textcolor{mydarkred}{W_{\vk}} , \qquad \vv_t = \boldsymbol{x}_t \textcolor{mydarkred}{W_{\vv}} , \qquad \vq_t = \boldsymbol{x}_t \textcolor{mydarkred}{W_{\vq}}, \qquad \eta_t = \boldsymbol{x}_t \textcolor{mydarkred}{W_{\eta}}, \qquad \alpha_t = \boldsymbol{x}_t \textcolor{mydarkred}{W_{\alpha}},\\
    &\min_{\textcolor{c4}{\M}} \:\: \mathcal{L}\left(\textcolor{c4}{\M}; \vk_t, \vv_t \right), \qquad \qquad \:\:\: \textcolor{c4}{\text{with an optimization algorithm}} \\
    &\boldsymbol{y}_t = \textcolor{c4}{\M_{t}} \vq_t \:.
\end{align}
For the sake of clarity, we use \textcolor{mydarkred}{red} (resp. \textcolor{c4}{blue}) to highlight computations/weight in the upper level (resp. lower level). Similar to example in \autoref{fig:mlp-vs-lin-attention}, we can add a new level for each of $W_{\vk}, W_{\vv}, W_{\vq}, W_{\eta},$ and $W_{\alpha}$ and allow them to be updated in-context. For the sake of efficiency, a simple version is to share the values for all the components in the nested system of associative memories: 
\begin{align}
    & \:\vk_t =  \textcolor{c4}{\M_{\vk, t-1}} \left(\boldsymbol{x}_t \right), \qquad \vv_t = \textcolor{c4}{\M_{\vv, t-1}}\left(\boldsymbol{x}_t \right) , \qquad \vq_t =  \textcolor{c4}{\M_{\vq, t-1}}\left(\boldsymbol{x}_t \right), \qquad \eta_t = \textcolor{c4}{\M_{\eta, t-1}}\left(\boldsymbol{x}_t \right), \qquad \alpha_t = \textcolor{c4}{\M_{\alpha, t-1}} \left(\boldsymbol{x}_t \right),\\
    &\min_{\textcolor{c4}{\M_{\square}}} \:\: \mathcal{L}\left(\textcolor{c4}{\M_\square}; \square_t, \vv_t \right), \qquad \quad \textcolor{c4}{\text{with an optimization algorithm}}, \quad \square \in \{\vk, \vv, \vq, \eta, \alpha \}, \\
    &\min_{\textcolor{c4}{\M_{\text{mem}}}} \:\: \mathcal{L}\left(\textcolor{c4}{\M_{\text{mem}}}; \vk_t, \vv_t \right), \qquad \textcolor{c4}{\text{with an optimization algorithm}}, \\ 
    & \:\boldsymbol{y}_t = \textcolor{c4}{\M_{\text{mem}, t}}\left(\vq_t \right) \:, 
\end{align}
where the initial states of all memories, i.e., $\textcolor{mydarkred}{\M_{\square, 0}}$ for any $\:\square \in \{\vk, \vv, \vq, \eta, \alpha, \text{memory} \}$ are meta-learned across all sequences/contexts. As discussed earlier, the meta-learning of the initial states of memories is essential for both fast-adaption, training stability, robustness to noise in the data.

This design provides a fully adaptive memory, where all the components can adapt themselves in-context. It, however, (1) still lacks self-modification, where the model in response to new data changes its own parameters or learning process~\citep{schmidhuber2003godel}; (2) has suboptimal design as it shares of keys and values for all the memories. In continual learning, where the model requires consistent weight/knowledge update in response to new data, it is critical for the model to not solely rely on data, and instead learns how to modify itself when it is needed. Motivated by the above points, and inspired by the self-modifying mechanisms that generate their own values based on the context~\citep{schmidhuber2003godel, schmidhuber1993self, irie2022modern}, we present self-modifying deep associative memory where the models generate their own values:
\begin{align}
    &\boldsymbol{y}_{t} = \textcolor{c4}{\M_{\text{memory}, t-1}}\left(\vq_t \right), \qquad \vk_t =  \textcolor{c4}{\M_{\vk, t-1}} \left(\boldsymbol{x}_t \right), \qquad \vv_t =  \textcolor{c4}{\M_{\vv, t-1}}\left(\boldsymbol{x}_t \right), \qquad \eta_t = \textcolor{c4}{\M_{\eta, t-1}}\left(\boldsymbol{x}_t \right), \qquad \alpha_t = \textcolor{c4}{\M_{\alpha, t-1}} \left(\boldsymbol{x}_t \right),\\
    &\hat{\vv}_{\square, t} =  \textcolor{c4}{\M_{\square, t-1}} \left(\vv_t \right),  \qquad \:\: \quad \text{(Generating its own values for each memory)}\\ \label{eq:AM-srt}
    &\min_{\textcolor{c4}{\M_{\square}}} \:\: \mathcal{L}\left(\textcolor{c4}{\M_\square}; \vk_t, \hat{\vv}_{\square, t} \right), \qquad \textcolor{c4}{\text{with an optimization algorithm}}, \quad \square \in \{\vk, \vv, \vq, \eta, \alpha, \text{memory} \},
\end{align}
where $\vq_t = \boldsymbol{x}_t \textcolor{mydarkred}{W_{\vq}}$ is the only non-adaptive projection, $\eta_t$ is the learning rate in optimization process, and $\alpha_t$ is the retention gate (forget gate or weight decay) in the optimization process. Note that, again, the initial states of all memories, i.e., $\textcolor{mydarkred}{\M_{\square, 0}}$ for any $\:\square \in \{\vk, \vv, \vq, \eta, \alpha, \text{memory} \}$ are meta-learned across all sequences/contexts, and so are optimized in the higher levels (or outer-loop).

Learning the mappings for associative memory modules (see \autoref{eq:AM-srt}) requires a choice of optimization algorithm as well as an objective $\mathcal{L}$ that measures the quality of mappings. A simple and common choice for objective and optimization process are $L_2$-regression loss, and gradient descent algorithm. As for the objective, we use $L_2$-regression loss, i.e., $\mathcal{L}(\M; \vk, \vv) = \| \M(\vk) - \vv \|_2^2$. As discussed earlier (see \autoref{sec:L2-backprop}), the choice of optimizer highly depends on the context of optimization. For example, gradient descent from associative memory perspective is based on dot-product similarity and so the update at each step, is solely based on the input and does not incorporate the previous data samples to the update. When performing optimization in the token space, however, we know tokens are highly correlated. Therefore, following our discussion in \autoref{sec:L2-backprop}, we use our DGD with weight decay, resulting in general update rule of:
\begin{align}\label{eq:elements}
    &\boldsymbol{y}_{t} = \textcolor{c4}{\M_{\text{memory}, t-1}}\left(\vq_t \right), \qquad \vk_t =  \textcolor{c4}{\M_{\vk, t-1}} \left(\boldsymbol{x}_t \right), \qquad \vv_t =  \textcolor{c4}{\M_{\vv, t-1}}\left(\boldsymbol{x}_t \right), \qquad \eta_t = \textcolor{c4}{\M_{\eta, t-1}}\left(\boldsymbol{x}_t \right), \qquad \alpha_t = \textcolor{c4}{\M_{\alpha, t-1}} \left(\boldsymbol{x}_t \right),\\
    &\hat{\vv}_{\square, t} =  \textcolor{c4}{\M_{\square, t-1}} \left(\vv_t \right),  \qquad \qquad \qquad \qquad \qquad \qquad \qquad \qquad \qquad  \:\:\: \quad \text{(Generating its own values for each memory)}\\ \label{eq:update-srt}
    & \textcolor{c4}{\M_{\square, t}} = \textcolor{c4}{\M_{\square, t-1}} \left(\alpha_t \boldsymbol{I} \: - \: \eta_t \vk_t \vk_t^{\top} \right) - \eta_t \nabla \mathcal{L}_{\textcolor{c4}{\M_{\square, t-1}}} \left( \textcolor{c4}{\M_{\square, t-1}}; \vk_t, \hat{\vv}_{\square, t}\right), \quad \square \in \{\vk, \vv, \vq, \eta, \alpha, \text{memory} \}.
\end{align}
Here, the architecture of the memories are arbitrary and even we are not forced to use the same architecture for all components. We use a 2-layer MLP block as the architecture of all the memories:
\begin{align}
    \M_{\square} (\cdot) = (\cdot) + W_{\square, 1} \sigma(W_{\square, 2} (\cdot)).
\end{align}

\subsection{Fast and Parallelizable Training}\label{sec:Parallel-srt}
In the above, we discussed how to design a model that can learn to generate its own latent values and so modify itself. The main challenge from the practical point of view is the efficiency of the method and if its training is parallelizable. We follow the chunk-wise training algorithm of non-linear update rules~\citep{sun2024learning, behrouz2024titans} and use update frequency of $f_{\square} = \frac{L}{C_{\square}}$, where $L$ is the context length. While there is no limitation to use different chunk-sizes, in our experiments, we use two different value of chunk sizes, one for the update of $\M_{\text{memory}}(\cdot)$ and the other for all the other memories in the self-referential Titans. 

In more details, given an input sequence $\{\boldsymbol{x}_t\}_{t=1}^{L}$ and chunk size $1 \leq C \leq L$, we split the sequence into $\lceil \frac{L}{C} \rceil$ chunks of $\{\boldsymbol{x}_{((i-1)C + t)} \}_{t = 1}^{C}$ for $i = 1, \dots , \lceil \frac{L}{C} \rceil$, and then generate all elements in \autoref{eq:elements} at the end of each chunk for the next chunk. This allows for generating all the elements for the entire chunk in parallel, before starting the computation for this chunk. Furthermore, to update the memory modules based on \autoref{eq:update-srt}, we take the gradient with respect to the last state of the previous chunk. Again, this allows for computing all the gradients for the next chunk in parallel. In more details, given this chunk-wise updating procedure, the update rule for the self-referential Titans is computed as:

\begin{align}\nonumber
    &\boldsymbol{y}_{t} = \textcolor{c4}{\M_{\text{memory}, C \times \lceil \frac{t}{C}\rceil}}\left(\vq_t \right), \quad \vk_t =  \textcolor{c4}{\M_{\vk, C \times \lceil \frac{t}{C}\rceil}} \left(\boldsymbol{x}_t \right), \quad \vv_t =  \textcolor{c4}{\M_{\vv, C \times \lceil \frac{t}{C}\rceil}}\left(\boldsymbol{x}_t \right), \quad \eta_t = \textcolor{c4}{\M_{\eta, C \times \lceil \frac{t}{C}\rceil}}\left(\boldsymbol{x}_t \right), \quad \alpha_t = \textcolor{c4}{\M_{\alpha, C \times \lceil \frac{t}{C}\rceil}} \left(\boldsymbol{x}_t \right),\\ \nonumber
    &\hat{\vv}_{\square, t} =  \textcolor{c4}{\M_{\square, C \times \lceil \frac{t}{C}\rceil}} \left(\vv_t \right),  \qquad \qquad \qquad \qquad \qquad \qquad \qquad \qquad \qquad  \: \qquad \text{(Generating its own values for each memory)}\\ \label{eq:update-srt2}
    & \textcolor{c4}{\M_{\square, t}} = \textcolor{c4}{\M_{\square, t-1}} \left(\alpha_t \boldsymbol{I} \: - \: \eta_t \vk_t \vk_t^{\top} \right) - \eta_t \nabla \mathcal{L}_{\textcolor{c4}{\M_{\square, C \times \lceil \frac{t}{C}\rceil}}} \left( \textcolor{c4}{\M_{\square, C \times \lceil \frac{t}{C}\rceil}}; \vk_t, \hat{\vv}_{\square, t}\right), \qquad \: \square \in \{\vk, \vv, \vq, \eta, \alpha, \text{memory} \}.
\end{align}
Here, the architecture of the memories are arbitrary and even we are not forced to use the same architecture for all components. We use a 2-layer MLP block as the architecture of all the memories:
\begin{align}
    \M_{\square} (\cdot) = (\cdot) + W_{\square, 1} \sigma(W_{\square, 2} (\cdot)).
\end{align}
Since all the gradients as well as new keys, values, learning-rates, and weight decays can be computed in parallel before starting the processing of the current chunk, the above updates accepts the fast parallelizable dual form that is discussed by \citet{sun2024learning} and \citet{behrouz2024titans}. To better illustrate the above update rule for self-referential Titans, let us derive the recurrent formula for the simplest case of matrix-valued memory. We derive the recurrent form for two different objectives:
\begin{itemize}
    \item Dot-product similarity $\mathcal{L}(\M; \vk, \vv) = - \inner{\M \vk}{\vv}$: Given this objective and linear memory, the gradient is calculated as $\vv \vk^\top$, which results in update rule of:
    \begin{align}
        & \textcolor{c4}{\M_{\square, t}} = \textcolor{c4}{\M_{\square, t-1}} \left(\alpha_t \boldsymbol{I} \: - \: \eta_t \vk_t \vk_t^{\top} \right) - \eta_t  \hat{\vv}_{\square, t} \vk_t^{\top}, \qquad \: \square \in \{\vk, \vv, \vq, \eta, \alpha, \text{memory} \}
    \end{align}
    \item $L_2$-regression loss: Given this objective and linear memory, the gradient is calculated as $(\M \vk - \vv)\vk^{\top}$, which results in update rule of:
    \begin{align}
        & \textcolor{c4}{\M_{\square, t}} = \textcolor{c4}{\M_{\square, t-1}} \left(\alpha_t \boldsymbol{I} \: - \: \eta_t \vk_t \vk_t^{\top} \right) - \eta_t  \left( \M_{\square, C \times \lceil \frac{t}{C}\rceil} \vk_t - \hat{\vv}_{\square, t} \right) \vk_t^{\top}, \qquad \: \square \in \{\vk, \vv, \vq, \eta, \alpha, \text{memory} \}.
    \end{align}
\end{itemize}

\subsection{Hope Neural Learning Module} \label{sec:hope-final} 
In the previous sections, we first discussed Continuum Memory System (CMS) that allows for more persistent storage of memories and defines memory as a spectrum of blocks with different frequencies of update. Due to the larger capacity and constraints for scaling the parameters, often CMS requires simple learning rule but higher capacity to store more persistent knowledge. On the other hand, in the previous section, we discussed the design of a self-modifying Titans, where it can generate its own keys and so learning update to better adapt to the context. Contrary to CMS, the self-modifying Titans has a small capacity but is using a complex and expressive learning rule. Accordingly, these two systems seem to be complementary and their combination can enhance the model expressiveness from different aspects. 

To this end, we present \model{} architecture: A neural learning module that incorporates self-modifying Titans followed by Continuum Memory System. The \model{} design is illustrated in \autoref{fig:Hope-vs-Transformer}. Formally, let $\boldsymbol{x}_t \in \mathbb{R}^{d}$ for $t = 1, \dots, L$ be the input, the \model{} forward pass is defined as (we remove the normalization and convolution layers for the sake of clarity):
\begin{align}\label{eq:hope1}
    &\boldsymbol{o}_{t} = {\M_{\text{memory}, t-1}}\left(\vq_t \right), \qquad \vk_t =  {\M_{\vk, t-1}} \left(\boldsymbol{x}_t \right), \qquad \vv_t =  {\M_{\vv, t-1}}\left(\boldsymbol{x}_t \right), \qquad \eta_t = {\M_{\eta, t-1}}\left(\boldsymbol{x}_t \right), \qquad \alpha_t = {\M_{\alpha, t-1}} \left(\boldsymbol{x}_t \right),\\ \label{eq:hope2}
    &\hat{\vv}_{\square, t} =  {\M_{\square, t-1}} \left(\vv_t \right),   \\ \label{eq:hope3}
    &{\M_{\square, t}} = {\M_{\square, t-1}} \left(\alpha_t \boldsymbol{I} \: - \: \eta_t \vk_t \vk_t^{\top} \right) - \eta_t \nabla \mathcal{L}_{{\M_{\square, t-1}}} \left( {\M_{\square, t-1}}; \vk_t, \hat{\vv}_{\square, t}\right), \quad \square \in \{\vk, \vv, \vq, \eta, \alpha, \text{memory} \}. \\ \label{eq:hope4}
    &\vy_t= \texttt{MLP}^{(f_k)}(\texttt{MLP}^{(f_{k-1})}(\cdots \texttt{MLP}^{(f_1)}(\boldsymbol{o}_t))),
\end{align}
where the block's output for token $t$ is $\vy_t$. In our experiments, we also normalize $\vq$ and $\vk$ with $L_2$ normalization and also use local convolutions with window size of 4.

\head{Hope-Attention} We also use another variant of \model, in which we simply replace the self-modifying Titans with softmax global attention~\citep{transformers}. 
\section{Experiments}\label{sec:experiments}
In this section, we empirically evaluate the performance of different components we discussed throughout the paper. More specifically, (1) we first focus on the presented optimization algorithms and compare them with state-of-the-art methods; (2) We then focus on in-context and continual learning tasks and show how the nested learning paradigm and more specifically higher order in-context learning enhances the capabilities of models. We compare the continuum memory system with simple MLP layers and discuss how a pre-trained model can be adapted to be a continual learner; (3) We then focus on the language modeling and long context understanding of \model{} model and compare it with Transformers and modern recurrent architectures. The details of the experiments and their setups are explained in the corresponding sections.

\begin{figure*}
    \centering
    \includegraphics[width=0.31\linewidth]{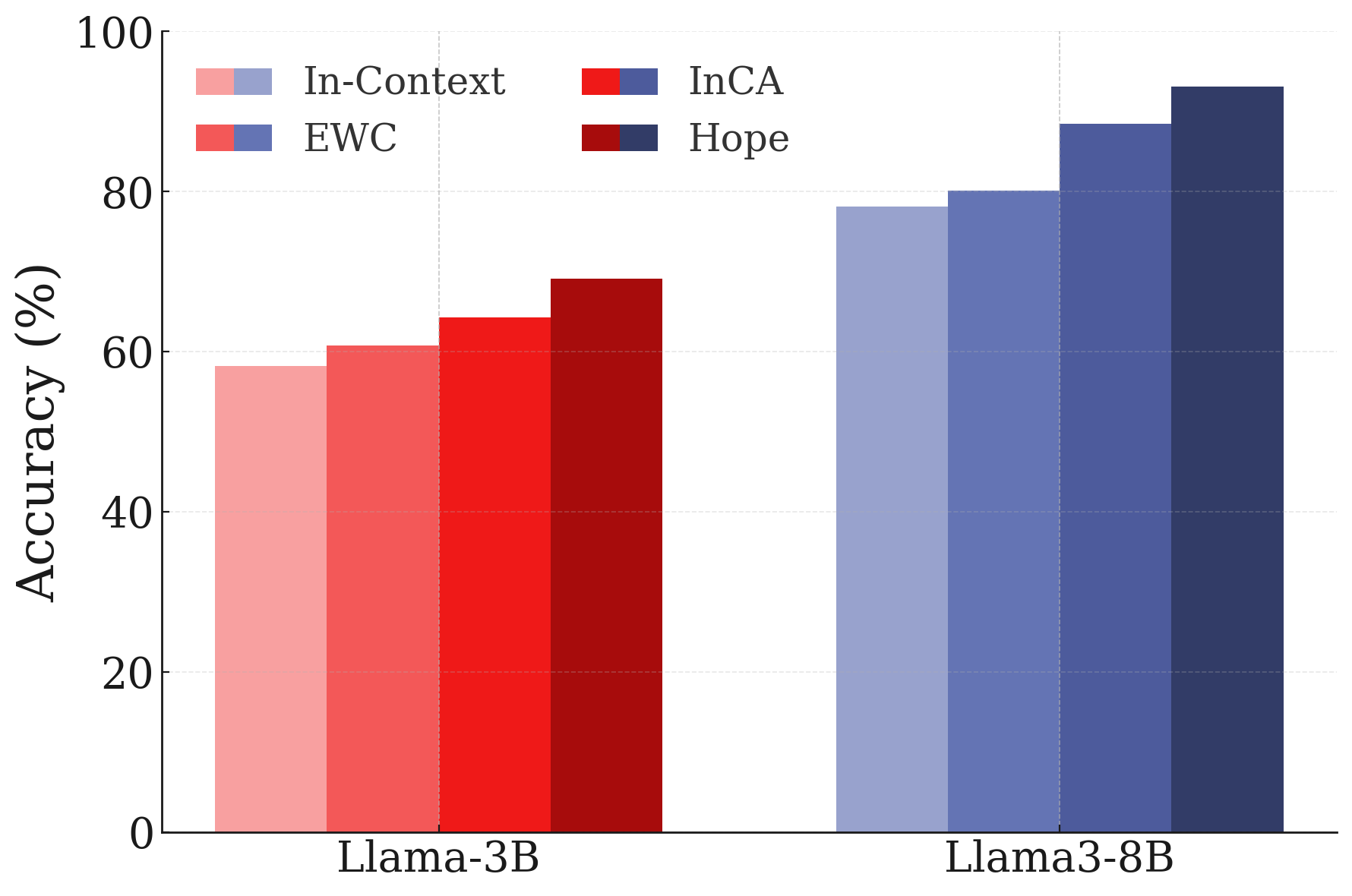}~\hfill~
    \includegraphics[width=0.31\linewidth]{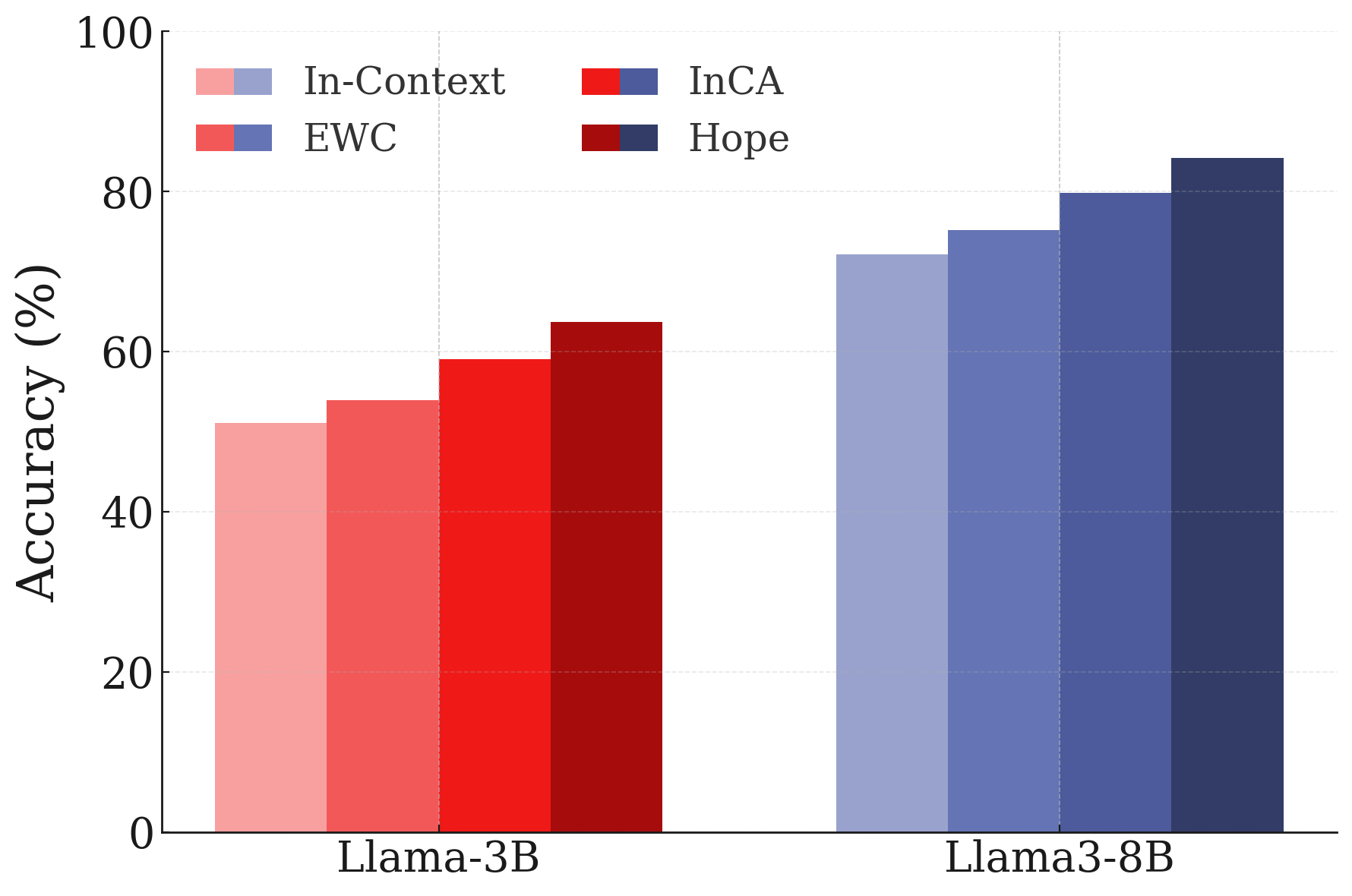}~\hfill~
    \includegraphics[width=0.31\linewidth]{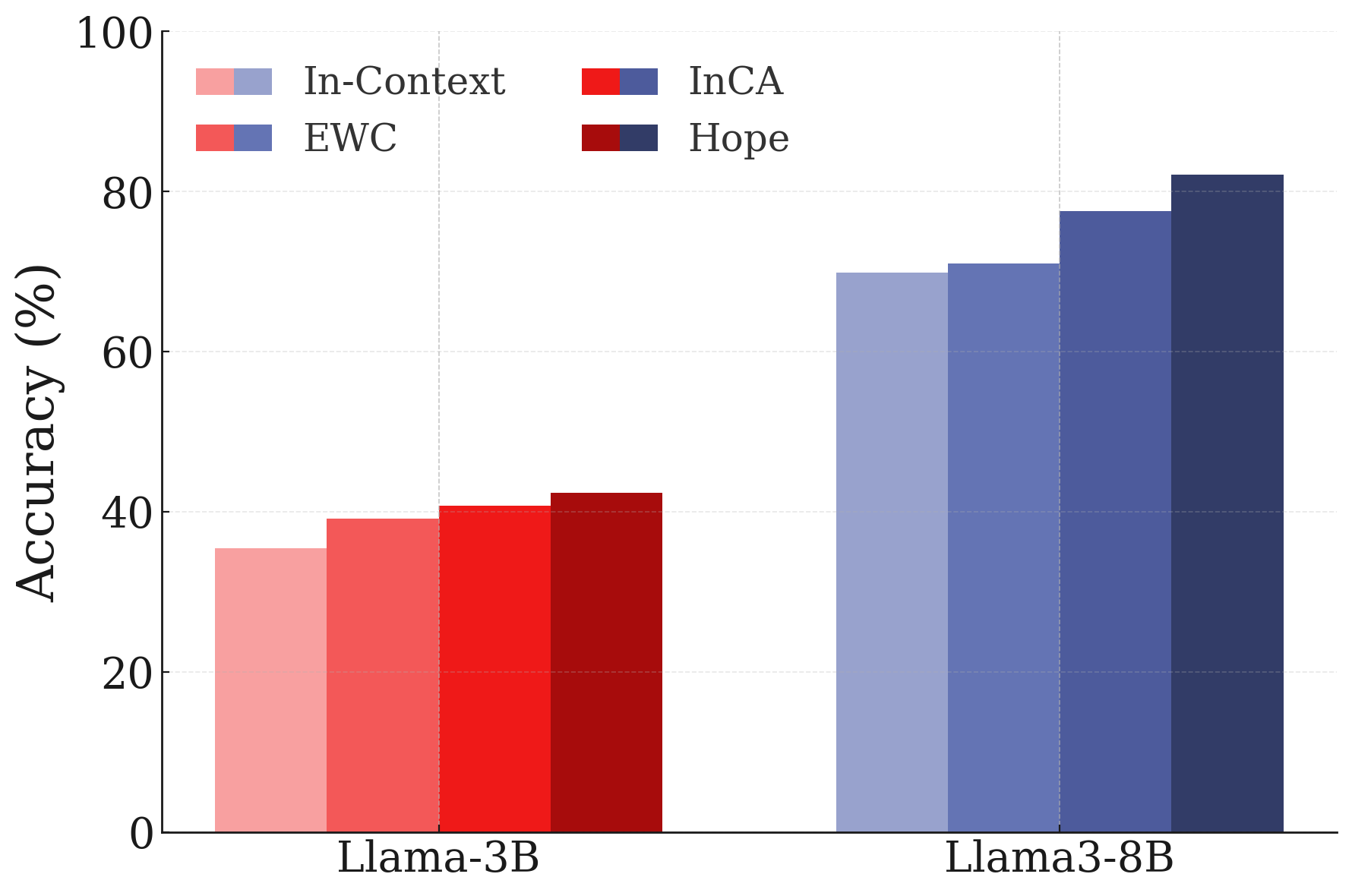}
    \caption{Class-incremental learning in the text classification domain on (Left) CLINC dataset~\citep{larson2019evaluation}, (Middle) Banking dataset~\citep{casanueva2020efficient}, and (Right) DBpedia dataset~\citep{auer2007dbpedia}. \model-enhanced architecture achieves the best accuracy among other continual learning methods, including ICL.}
    \label{fig:CIL}
\end{figure*}

\begin{figure*}
    \centering
    \includegraphics[width=0.33\linewidth]{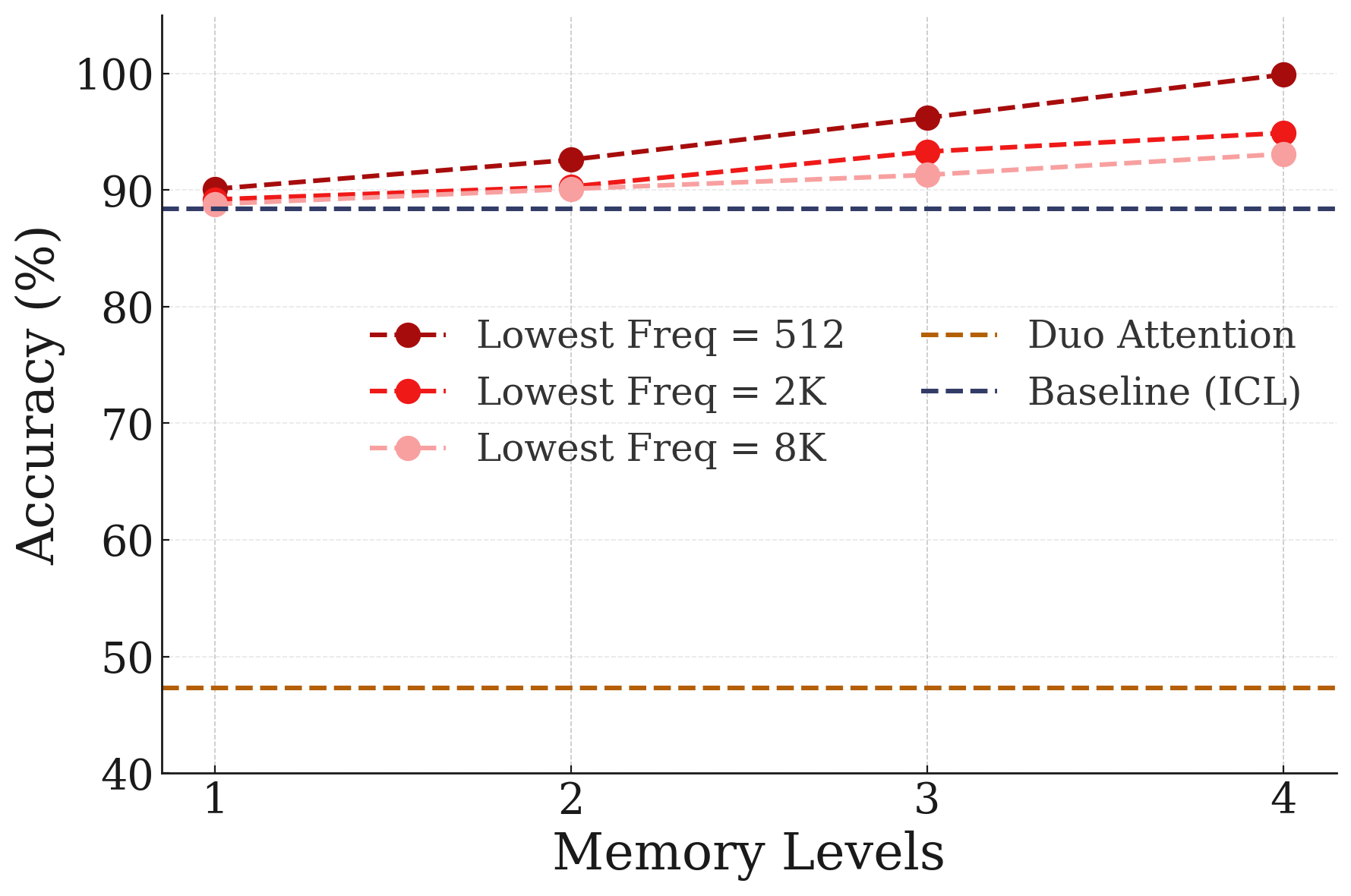}
    \includegraphics[width=0.33\linewidth]{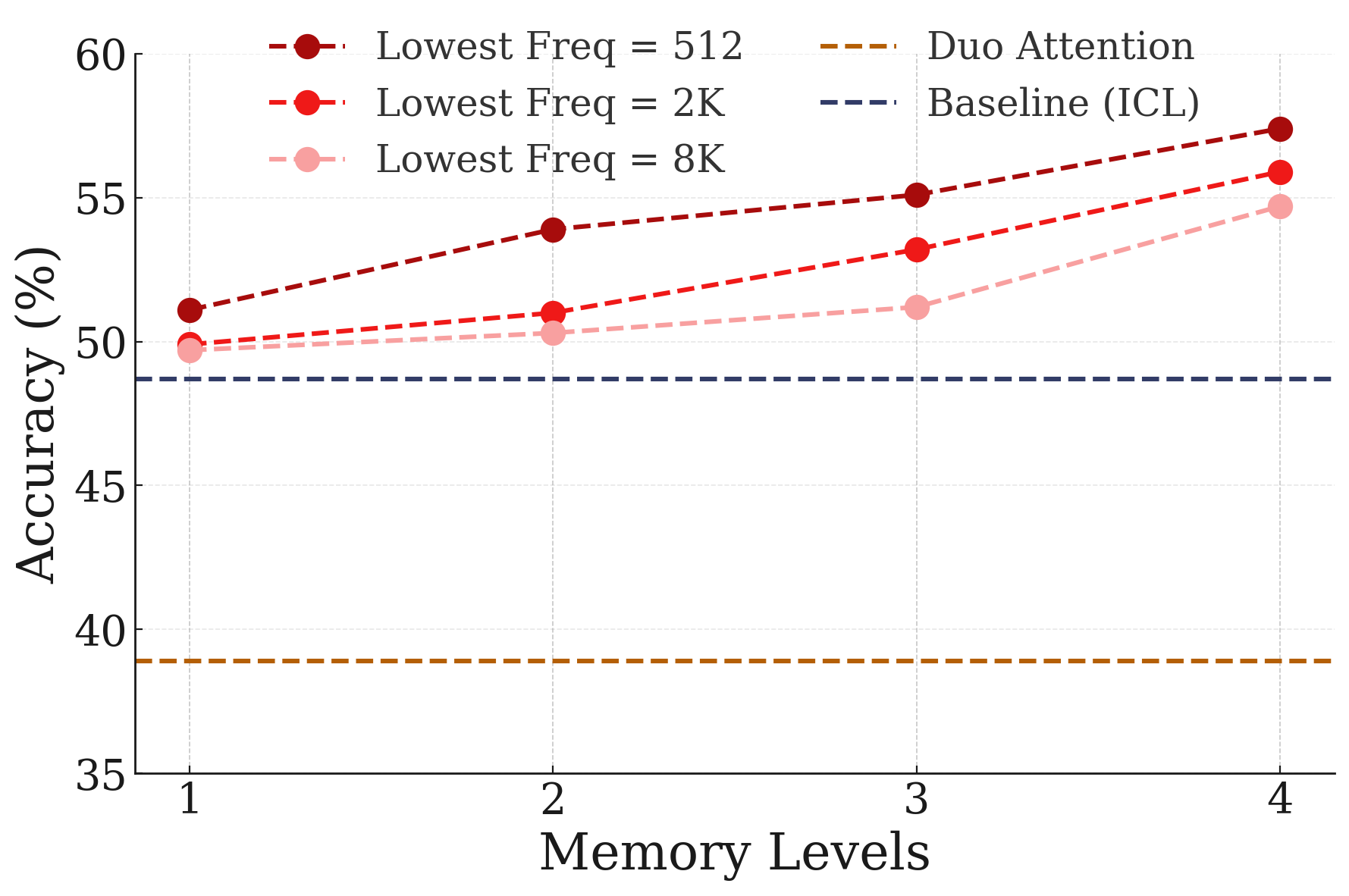}
    \includegraphics[width=0.33\linewidth]{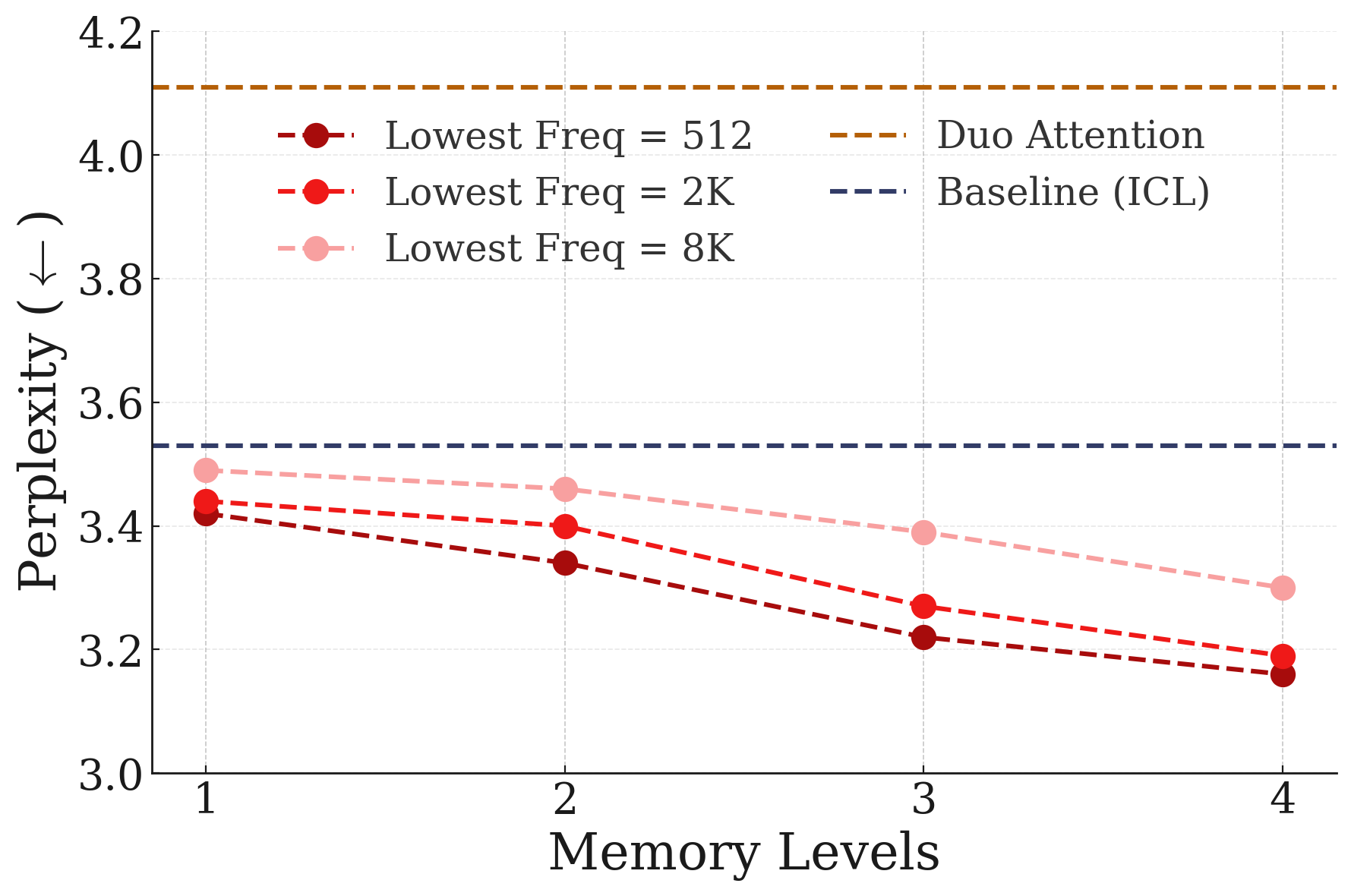}
    \caption{The effect of memory levels on the in-context learning performance of the model in (Left) MK-NIAH of RULER~\citep{hsieh2024ruler}, (Middle) LongHealth~\citep{adams2025longhealth}, (Right) QASPER~\citep{dasigi2021dataset} benchmarks. Note that for QASPER benchmark (Right), the lower values indicate better performance.}
    \label{fig:effect-level}
\end{figure*}

\subsection{\model: Continual Learning and Long Context Understanding}
One of the main goals of NL is to enhance the continual learning capabilities, and so in this section, we evaluate NL and its implications such as Continuum Memory System (CMS) and \model{} on multiple continual learning and long context understanding tasks. For each of the tasks, we use the best reported results on the benchmark as the baselines.

\head{Class Incremental Learning}
First, we focus on class-incremental learning tasks on three datasets: 
\begin{itemize}
    \item CLINC~\citep{larson2019evaluation}: CLINC is a multi-domain intent classification benchmark designed for task-oriented dialog systems, with a special focus on detecting out-of-scope (OOS) queries. It has 150 in-scope intent classes spanning 10 broad domains (e.g. Banking, Travel, Home, Weather, Small Talk, etc.) with 23.7K total queries, of which 22.5K are in-scope and 1.2K are out-of-scope. 
    \item Banking~\citep{casanueva2020efficient}: Banking dataset is a single-domain intent classification benchmark focused on fine-grained customer service queries in the banking domain. In this dataset, each example is a short customer query (e.g. “How can I reset my card PIN?”) that must be classified into the correct banking intent/category. There are 3083 total examples labeled with the 77 intents (heavily imbalanced classes).
    \item DBpedia~\citep{auer2007dbpedia}: DBpedia is a text classification benchmark from Wikipedia where article abstracts are expected to be categorized into ontology topic classes. In other words, given a short Wikipedia description, the goal is to predict its high-level topic/category (such as whether the article is about a book, a film, an animal, a place, etc.). The dataset has over 340K examples labeled across those 70 second-level classes, but we sample 10K training and 1K test instances for the 70-class DBpedia task.
\end{itemize}
As the backbone of our \model{} models we use Llama3-8B and Llama-3B~\citep{dubey2024llama}, and then employ our technique discussed in \autoref{sec:pre-trained-mlp} to make the MLP blocks capable of adaption, placing them in different levels with different frequency updates followed by continual pre-training with 15B tokens. Following \citet{momeni2025context}, we use simple in-context learning (ICL) capability of Llama-3 models (with the same process of continual pre-training with 15B tokens but without any change in the MLP blocks), Elastic Weight Consolidation (EWC)~\citep{kirkpatrick2017overcoming}, and In-context
Continual Learning with an External Learner (InCA)~\citep{momeni2025context}  as the baselines of our evaluation. The results are reported in \autoref{fig:CIL}. \model{} shows the best performance across all continual learning baselines, including models with external learner (i.e., InCA). Comparing \model{} with ICL, the main difference comes from \model's multiple levels of in-context learning (or equivalently, different frequency of updates for MLP blocks), indicating the effectiveness of CMS's design for enhancing continual learning capabilities. Furthermore, the superior performance of \model{}, compared to InCA and EWC, indicates that the knowledge transfer between levels plays a critical role in the performance of the model.

\head{The Effect of Levels on In-context Learning}
Despite showing improvement when using CMS in the above tasks, to better understand and evaluate the effect of levels and their frequency on the in-context level ability of the model, we perform in-context question/answering and multi-key long context understanding. More specifically, we use:
\begin{itemize}
    \item LongHealth~\citep{adams2025longhealth}: This is a benchmark for long-context clinical question answering with multiple-choice QA tasks based on extensive fictional patient records, testing an LLM’s ability to extract and reason over detailed medical documents. The dataset includes 20 comprehensive patient case documents (across various diseases), each about 5.1K–6.8K words in length, and we use 200 questions sampled from patient records. 
    \item QASPER~\citep{dasigi2021dataset}: This benchmark is an information-seeking QA dataset centered on full-length NLP research papers. In particular, it contains around 5K QA pairs grounded in around 1.6L NLP research papers. Also, we use the full text of each paper as the context for the model. 
    \item MK-NIAH~\citep{hsieh2024ruler}: We use the task of multiple keys in needle-in-haystack from RULER~\citep{hsieh2024ruler}. This setup requires models to not only locate but also extract multiple pieces of information distributed throughout a long text. 
\end{itemize}
As for the baseline, we use ICL, which is the same as \model{} with 1-level of memory, and also DuoAttention~\citep{xiao2025duoattention}. It is notable that methods such as Cartridges~\citep{eyuboglu2025cartridges} have shown promising performance, even better than DuoAttention and sometimes ICL. Here, however, we exclude their comparison with \model{} mainly due to the fact that \model{} has higher memory usage and there are fundamental differences in their computational costs (e.g., self-studying, etc.), requiring further experiments with careful and controlled design in the future studies. For the variants of our model, we use different number of memory levels with different frequencies, which are grouped based on the lowest frequency. Note that the lowest-frequency memory corresponds to the most persistent memory of the model and so we expect models with higher lowest frequency to be more adaptive.

\begin{table*}
    \begin{minipage}[b]{0.6\linewidth}
        \centering
    \small
    \captionof{table}{
    Needle-In-A-Haystack experiments with: (1) Single needle with three levels of difficulty: single-needle tasks—S-NIAH-1 (passkey retrieval), S-NIAH-2 (numerical needle), and S-NIAH-3 (UUID-based needle); (2) multi-query; (3) multi-key; and (4) multi-value settings of the benchmark.} \label{tab:ruler}
    \hspace{3ex}
    \resizebox{\linewidth}{!}{
    \begin{tabular}{lccccccccc}
    \toprule
     & \multicolumn{3}{c}{\textbf{S-NIAH-1}}  & \multicolumn{3}{c}{\textbf{S-NIAH-2}} & \multicolumn{3}{c}{\textbf{S-NIAH-3}}  \\
    & \multicolumn{3}{c}{(pass-key retrieval)} & \multicolumn{3}{c}{(number in haystack)} & \multicolumn{3}{c}{(uuid in haystack)}  \\
    \cmidrule(lr){2-4} \cmidrule(lr){5-7} \cmidrule(lr){8-10}
    \textbf{Model}  & 4K  & 8K & 16K & 4K & 8K & 16K & 4K & 8K & 16K  \\
    \midrule
    Transformer          &  88.6 & 76.4 & 79.8 & 100 & 98.8 & 94.2 & 78.0 & 69.2 & 40.8 \\
    \rowcolor{mygray} \model-Attention & 100 & 100 & 100 & 100 & 98.4 & 94.4 & 76.8 & 68.8 & 42.4 \\
    \midrule
    RWKV-7 & 100 & 100 & 99.6 & 93.8 & 44.8 & 12.6 & 63.8 & 13.2 & 5.8\\
    Comba & 100 & 100 & 99.4  & 92.6 & 47.2 & 13.4 & 62.4 & 13.8 & 7.4\\
    DLA             & 96.4 & 71.2 & 44.0 & 79.6 & 42.6 & 28.2 & 18.2 & 8.8 & 4.0 \\
    Titans      & 100 & 100 & 100 & 99.6 & 84.6 & 75.4 & 74.2 & 42.8 & 21.2 \\
    \midrule
    \rowcolor{mygray}\model & 100 & 100 & 100 & 99.2 & 88.4 & 78.2 & 73.2 & 46.2 & 24.8 \\
    \toprule
    \addlinespace
    & \multicolumn{3}{c}{\textbf{MK-NIAH-1}} & \multicolumn{3}{c}{\textbf{MQ-NIAH}}& \multicolumn{3}{c}{\textbf{MV-NIAH}} \\
    & \multicolumn{3}{c}{(multi-key line retrieval)} & \multicolumn{3}{c}{(multi-query)} & \multicolumn{3}{c}{(multi-value)}  \\
    \cmidrule(lr){2-4} \cmidrule(lr){5-7} \cmidrule(lr){8-10}
    \textbf{Model}  & 4K & 8K & 16K & 4K & 8K & 16K & 4K & 8K & 16K  \\
    \midrule
    Transformer      & 79.4 & 83.0 & 61.4 & 58.9 & 48.0 & 29.8 & 37.5 & 34.1 & 21.5 \\
    \rowcolor{mygray} \model-Attention & 80.2 & 84.8 & 60.8 & 60.4 & 47.8 & 30.6 & 35.2 & 34.4 & 24.8 \\
    \midrule
    RWKV-7 & 21.4   & 18.8  &  9.6    &  20.4  & 14.8  &  8.6  & 16.2 & 13.4 & 6.8\\
    Comba &  21.4   & 19.4  &   8.2   &  21.8  & 15.2  &  6.4  & 16.5 & 13.5   & 7.2 \\
    DLA             & 27.4 & 20.0 & 11.8 & 26.4 & 22.0 & 6.4 & 25.6 & 12.8 & 9.6 \\
    Titans       & 26.4 & 23.6 & 8.2 & 22.8 & 19.8 & 9.4 & 24.6 & 15.1 & 8.2 \\
    \midrule
    \rowcolor{mygray}\model & 29.4 & 24.8 &  14.8 & 31.7  & 24.8 & 14.2 & 31.4 & 17.2 & 11.4 \\
    \bottomrule
    \end{tabular}
    }
    \centering
    \end{minipage}~\hspace{1ex}~
    \begin{minipage}{0.37\linewidth}
            \begin{minipage}{\linewidth}
                \centering
                \includegraphics[width=\linewidth]{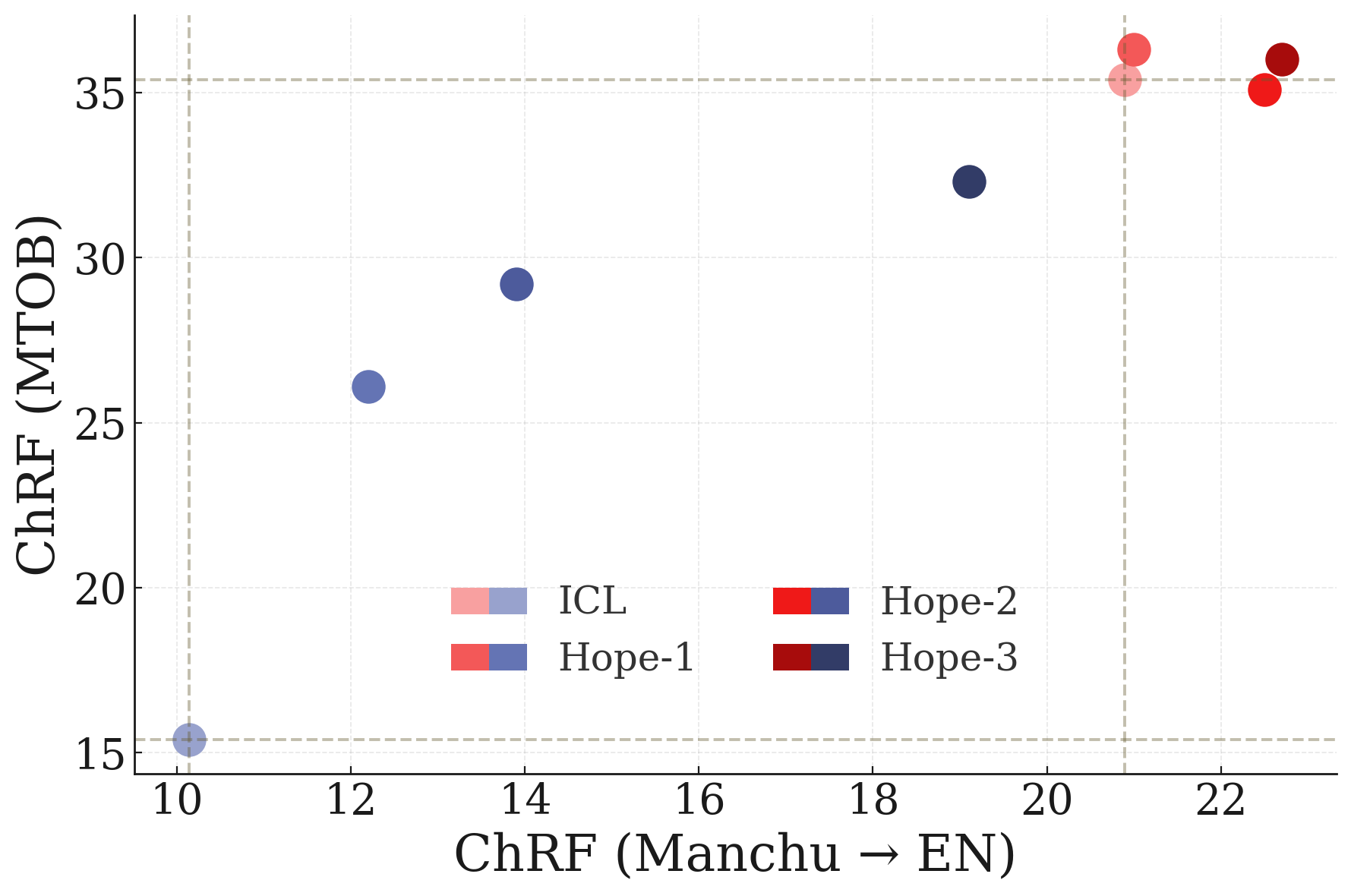}
                \captionof{figure}{Continual Translation of a Novel Language (CTNL) task. \textcolor{darkred}{Red} points are the results with only one language.  \textcolor{darkblue}{Blue} points are the results in the continual leanring setup.}
                \label{fig:icl-translate}
            \end{minipage}
            \vspace*{10ex}
            \begin{minipage}{\linewidth}
                \centering
                \vspace{1ex}
                \includegraphics[width=\linewidth]{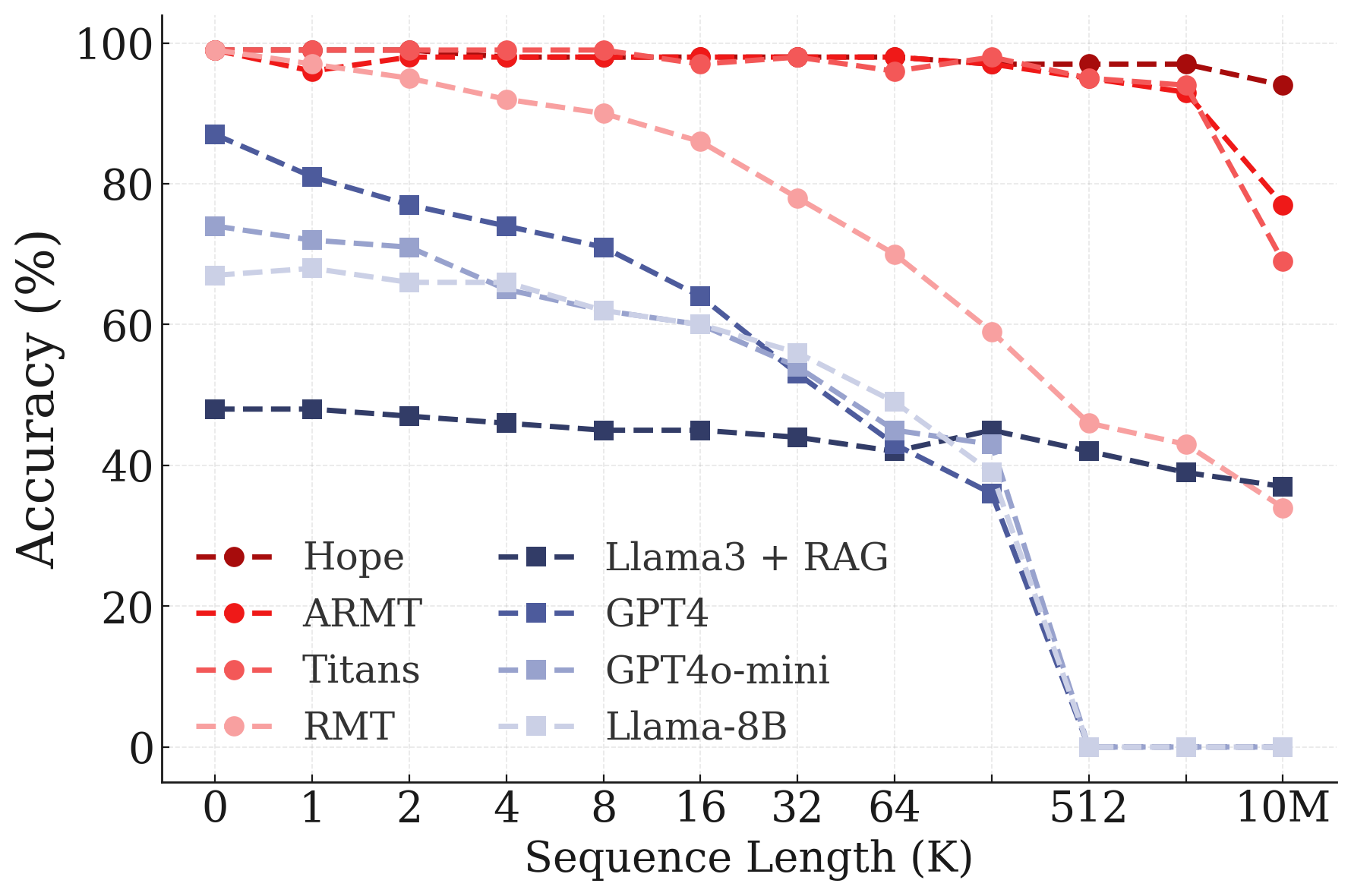}
                \captionof{figure}{BABILong Benchmark. \textcolor{darkred}{Red} points are the results of fine-tuned models, and \textcolor{darkblue}{Blue} points are the large models' zero-shot results. }
                \label{fig:babilong}
                \vspace{-7ex}
            \end{minipage}
    \end{minipage}
\end{table*}

The results are reported in \autoref{fig:effect-level}. \model{} with any number of levels and also with any lowest frequency outperforms both ICL baseline as well as the efficient DuoAttention. Furthermore, comparing the \model's variants with themselves, the results suggest that: (1) The more levels of memory can help with in-context learning capability of the model and enhances its long-term memory and so long-context understanding;  and (2) The higher the lowest-frequency the lower the performance. As discussed above, the lowest-frequency memory is the most persistent memory of the model and so increasing it means that the model has a weaker but more adaptive long-term memory. Due to the fact that the frequency of update directly affects the efficiency of the model, one might find ``Lowest Frequency = 2K'' an optimal setup as it offers significantly more efficient forward pass, while achieves close performance with ``Lowest Frequency = 512''.

\head{Learning a New Language In-Context}
Throughout the paper, we argued that pre-training can be seen as an in-context learning process, where the context is the entire pre-training data. Accordingly, when we use more levels in the neural learning module, or performing in-context learning on different context flows, we expect the model to show better adaptability and continual learning capabilities. To this end, with combining two existing benchmarks of MTOB~\citep{tanzer2024a} and Manchu~\citep{pei2025understanding}, we design a new continual learning task that the LLM learns two new languages in context and is expected to translate phrases to English. 
Then for our Continual Translation of a Novel Language (CTNL) task, we consider two setups: (1) Learning and testing the performance of the model on each language separately (in \textcolor{darkred}{red} color). This is a baseline we use to measure the catastrophic forgetting when comparing with the next setting; and (2) The model, first, learns these two languages in a sequential manner (in \textcolor{darkblue}{blue} color) and then is asked to translate the phrases to English. We use ICL as the baseline, and to study the importance of multi-level design of \model, we use different variants of \model{} with 1, 2, and 3 additional levels of memory, which we refer to as \model-1, \model-2, and \model-3, respectively.

The results are reported in \autoref{fig:icl-translate}. More specifically, each point is the performance of one model in one of the setups such that the $\vx$-axis (resp. $\vy$-axis) indicates the model ChRF in translation of Manchu $\rightarrow$ English (resp. Kalamang $\rightarrow$ English). In the first setup (in-context translation without continual learning), all \model's variants performs better or on-par compared to ICL, supporting the importance of CMS design. In the second setup (i.e., continual translation), however, ICL faces dramatic performance drop and almost rely on its capabilities achieved in its pre-training (i.e., catastrophic forgetting about the knowledge in the context). On the other hand, increasing the memory levels in the \model{} shows clear improvement and \model-3 almost recovers the ICL capability in the first setup, without continual learning. These results further support the importance of CMS design in continual learning and the ability of the model to adapt itself to the new tasks.

\subsection{\model: Long Context Understanding}\label{sec:exp-niah}
In the previous section, we evaluated the performance of \model-Attention, when the MLP blocks are adapted to perform in-context learning. In this part, we evaluate the performance of \model{} in long context understanding, when it starts learning from scratch. To this end, we use about 50B tokens from a mixture of FineWeb-Edu~\citep{penedo2024fineweb} and long-context documents with a vocabulary size of 32K to train all the models from scratch. All models are optimized using AdamW with tuned learning rate for each model and with the default optimizer configuration in \citet{behrouz2024titans}. We focus on two popular benchmarks of RULER~\citep{hsieh2024ruler} and BABILong~\citep{kuratov2024babilong}:

\head{Needle-in-a-Haystack (NIAH) Tasks}
In the first part, we focus on the needle-in-a-haystack with different setups of: (1) single needle but different types (i.e., pass-key, number, and uuid), (2) multi-key, (3) multi-query, and (4) multi-value, all follows \citet{hsieh2024ruler}. As for the baselines, we use RetNet~\citep{sun2023retentive} and DeltaNet~\citep{schlag2021linear} as the representative of the models \emph{purely} based on Hebbian- and Delta-rule, and experimented with diverse set of modern \emph{linear} recurrent models and so used the linear models with the best performance: i.e., RWKV-7~\citep{rwkv-repo} and Comba~\citep{hu2025improving}. As another group of baselines, we also compare with \emph{deep} memory modules with dot-product and $L_2$ regression objectives: i.e., DLA~\citep{behrouz2025atlas}, and Titans~\citep{behrouz2024titans}.

The results are reported in \autoref{tab:ruler}. Comparing with other attention-free models, \model{} achieves the best performance across all tasks and levels of difficulties. Particularly, comparing to linear memories, deep memory modules show better performance in longer sequences, mainly due to their higher memory capacity to compress more tokens. Comparing \model{} with Titans, the superior performance of \model, specifically in longer context length supports the importance of both self-referential update as well as the CMS design. Finally, we also evaluate the performance of \model-Attention and compare it with Transformers to better understand the contribution of CMS design. The results indicate that \model-Attention achieves a better performance compared to Transformers, supporting the advantage of having CMS in \model-Attention design.

\head{BABILong} Next, we evaluate the \model's performance on BABILong benchmark~\citep{kuratov2024babilong} and compare it with (1) large models such as GPT4 and GPT4o-mini~\citep{achiam2023gpt}; (2) middle-size Llama-8B model~\citep{dubey2024llama} with its RAG augmented version; and (3) the state-of-the-art small models in this tasks: i.e., RMT~\citep{bulatov2022recurrent}, ARMT~\citep{rodkin2024associative}, and Titans~\citep{behrouz2024titans}. We follow the original setup of the benchmark and fine-tune the small models with the same process as \citet{kuratov2024babilong}.

The results are reported in \autoref{fig:babilong}. Large models show significant performance drop with increasing the sequence length, where all fail around 128K-256K context length. The RAG augmented model also show a drop with increasing the context, but it is able to relatively maintain its performance after 256K context length. Among fine-tuned models, Titans, ARMT, and \model{} show competitive results until 1M context length, but the performance of both Titans and ARMT drop fast after that point. \model{} maintains its good performance even for 10M context length, mainly due to its CMS design. It is notable that the performance of all small models, including \model, can drop significantly when used without fine-tuning. The reason is, compressing large context (e.g., 10M), in addition to a powerful memory management in the high-frequency level, requires enough capacity to compress 10M tokens or at least tokens that are needed for the final answer. The fine-tuning step helps the models to adjust their lower-frequency levels to adapt fast and so properly manage their memory in the higher-frequency levels.

\begin{table*}[t!]
\centering
\caption{
Performance of models on language modeling and common-sense reasoning tasks.
}\label{tab:lm_results}
\centering
\resizebox{0.85\linewidth}{!}{
\centering
\begin{tabular}{l|c c|c c c c c c c c c}
\toprule
\textbf{Model}  & \textbf{Wiki.}  &  \textbf{LMB.} &  \textbf{LMB.} & \textbf{PIQA} &    \textbf{Hella.} & \textbf{Wino.} & \textbf{ARC-e} &  \textbf{ARC-c} &  \textbf{SIQA}  & \textbf{BoolQ} &  \textbf{Avg.} \\
 & ppl $\downarrow$  &  ppl $\downarrow$  &  acc $\uparrow$  & acc $\uparrow$ &   acc\_n $\uparrow$  & acc $\uparrow$  & acc $\uparrow$ & acc\_n $\uparrow$ &  acc $\uparrow$  & acc $\uparrow$ &   $\uparrow$  \\
\midrule
\midrule
\multicolumn{12}{c}{760M params / 30B tokens} \\
\midrule
 Transformer++ & 24.18 & 24.27 & 37.1 & 67.2 & 43.8 &	53.0 & 65.6	& 33.4 &  39.1  & 61.7  & 50.11 \\
 Samba$^{*}$ & 21.07 & 22.85 & 39.2 & 68.9 & 47.8 &	53.1 & 65.8	& 34.9 &  38.9  & 63.1  & 51.46 \\
 \midrule
 RetNet & 25.77 & 24.19 & 34.5 & 66.8 & 41.2 &	51.9 & 63.6	& 32.5 &  38.8  & 56.2  &  48.19\\
 DeltaNet & 24.52 & 24.38 & 36.8 & 67.3 & 44.5 &	51.8 & 64.2	& 32.7 &  39.6  & 60.1  & 49.63 \\
 RWKV-7 & 23.75 & 23.08 & 37.1 & 67.3 & 47.6 & 52.2 & 64.7 & 34.2 & 39.4 & 61.9 & 50.55\\
 Comba & 22.41 & 22.19 & 37.5 & 66.9 & 48.2 & 52.4 & 65.1 & 34.1 & 40.1 & 62.8 & 50.89 \\
  \midrule
   TTT & 24.17 & 23.51 & 34.7 & 67.3 & 43.9 & 51.0 & 64.5 & 33.8 & {40.2} & 59.6 & 47.32 \\
 Miras (Memora) &  22.28 & 22.31 & 38.2 & 67.8  & 49.3 & 53.3 & 63.6 & 36.1 & 40.9 & 63.0 & 51.53\\
 DLA &  23.12 & 22.09  &  36.1   &  68.0 &  47.9 &  52.7 & 65.8  & 34.6 & 39.1 & 59.6 &  50.48\\
Titans & 20.08 & 21.52 & 38.1 & 69.1  & 48.5 & 52.7 & 66.2 & 35.7 & 40.3 & 62.8 & 51.68\\
\midrule
\rowcolor{mygray} \model & 18.68 & 20.07 & 38.8 & 69.2  & 49.1 & 53.6 & 66.8 & 36.1 & 41.2 & 63.4 & 52.28\\
  \toprule
\multicolumn{12}{c}{1.3B params / 100B tokens} \\
\midrule
 Transformer++ & 17.92 & 17.73 & 42.6 & 71.4 & 52.3 &	54.1 & 69.9	& 36.5 &  41.8  & 58.4  & 53.38\\
 Samba$^*$ & 16.15 & 13.21 & 45.2 & 71.5 & 53.8 &	55.8 & 69.1	& 36.7 &  40.6  &  {63.0}  & 54.46 \\
 \midrule
 RetNet & 18.91 & 17.04 & 41.2 & 71.3 & 49.1 &	55.2 & 67.5	& 34.1 &  {41.4}  & {61.0}  & 52.60 \\
 DeltaNet & 18.62 & 17.10 & 41.6 & 70.1 & 49.4 &	52.7 & 67.6	& 35.2 &  39.7  & 54.8  & 51.39 \\
 RWKV-7 & 18.44 & 15.96 & 46.7 & 72.4 & 54.9 & 57.5 & 71.6 & 38.2 & 40.7 & 60.4 & 55.30\\
 Comba & 18.16 & 14.87 & 46.9 & 73.1 & 54.5 & 57.7 & 72.0 & 39.1 & 40.2 & 60.6 & 55.39 \\
\midrule
TTT & 18.42 & 14.51 &  46.8 & 72.9  & 55.2 &  59.0 & 71.8 & 39.5 &  39.8 & 59.6 & 55.58 \\
 Miras (Memora) &  15.90 & 12.04  &  48.7   &  73.1 &  56.0 &  57.4 & 71.5  & 37.9 & 40.2 & 61.3 &  55.76\\
DLA & 16.31 & 12.29 & 44.5   & 70.6 & 53.9 & 54.2 & 69.6 & 36.0 & 40.8  & 60.2 & 53.72\\
Titans & 15.60 & 11.41 &  49.1 & 73.1  & 56.3 &  59.8 & 72.4 & 40.8 &  42.1 & 61.0 & 56.82\\
\midrule
\rowcolor{mygray} \model & 14.39 & 10.08 &  51.0 & 73.9 & 57.5 & 61.2 & 73.8 & 42.7 & 42.8 & 61.4 & 58.04 \\
\bottomrule
\multicolumn{5}{l}{$^*$ is a hybrid of attention + linear RNN~\citep{ren2024samba}.}
\end{tabular}
}
\end{table*}

\subsection{\model: Language Modeling and Common-Sense Reasoning}\label{sec:exp-lm}
In this section, we aim to study \model{} as a backbone of a language model and evaluate it on common language modeling and common-sense reasoning tasks with the setup of: 
\begin{itemize}
    \item Datasets: We evaluate \model{} and baselines on Wikitext~\citep{merity2017pointer}, LMB~\citep{paperno-etal-2016-lambada}, PIQA~\citep{bisk2020piqa}, HellaSwag~\citep{zellers-etal-2019-hellaswag}, WinoGrande~\citep{sakaguchi2021winogrande},  ARC-easy (ARC-e) and ARC-challenge (ARC-c)~\citep{clark2018think}, SIQA~\citep{sap-etal-2019-social}, and BoolQ~\citep{clark-etal-2019-boolq} benchmarks.
    \item {Baselines}: As for the baselines, similar to \autoref{sec:exp-niah}, we use RetNet~\citep{sun2023retentive} and DeltaNet~\citep{schlag2021linear} as the representatives of the models that are \emph{purely} based on Hebbian- or Delta-rule, and two modern \emph{matrix-valued} recurrent models with the best performance compared to others: i.e., RWKV-7~\citep{rwkv-repo} and Comba~\citep{hu2025improving}. As another group of baselines, we compare with attention-free \emph{deep} memory modules with diverse internal attentional bias of dot-product, $L_2$, and $L_p$ regression: i.e., TTT~\citep{sun2024learning}, Miras~\citep{behrouz2025Miras}, DLA~\citep{behrouz2025atlas} and Titans~\citep{behrouz2024titans}. Finally, we also compare with Transformers~\citep{transformers, dubey2024llama} as well as the hybrid of attention and linear RNN, Samba~\citep{ren2024samba}. 
    \item Training: We train models with about 760M and 1.3B parameters, trained with 30B and 100B tokens, respectively, from a mixture of FineWeb-Edu~\citep{penedo2024fineweb} and long-context documents with a vocabulary size of 32K to train all the models from scratch. All models are optimized using AdamW with tuned learning rate for each model and with the default optimizer configuration in \citet{behrouz2024titans}.
\end{itemize}
The results are reported in \autoref{tab:lm_results}. \model{} outperforms all the baselines on the average performance in both language modeling and common-sense reasoning benchmarks. Interestingly, with scaling the parameters, \model{} show higher performance gain compare to other attention-free models.

\subsection{\model: In-context Recall Tasks and MAD Synthetic Benchmark}\label{sec:retrieval-tasks}
In-context recall is often referred to as one of the challenging benchmarks for attention-free models. In this section, we follow \citet{arora2024simple} and perform experiments on SWDE~\citep{lockard2019openceres}, NQ~\citep{kwiatkowski2019natural}, DROP~\citep{dua2019drop}, FDA~\citep{arora2023language}, SQUAD~\citep{rajpurkar2016squad}, and TQA~\citep{kembhavi2017you} to evaluate the effectiveness of \model's design. We use the same set of baselines and experimental setup as the above previous section. The results are reported in \autoref{tab:recal}. Transformers achieve the best performance, while \model{} show competitive results, outperforming all the attention-free baselines and closing the gap with Transformers.

\begin{minipage}[t]{0.48\textwidth}
    \centering
    \captionof{table}{The performance of \model{} and baselines in short in-context recall tasks. \model{} outperforms all attention-free models and close the gap with Transformers.}
    \label{tab:recal}
    \resizebox{\linewidth}{!}{
    \begin{tabular}{l c c c c c c }
    \toprule
         &  \multirow{1}{*}{SWDE} & \multirow{1}{*}{NQ} & \multirow{1}{*}{DROP} & FDA & SQUAD & \multirow{1}{*}{TQA} \\
         \midrule
         \midrule
        Transformers & 71.4 & 22.0 & 23.9 & 67.3 & 39.4 & 59.1 \\
        RWKV-7 & 52.3 & 17.8 & 21.7 & 32.8 & 28.5 & 56.2 \\
        Comba & 53.9 & 19.1 & 21.9 & 35.5 & 30.2 & 56.4 \\
        Titans & 60.8 & 20.3 & 22.0 & 37.6 &  31.8 & 57.5 \\
                 \midrule
        \rowcolor{mygray} \model{} & 65.9 &  21.2 & 22.8 & 41.9 & 33.0 & 57.7 \\
    \toprule
    \end{tabular}
    }
\end{minipage}~\hspace{1ex}~
\begin{minipage}[t]{0.51\textwidth}
    \centering
    \captionof{table}{Performance of \model{} and baselines on the synthetic benchmark of MAD~\citep{poli2024mechanistic}. \model{} outperforms all the baselines, including Transformers.}
    \label{tab:MAD}
    \resizebox{\linewidth}{!}{
    \begin{tabular}{l c c c c c}
    \toprule
         &  \multirow{2}{*}{Compress.} & \multirow{2}{*}{ICR} & \multirow{2}{*}{Fuzzy ICR} & Selective & \multirow{2}{*}{Memory} \\
         &  & & & Copying \\
         \midrule
         \midrule
        Transformers & 49.4 & 100 & 47.9 & 96.2 & 83.7\\
         RWKV-7 & 45.1 & 100 & 32.8 & 95.6 & 82.2\\
         Comba & 46.3 & 100 & 32.8 & 96.4 & 82.9\\
         Titans & 49.8 & 100 & 50.0 & 99.4 & 83.4 \\
         \midrule
        \rowcolor{mygray} \model & 51.2 & 100 & 52.1 & 99.7 & 85.2\\
    \toprule
    \end{tabular}
    }
\end{minipage}

We also study the performance of \model{} on MAD benchmark~\citep{poli2024mechanistic}, which is a synthetic benchmark, evaluating the performance of models in recall, memorization, compression, and copying tasks. The results are reported in \autoref{tab:MAD}. \model{} achieves the best results compared to baselines.

\begin{table*}[t]
    \centering
    \small
    \caption{Accuracies of various models on the formal language recognition tasks.}
    \label{tab:formal-language}
    \resizebox{0.95\linewidth}{!}{
        \begin{tabular}{l c rrrrrrrrrrrrr}
        \toprule
         & & \multicolumn{6}{c}{Non-Star-Free Regular} & \multicolumn{6}{c}{\textcolor{black!40!white}{Counter}} \\  \cmidrule(r{.45em} l{.50em}){3-8}  \cmidrule(r{.45em} l{.50em}){9-14}
         & Parallel & \multicolumn{2}{c}{Parity} & \multicolumn{2}{c}{$(aa)^*$}  & \multicolumn{2}{c}{$(abab)^*$} & \multicolumn{2}{c}{\textcolor{black!40!white}{$a^nb^n$}} & \multicolumn{2}{c}{\textcolor{black!40!white}{$a^nb^nc^n$}} & \multicolumn{2}{c}{\textcolor{black!40!white}{Shuffle-2}}\\  \cmidrule(r{.45em} l{.50em}){3-8}  \cmidrule(r{.45em} l{.50em}){9-14}
        Model & Training & Bin0 & Bin1 & Bin0 & Bin1 & Bin0 & Bin1 & \textcolor{black!40!white}{Bin0} & \textcolor{black!40!white}{Bin1} & \textcolor{black!40!white}{Bin0} & \textcolor{black!40!white}{Bin1} & \textcolor{black!40!white}{Bin0} & \textcolor{black!40!white}{Bin1} 
         \\
        \midrule
        LSTM & \xmark & \textbf{100.0} & \textbf{100.0} & \textbf{100.0} & \textbf{100.0} & \textbf{100.0} & \textbf{100.0} & \textcolor{black!40!white}{\textbf{100.0}} & \textcolor{black!40!white}{\textbf{100.0}} & \textcolor{black!40!white}{\textbf{100.0}} & \textcolor{black!40!white}{\textbf{100.0}} & \textcolor{black!40!white}{\textbf{100.0}} & \textcolor{black!40!white}{\textbf{100.0}} \\
        Transformer & \checkmark & 46.4 & 0.0 & 0.0 & 0.0 & 0.0 & 0.0 & \textcolor{black!40!white}{\textbf{100.0}} & \textcolor{black!40!white}{\textbf{100.0}} & \textcolor{black!40!white}{\textbf{100.0}} & \textcolor{black!40!white}{\textbf{100.0}} & \textcolor{black!40!white}{\textbf{100.0}} & \textcolor{black!40!white}{\textbf{100.0}} \\ \midrule
        Linear & \checkmark & 78.1 & 0.0 & 0.0 & 0.0 & 0.0 & 0.0 & \textcolor{black!40!white}{\textbf{100.0}} & \textcolor{black!40!white}{\textbf{100.0}} & \textcolor{black!40!white}{\textbf{100.0}} & \textcolor{black!40!white}{\textbf{100.0}} & \textcolor{black!40!white}{\textbf{100.0}} & \textcolor{black!40!white}{\textbf{100.0}}  \\
        DeltaNet & \checkmark & 98.2 &  10.1 & 0.0 & 0.0 & 0.0 & 0.0 & \textcolor{black!40!white}{\textbf{100.0}} & \textcolor{black!40!white}{\textbf{100.0}} & \textcolor{black!40!white}{\textbf{100.0}} & \textcolor{black!40!white}{\textbf{100.0}} & \textcolor{black!40!white}{\textbf{100.0}} & \textcolor{black!40!white}{\textbf{100.0}} \\
        SRWM & \xmark & \textbf{100.0} & \textbf{100.0} & \textbf{100.0} & \textbf{100.0} & \textbf{100.0} & \textbf{100.0} & \textcolor{black!40!white}{\textbf{100.0}} & \textcolor{black!40!white}{\textbf{100.0}} & \textcolor{black!40!white}{\textbf{100.0}} & \textcolor{black!40!white}{\textbf{100.0}} & \textcolor{black!40!white}{\textbf{100.0}} & \textcolor{black!40!white}{\textbf{100.0}}  \\
        \midrule
          \rowcolor{mygray} \model & \checkmark & \textbf{100.0} & \textbf{100.0} & \textbf{100.0} & \textbf{100.0} & \textbf{100.0} & \textbf{100.0} & \textcolor{black!40!white}{\textbf{100.0}} & \textcolor{black!40!white}{\textbf{100.0}} & \textcolor{black!40!white}{\textbf{100.0}} & \textcolor{black!40!white}{\textbf{100.0}} & \textcolor{black!40!white}{\textbf{100.0}} & \textcolor{black!40!white}{\textbf{100.0}} \\
        \bottomrule
        \end{tabular}
}
\end{table*}

\subsection{Language Recognition Tasks}\label{sec:exp-language}
One of the critical limitations of Transformers is on non-parallelizable tasks, where the recurrence plays a significant rule in achieving proper performance. An example of such tasks is state tracking problem, where the model needs to track its state given a sequence of instructions (movement, etc.), and several studies have shown that Transformers both in theory and practice significantly underperforms models with non-linear recurrence~\citep{merrill2024the, grazzi2025unlocking}. In this section, we focus on formal language recognition tasks and follow the construction of the benchmark by \citet{irie2023practical}. The results are reported in \autoref{tab:formal-language}. \model{} achieves the perfect score on all the tasks, similar to other non-linear recurrent models; e.g., LSTM~\cite{LSTM} and SRWM~\citep{irie2022modern}. The main advantage of \model, however, is the fact that when it is needed, it has parallelizable training and so can scale to larger scales for language modeling tasks.

\begin{table*}
\begin{minipage}{0.6\linewidth}
    \centering
    \captionof{table}{Ablation Study on \model. All components of \model{} are positively contributing to its performance. }
    \label{tab:ablation}
    \resizebox{0.78\linewidth}{!}{
    \begin{tabular}{l c c}
    \toprule
    \multirow{2}{*}{Model}     & Language Modeling & Reasoning \\
    & ppl $\downarrow$ & acc $\uparrow$ \\
    \midrule
    \midrule
    \rowcolor{mygray} \model{}         &   12.24  &  58.1  \\ 
    \midrule
    w/o DGD                &    13.41     &      56.5     \\
    w/o Momentum           &   13.58  &  56.9  \\
    w/o weight decay            &  13.71   &   57.2  \\
    w/o CMS                     &    13.04      &     57.3     \\
    w/o inner-projection $\vk$  &   13.77  &   56.9  \\
    w/o inner-projection $\vv$  &   13.90  &  55.1   \\
    w/o inner-projection $\vq$  &   12.19  &  57.4   \\
    \toprule
    \end{tabular}
    }
\end{minipage}
~\hfill~
\begin{minipage}{0.37\linewidth}
        \centering
        \includegraphics[width=0.87\linewidth]{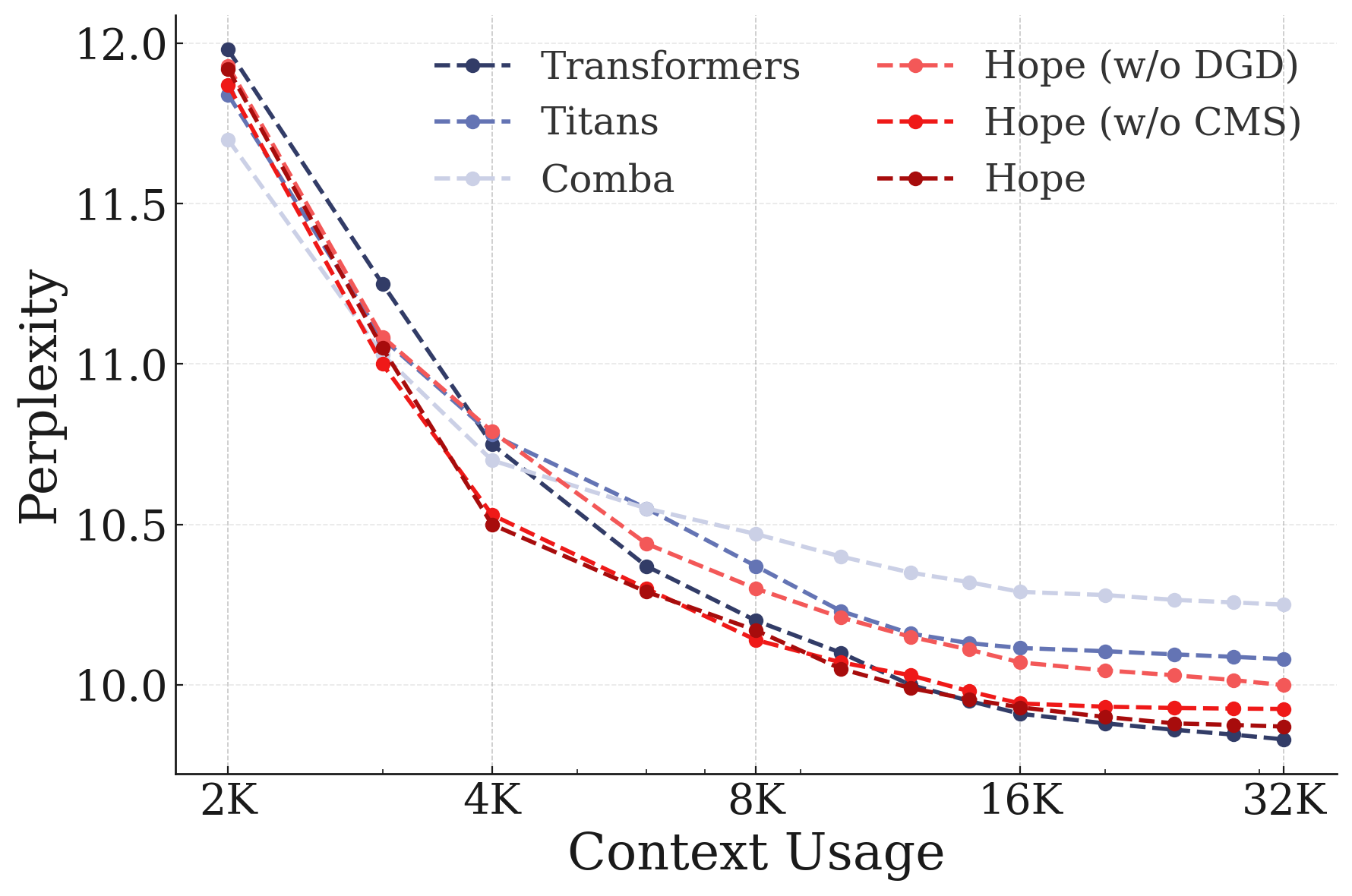}
        \captionof{figure}{The effect of context usage on the perplexity of the model. We expect the perplexity of a model with powerful memory management to decrease with more context.}
        \label{fig:placeholder}
\end{minipage}
\end{table*}

\subsection{\model: Ablation Studies and Scaling}
In this section, we first study the importance of design choices we have made for the \model{} architecture by performing ablation studies, where we remove or change one of the \model's components at a time. Note that, in the previous experiments, we have evaluated the significance of some components: E.g., \autoref{fig:icl-translate} and \autoref{fig:effect-level} have already been shown the effect of number of levels as well as the frequency of update on the performance of \model{} in continual learning. In this section, to compare different variants, we use the average perplexity of models in language modeling as well as the average accuracy in common-sense reasoning tasks. The results are reported in \autoref{tab:ablation}. (1) The first row, replaces Delta Gradient Descent with a simple gradient descent in the design of self-modifying Titans; (2) The second row removes the momentum term in the self-modifying Titans; (3) removes the weight decay; (4) removes the CMS from \model's architecture; (5, 6, 7) transfer the projections for $\vk, \vv, \vq$ from higher-frequency level to the lowest-frequency level. All the components of \model{} contribute to its superior performance in these tasks and removing or changing each of them can damage the model's perplexity in language modeling and/or accuracy in common-sense reasoning tasks.

\begin{table*}
\begin{minipage}{0.612\linewidth}
    \centering
        \includegraphics[width=0.5\linewidth]{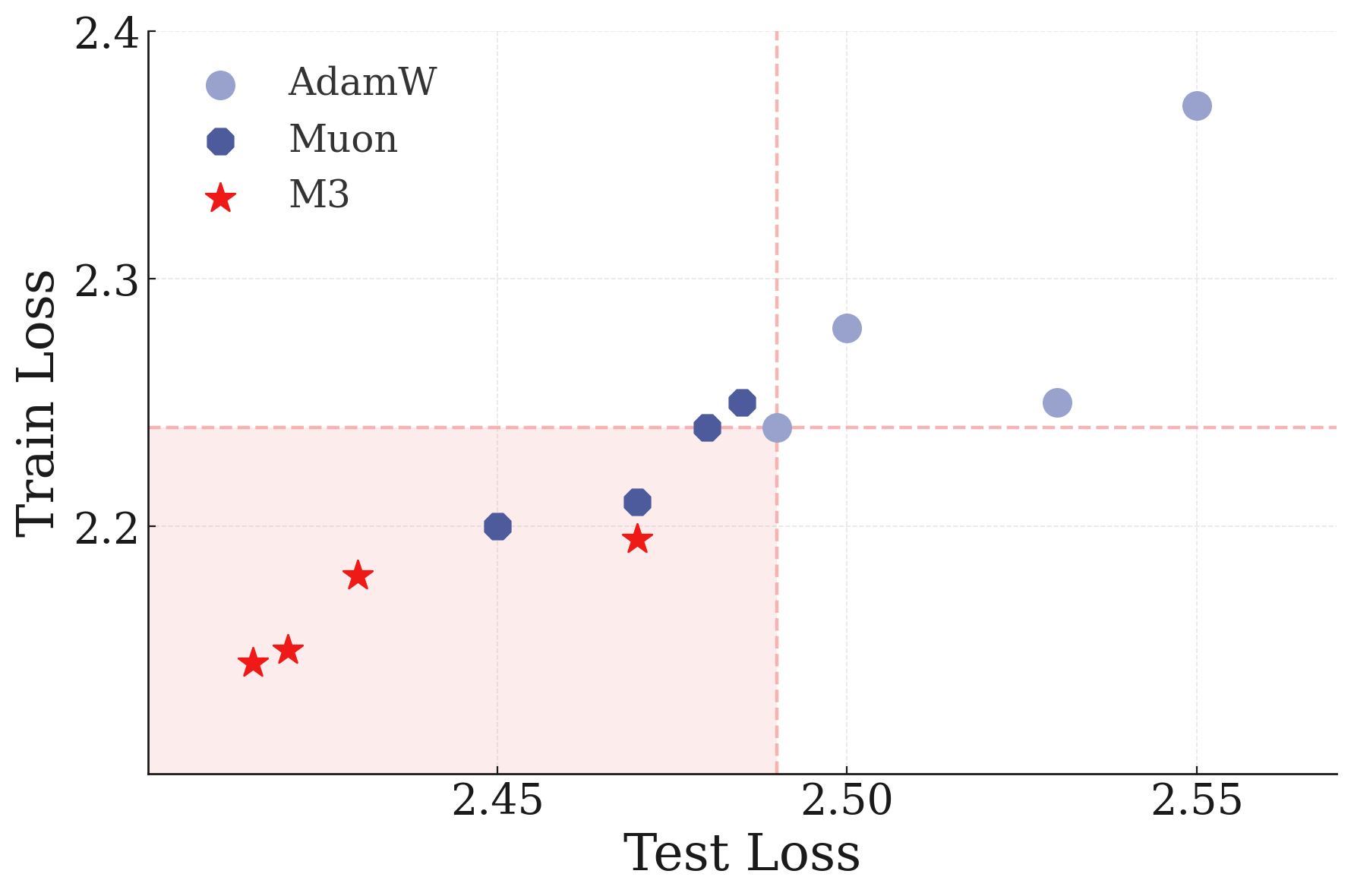}~\hfill~
        \includegraphics[width=0.5\linewidth]{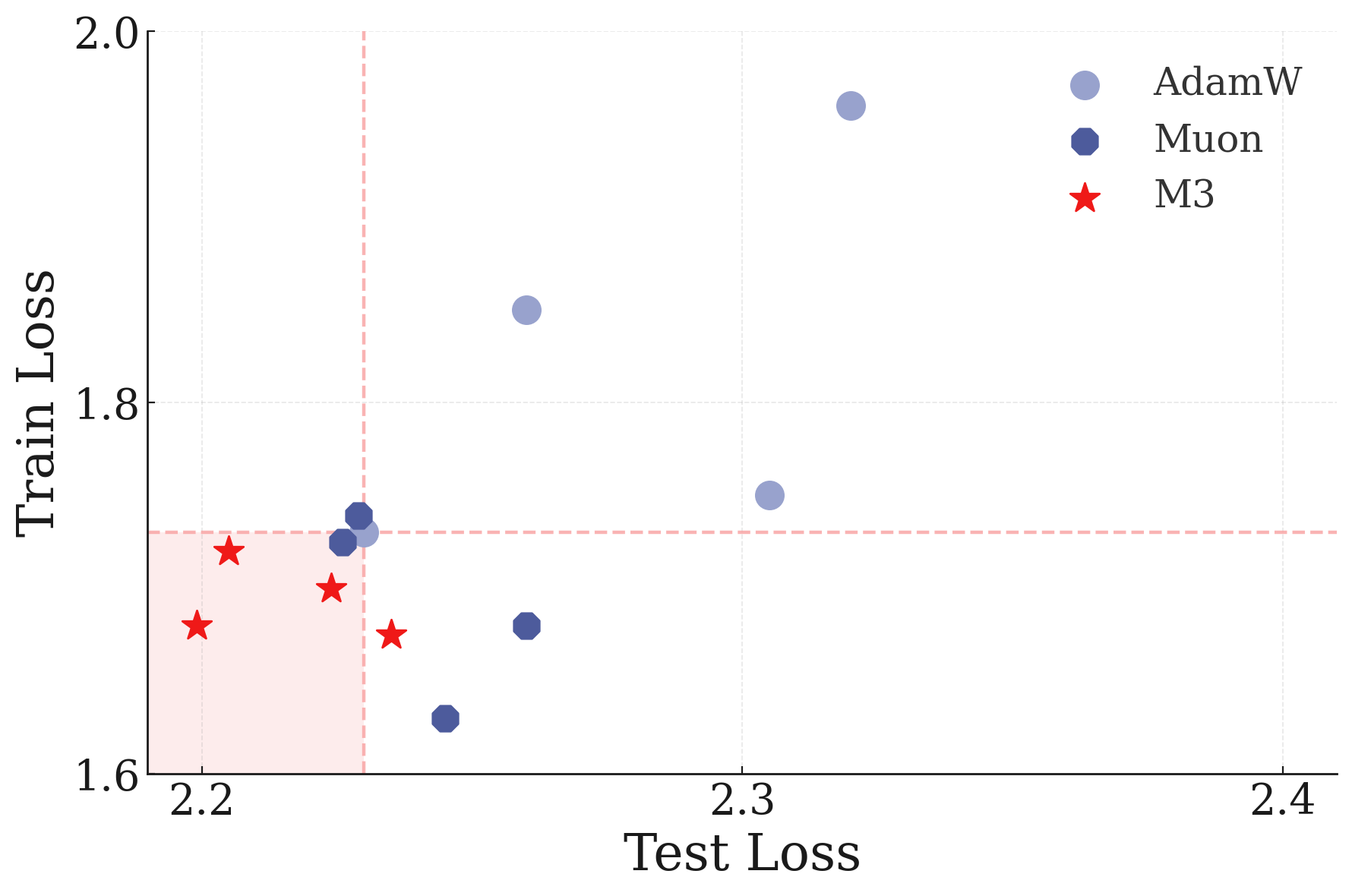}
        \captionof{figure}{ViT test and train loss on ImageNet-21K~\citep{ridnik2021imagenetk}, trained with AdamW, Muon, and our M3 optimizers. }
        \label{fig:optimizer-imagenet}
\end{minipage}~\hfill~
\begin{minipage}{0.33\linewidth}
        \centering
        \includegraphics[width=0.92\linewidth]{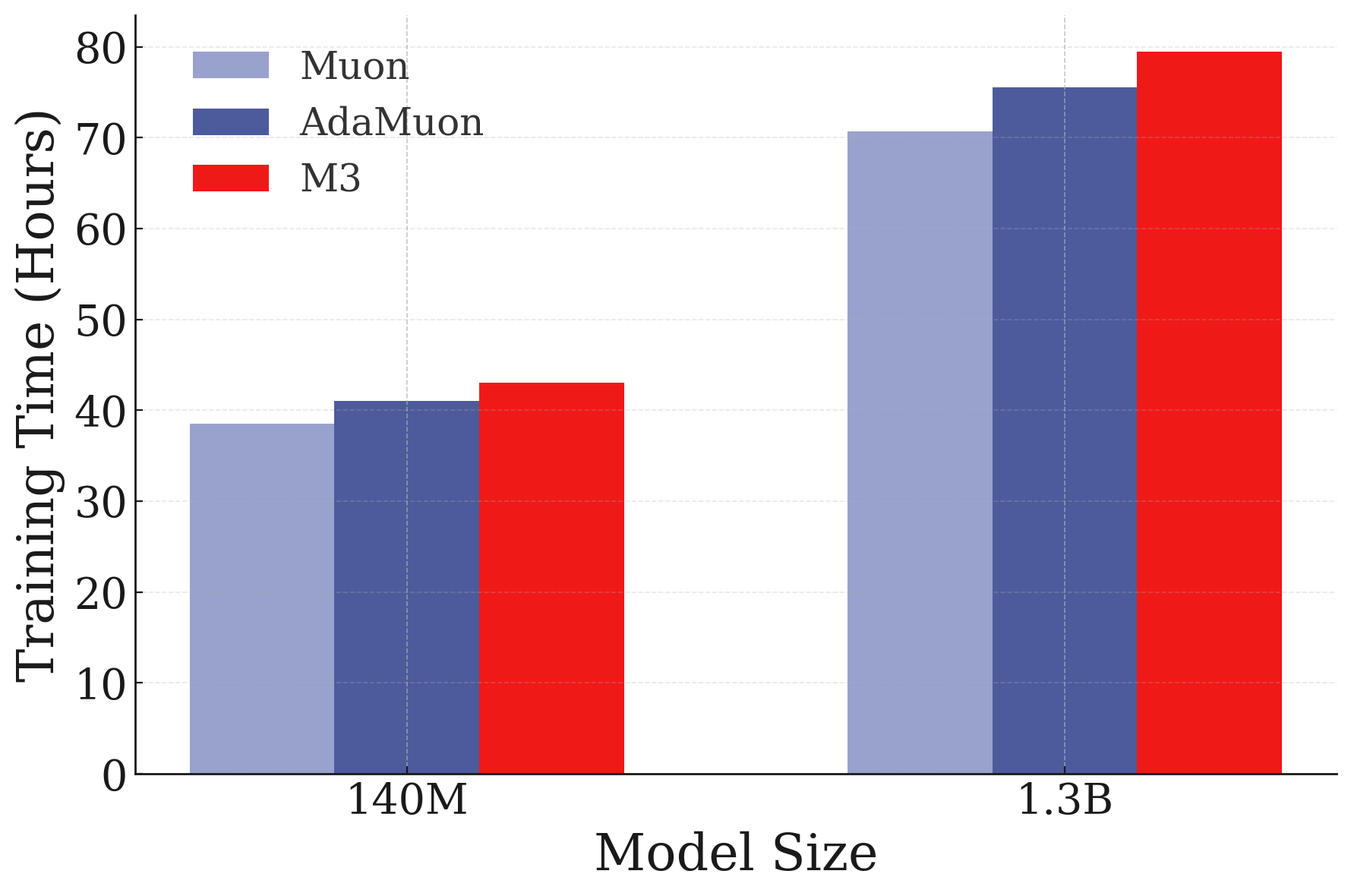}
        \captionof{figure}{Training time of models with 140M and 1.3B parameters with Muon, AdaMuon, and M3 optimizers. }
        \label{fig:optimizer-efficiency}
\end{minipage}
\end{table*}

\subsection{Expressive Optimizers}\label{sec:exp-deep-optimizer}
In this section, we evaluate the performance of our M3 optimizer from both aspects of effectiveness in finding the solution and also efficiency of training in large scales:

\head{ImageNet} In this experiment, we focus on ViT~\citep{dosovitskiy2021an} architecture for vision tasks and pre-train it on ImageNet-21K, which consists of 11M images corresponding to 10,450 classes. We use a patch size of 16, MLP dimension of 1536 and 3072 for the two scales of 24M and 86M models, respectively. We control all other components and finetuned hyperparameters for each optimizer separately to make sure a fair comparison. We vary the optimizer for training the ViT to understand which of the optimizers find more effective solutions in the same number of training steps. The training and test loss of trained models with different optimizers are reported in Figure~\ref{fig:optimizer-imagenet}. Our M3 shows the best training/test loss compared to both AdamW and Muon.

\head{Efficiency in Large Models} Due to the small size of the model that is needed for ImageNet, we also perform an efficiency comparison when training a language model. To this end, we use two scales of 140M and 1.3B and train the same Transformer model using different optimizers of Muon~\citep{jordanmuon}, AdaMuon~\citep{si2025adamuon}, and our M3. The results are reported in Figure~\ref{fig:optimizer-efficiency}. Our M3 optimizer, due to the use of multiple momentums (memories) is reletively slower compare to the Muon optimizer, and show on par efficiency with AdaMuon.

\section{Conclusion}\label{sec:concolusion}
In this paper, we introduced \emph{Nested Learning (NL)} a new learning paradigm, in which modern machine learning systems are modeled as inter-connected, multi-level optimization problems, each with its own context flow and update frequency. Within this view, both architectures and optimizers are instances of nested systems of associative memories that compress their own context (either is tokens, gradients, or higher-level signals) into internal parameters. This perspective reframes pre-training, in-context learning, and continual learning as manifestations of the same underlying mechanism: learning to compress and reuse context at different levels and time scales. This perspective recasts backpropagation, momentum, and preconditioning as associative memory mechanisms, and explains a wide family of existing methods as specific design points in a larger (previously hidden) space.

Building on NL's viewpoint, we derived generalized gradient-based updates: e.g., Delta Gradient Descent, Delta Momentum, Multi-scale Momentum Muon (M3), and reinterpreted modern sequence architectures nested associative memories. To enhance the memory processing, we introduced Continuum Memory System (CMS), a new formulation for memory that generalizes the traditional viewpoint of ``long-term/short-term memory blocks''. Our \model{} architecture based on a self-modifying Titans and CMS improves continual learning and long-context reasoning capabilitites, while remaining competitive as a general backbone.

\headdot{Is Catastrophic Forgetting Solved?} 
While \model{} and CMS have shown promising results in reducing catastrophic forgetting in the tasks we empirically studied, the undesirable phenomenon of catastrophic forgetting is not ``solved''  in general. From nested learning viewpoint on learning and backpropagation process, catastrophic forgetting is a natural consequence of compression, where the limited capacity of the network forces the model to forget so that it retains capacity for new information. We view NL as a roadmap rather than a destination: it suggests that progress on continual learning, long-context reasoning, modern optimizers, and self-modifying models will come from better exploiting the extra design axis of \emph{levels} rather than from ever-deeper static networks.

\newpage
\printbibliography
\newpage
\appendix

\section{Generalized Formulation for Nested Learning and Nested Systems}
In this section, we discuss the generalized version of nested systems and \nsam, we discussed in \autoref{sec:nested-learning}:

\begin{dfn}[(Generalized) Nested System]
    A (ordered) nested system is a system with $K$ (ordered) levels such that each level $1 \leq k \leq K$ consists of a set of optimization problems $\{ (\mathcal{L}^{(k)}_{i},  \mathcal{C}^{(k)}_i, \boldsymbol{\Theta}^{(k)}_{i}) \}_{i = 1}^{N_k}$, where $\mathcal{L}_{i}(\cdot; \cdot)$ is the optimization objective in the $i$-th problem, $\mathcal{C}_i$ is its context (the data that is optimized on), $\boldsymbol{\Theta}_{i}$ is the set of its parameters, and each optimization problem is optimized using gradient descent, or equivalently:
    \begin{align}
        {\boldsymbol{\theta}_{i}}^{(k)}_{t+1} = \arg\min_{\boldsymbol{\Phi}^{(k)}_{i}} \:\:  \mathcal{L}^{(k)}_i(\boldsymbol{\Phi}^{(k)}_{i}; x_{t+1}) + \frac{1}{2{\eta_{i}}^{(k)}_{t+1}}\:\|\boldsymbol{\Phi}^{(k)}_{i} - {\boldsymbol{\theta}_{i}}^{(k)}_{t} \|^{2}_2 \qquad \text{where}\:\: x_{t+1} \thicksim \mathcal{C}^{(k)}_i, \quad \text{and} \quad {\boldsymbol{\Phi}_{i}}^{(k)} \in  \boldsymbol{\Theta}^{(k)}_{i}.
    \end{align}
\end{dfn}

\begin{dfn}[(Generalized) Nested System of Associative Memories] \label{dfn:gnsam}
    A nested system of associative memory (\nsam) is a system with $K$ (ordered) levels such that each level $1 \leq k \leq K$ consists of a set of optimization problems $\{ (\mathcal{L}^{(k)}_{i},  \mathcal{C}^{(k)}_i, \boldsymbol{\Theta}^{(k)}_{i}) \}_{i = 1}^{N_k}$, where $\mathcal{C}_i = \{(\vk^{(i)}_j, \vv^{(i)}_j) \}_{j = 1}^{L_{i}}$ is a set of ground-truth mappings, $\mathcal{L}_{i}(\cdot; \cdot, \cdot)$ is the optimization objective that measures the quality of memory learned mappings in the $i$-th problem, $\boldsymbol{\Theta}_{i}$ is the set of memory parameters, and each optimization problem is optimized using gradient descent:
    \begin{align}
        {\boldsymbol{\theta}_{i}}^{(k)}_{t+1} = \arg\min_{\boldsymbol{\Phi}^{(k)}_{i}} \:\:  \mathcal{L}^{(k)}_i(\boldsymbol{\Phi}^{(k)}_{i}; \vk^{(i)}_{t+1}, \vv^{(i)}_{t+1}) + \frac{1}{2{\eta_{i}}^{(k)}_{t+1}}\:\|\boldsymbol{\Phi}^{(k)}_{i} - {\boldsymbol{\theta}_{i}}^{(k)}_{t} \|^{2}_2 \quad \:\: \text{where}\:\: (\vk^{(i)}_{t+1}, \vv^{(i)}_{t+1}) \thicksim \mathcal{C}^{(k)}_i, \:\:\:\: \text{and} \:\:\: \:{\boldsymbol{\Phi}_{i}}^{(k)} \in  \boldsymbol{\Theta}^{(k)}_{i}.
    \end{align}
\end{dfn}

\section{Adam, AdaGrad, and Other Similar Optimizers as Associative Memory Modules}\label{app:adam}
Revisiting the momentum term without using the chain rule in backpropagation,
\begin{align}\nonumber
    &{W_{\ell}}_{_{t+1}} = {W_{\ell}}_{_{t}} + {\boldsymbol{m}_{\ell}}_{_{t+1}}\\ \label{eq:momentum2}
    &{\boldsymbol{m}_{\ell}}_{_{t+1}} = \alpha_{\ell, t+1} {\boldsymbol{m}_{\ell}}_{_{t}} - \eta_{\ell, t+1} \nabla_{{W_{\ell}}_{_{t}}} \mathcal{L}\left({W_{\ell}}_{_{t}}; \boldsymbol{x}_{t+1}\right), 
\end{align}
one can interpret the momentum as a key or value-less associative memory~\citep[see Section 5]{behrouz2025Miras}, where the gradient terms $\nabla_{{W_{\ell}}_{_{t}}} \mathcal{L}\left({W_{\ell}}_{_{t}}; \boldsymbol{x}_{t+1}\right)$ are compressed into the momentum. We expect from a powerful momentum term to perfectly memorize all the past gradients in the training process so it can better model the current update to the weights. To this end, one can  start from  defining a simple objective as:
\begin{align}\label{eq:adam-objective2}
    \tilde{\mathcal{L}}_t = \sum_{i = 1}^{t} \| {\boldsymbol{m}_{\ell}}_{_{t}}  \odot {\boldsymbol{g}_{\ell}}_{_{i + 1}} - \vp_{\ell_{t}} \|_{2}^{2} \:\: +\:\: \lambda_{\ell} \|{\boldsymbol{m}_{\ell}}_{_{t}}\|^2_F ,
\end{align}
where ${\boldsymbol{g}_{\ell}}_{_{t + 1}} = - \nabla_{{W_{\ell}}_{_{t}}} \mathcal{L}\left({W_{\ell}}_{_{t}}; \boldsymbol{x}_{t+1}\right)$. This objective aims to find a momentum term that does not simply map the gradients to 1 (which also results in limited memory management in momentum), but it maps gradients to a global property of past data samples. The more expressive this global property is, the more accurately the momentum can incorporate the compressed information from past. Accordingly, the objective in \autoref{eq:adam-objective2} admits an optimal associative memory that maps gradients to $\vp_{\ell_{t+1}}$ as:
\begin{align}\nonumber
    &{\boldsymbol{m}^{(t)}_{\ell, i}}^{*} = \left[  \left( \boldsymbol{H}^{(t)}_{\ell, i} \: + \: \lambda_{\ell}\: \boldsymbol{I} \right)^{-1}\right] \odot \left( \boldsymbol{M}^{(t)}_{\ell, i}\right) = \left[  \left( \boldsymbol{H}^{(t)}_{\ell, i} \: + \: \lambda_{\ell}\: \boldsymbol{I} \right)^{-1}\right] \odot \tilde{\boldsymbol{M}}^{(t)}_{\ell, i+1} \odot \vp_{\ell_{t}}, \qquad \text{where} \\ \nonumber
    &\boldsymbol{M}^{(t)}_{\ell, i+1} = \boldsymbol{M}^{(t)}_{\ell, i} \: + \:  \beta_1 \: {\boldsymbol{g}_{\ell}}_{_{i + 1}} \odot \vp_{\ell_{t}} = \tilde{\boldsymbol{M}}^{(t)}_{\ell, i+1} \odot \vp_{\ell_{t}},\\ \nonumber
    &\tilde{\boldsymbol{M}}^{(t)}_{\ell, i+1} = \boldsymbol{M}^{(t)}_{\ell, i} \: + \:  \beta_1 \: {\boldsymbol{g}_{\ell}}_{_{i + 1}},\\ \label{eq:solution2}
    &\boldsymbol{H}^{(t)}_{\ell, i+1} = \boldsymbol{H}^{(t)}_{\ell, i} + \beta_2 \: {\boldsymbol{g}_{\ell}}_{_{i + 1}} \odot {\boldsymbol{g}_{\ell}}_{_{i + 1}} = \boldsymbol{H}^{(t)}_{\ell, i} + \beta_2 \: {\boldsymbol{g}_{\ell}}_{_{i + 1}}^{2}.
\end{align}
Given this solution, one can write the update step as:
\begin{align}\label{eq:general-adam}
    {W_{\ell}}_{_{i+1}} = {W_{\ell}}_{_{i}} - \eta_t \: {\boldsymbol{m}^{(t)}_{\ell, i}}^{*} = {W_{\ell}}_{_{i}} - \eta_i \: \left[  \left( \boldsymbol{H}^{(t)}_{\ell, i} \: + \: \lambda_{\ell}\: \boldsymbol{I} \right)^{-1}\right] \odot \tilde{\boldsymbol{M}}^{(t)}_{\ell, i+1} \odot \vp_{\ell_{t}}. 
\end{align}
We start with a simple case, where $\vp_{\ell_{t}}$ is the summation of the square of past gradients: i.e., $\vp_{\ell_{t}} = \sum_{i = 1}^{t} \vg_{\ell_{i+1}}^2$. With the choice of $\lambda \rightarrow 0$, \autoref{eq:general-adam} recovers simple gradient descent with momentum. That is, let $\lambda = 0$, we have:
\begin{align}
    {W_{\ell}}_{_{i+1}} = {W_{\ell}}_{_{i}} - \eta_t \: {\boldsymbol{m}^{(t)}_{\ell, i}}^{*} = {W_{\ell}}_{_{i}} - \eta_i \:  \left( \boldsymbol{H}^{(t)}_{\ell, i} \right)^{-1} \odot \tilde{\boldsymbol{M}}^{(t)}_{\ell, i+1} \odot \undermath{\boldsymbol{H}^{(t)}_{\ell, i} / \beta_2}{\vp_{\ell_{t}}} = {W_{\ell}}_{_{i}} - \eta_t \beta_2 \: \tilde{\boldsymbol{M}}^{(t)}_{\ell, i+1},
\end{align}
which based on the definition of $\tilde{\boldsymbol{M}}^{(t)}_{\ell, i+1}$, this update rule is equivalent to gradient descent with momentum. Next, we explore a more sophisticated design choice, where we use $\vp_{\ell_{t}}$ as the variance of data samples before token $t + 1$. In this case, formally,  $\vp_{\ell_{t}} = \sqrt{\sum_{i = 1}^{t} \vg_{\ell_{i+1}}^2}$ and so the update rule is as follows:
\begin{align}
    &{W_{\ell}}_{_{i+1}} = {W_{\ell}}_{_{i}} - \eta_t \: {\boldsymbol{m}^{(t)}_{\ell, i}}^{*} = {W_{\ell}}_{_{i}} - \eta_i \:  \left[  \left( \boldsymbol{H}^{(t)}_{\ell, i} \: + \: \lambda_{\ell}\: \boldsymbol{I} \right)^{-1}\right] \odot \tilde{\boldsymbol{M}}^{(t)}_{\ell, i+1} \odot \undermath{\boldsymbol{H}^{(t)^{\frac{1}{2}}}_{\ell, i} / \sqrt{\beta_2}}{\vp_{\ell_{t}}} \approx {W_{\ell}}_{_{i}} - \frac{\eta_t}{\sqrt{\beta_2}} \: \frac{\tilde{\boldsymbol{M}}^{(t)}_{\ell, i}}{\boldsymbol{H}^{(t)^{1/2}}_{\ell, i} \: + \varepsilon}
\end{align}
which is equivalent to the popular Adam optimizer~\citep{kingma2014adam}. Therefore, Adam is an optimal associative memory given the $L_2$ regression objective that is defined in \autoref{eq:adam-objective2}. In fact, the update component of Adam at each state aims to learn a mapping between the gradients and their variance (as a global property of past data samples). One interesting point about this formulation is about the frequency of elements. Looking at the first and second momentum in Adam optimizer, the frequency of update for both of these memories are the same as both are updated after each sample. Furthermore, the computation of each of them can be done in parallel and so are independent. Accordingly, it is one of the few examples that two components without any internal gradient flow, are placed in the same frequency and so level (see \autoref{sec:nop}).  

Going beyond element-wise update, we reformulate \autoref{eq:adam-objective2} with outer-product operation as:
\begin{align}\label{eq:adgrad-objective2}
    \tilde{\mathcal{L}}_t = \sum_{i = 1}^{t} \| {\boldsymbol{m}_{\ell}}_{_{t}}   {\boldsymbol{g}_{\ell}}_{_{i + 1}} - \vp_{\ell_{t}} \|_{2}^{2} \:\: +\:\: \lambda_{\ell} \|{\boldsymbol{m}_{\ell}}_{_{t}}\|^2_F ,
\end{align}
where ${\boldsymbol{g}_{\ell}}_{_{t + 1}} = - \nabla_{{W_{\ell}}_{_{t}}} \mathcal{L}\left({W_{\ell}}_{_{t}}; \boldsymbol{x}_{t+1}\right)$. Similar to \autoref{eq:solution2}, finding the optimal solution to the above objective is defined as:
\begin{align}\label{eq:solution-adagrad}
    &{\boldsymbol{m}^{(t)}_{\ell, i}}^{*} = \left[  \left( \boldsymbol{H}^{(t)}_{\ell, i} \: + \: \lambda_{\ell}\: \boldsymbol{I} \right)^{-1}\right]  \left( \boldsymbol{M}^{(t)}_{\ell, i}\right) = \left[  \left( \boldsymbol{H}^{(t)}_{\ell, i} \: + \: \lambda_{\ell}\: \boldsymbol{I} \right)^{-1}\right]  \tilde{\boldsymbol{M}}^{(t)}_{\ell, i+1} \vp_{\ell_{t}}, \qquad \text{where} \\
    &\boldsymbol{M}^{(t)}_{\ell, i+1} = \boldsymbol{M}^{(t)}_{\ell, i} \: + \:  \beta_1 \: \vp_{\ell_{t}} \: {\boldsymbol{g}_{\ell}}_{_{i + 1}}^{\top}   = \vp_{\ell_{t}}  \: \tilde{\boldsymbol{M}}^{(t)}_{\ell, i+1}  ,\\
    &\tilde{\boldsymbol{M}}^{(t)}_{\ell, i+1} = \boldsymbol{M}^{(t)}_{\ell, i} \: + \:  \beta_1 \: {\boldsymbol{g}_{\ell}}_{_{i + 1}}^{\top},\\
    &\boldsymbol{H}^{(t)}_{\ell, i+1} = \boldsymbol{H}^{(t)}_{\ell, i} + \beta_2 \: {\boldsymbol{g}_{\ell}}_{_{i + 1}} {\boldsymbol{g}_{\ell}}_{_{i + 1}}^{\top}.
\end{align}
With a similar choice as Adam, i.e., letting $\vp_{\ell_{t}}$ be the variance of the gradients so far, $\vp_{\ell_{t}} = \sqrt{\sum_{i = 1}^{t} \vg_{\ell_{i+1}} \vg_{\ell_{i+1}}^{\top}}$, then the above solution can be simplified as:
\begin{align}
    &{W_{\ell}}_{_{i+1}} = {W_{\ell}}_{_{i}} - \eta_t \: {\boldsymbol{m}^{(t)}_{\ell, i}}^{*} = {W_{\ell}}_{_{i}} - \eta_i \:  \left[  \left( \boldsymbol{H}^{(t)}_{\ell, i} \: + \: \lambda_{\ell}\: \boldsymbol{I} \right)^{-1}\right] \: \undermath{\boldsymbol{H}^{(t)^{\frac{1}{2}}}_{\ell, i} / \sqrt{\beta_2}}{\vp_{\ell_{t}}} \: \tilde{\boldsymbol{M}}^{(t)}_{\ell, i+1}   \approx {W_{\ell}}_{_{i}} - \frac{\eta_t}{\sqrt{\beta_2}} \: \boldsymbol{H}^{(t)^{-1/2}}_{\ell, i} \:\tilde{\boldsymbol{M}}^{(t)}_{\ell, i}
\end{align}
The above formulation is the AdaGrad with momentum~\citep{defossez2022a} and so generalizes the AdaGrad~\citep{duchi2011adaptive}, i.e., when $\beta_1 = 1$. Similarly, based on the connection of Adam optimizer~\citep{kingma2014adam} with other algorithms such as RMSProp~\citep{hinton2012neural}, SignSGD and its momentum-based variants~\citep{bernstein2018signsgd}, NAdam~\citep{dozat2016incorporating}, AMSGrad~\citep{reddi2016stochastic}, RAdam~\citep{Liu2020On}, and Lion~\citep{chen2023symbolic}, as well as considering AdaGrad's connection with optimizers such as Shampoo~\citep{gupta2018shampoo} and Soap~\citep{vyas2025soap}–i.e., as the approximation of the preconditioning term–we can conclude that all these optimizers can be re-formulated as associative memory that aims to compress the gradients.

\section{Delta Gradient Descent with Normalization}\label{app:DGD}
In \autoref{sec:L2-backprop} we discussed that gradient descent can be see as an associative memory, and be reformulated as:
\begin{align}
    W_{t+1} = \arg\min_{W} \:\: \inner{W\boldsymbol{x}_t}{\nabla_{y_t} \mathcal{L}(W_t;\boldsymbol{x}_t)} + \frac{1}{2\eta_t} \: \|W - W_{t}\|^{2}_2,
\end{align}
where each step aims at learning the negative of the gradient direction. Due to the fact that this learning rule only use update terms that depends on the current gradient, we defined $\mathbf{u}_t = -\nabla_{y_t} \mathcal{L}(W_t;\boldsymbol{x}_t)$, and extended the above process to Delta Gradient Descent with more expressive objective of $\mathcal{L}_2$ regression loss:
\begin{align}
    W_{t+1} = \arg\min_{W} \:\: \frac{1}{2}\|W\boldsymbol{x}_t - \mathbf{u}_t \|^{2}_2 +  \frac{1}{2\eta_t}\|W - W_{t}\|^{2}_2.
\end{align}
We assume that $\vx_t$ is normalized (e.g.  in normalized memory systems or in neural networks with normalization layers, $\|\vx_t\|_2 = \lambda$). Taking gradient to optimize the above objective, 
\begin{align}
    2(W_{t+1}\vx_t - \nabla_{y_t} \mathcal{L}(W_t,\vx_t))\vx_t^{\top} + 2 \eta_t (W_{t+1}-W_t) = 0,
\end{align}
which results in:
\begin{align}
    &W_{t+1}(\vx_t \vx_t^{\top} + \eta_t I) = \nabla_{y_t} \mathcal{L}(W_t,\vx_t) \vx_t^{\top} + \eta_t W_t, \\
    \Rightarrow\:\:&W_{t+1} = (\nabla_{y_t} \mathcal{L}(W_t,x_t) x_t^{\top} + \eta_t W_t)(\vx_t \vx_t^{\top} + \eta_t I)^{-1}.
\end{align}
To compute $( \vx_t \vx_t^{\top} + \eta_t I)^{-1}$ term with  Sherman-Morrison lemma, we have:
\begin{align}
    (x_tx_t^{\top} + \eta_t I)^{-1} = \frac{1}{\eta_t} (I - \frac{1}{\lambda^2+\eta_t} \vx_t \vx_t^{\top}),
\end{align}
and so:
\begin{align}
    &W_{t+1} = (\nabla_{y_t} \mathcal{L}(W_t,\vx_t) \vx_t^{\top} + \eta_t W_t)\frac{1}{\eta_t} (I - \frac{1}{\lambda^2+\eta_t}  \vx_t \vx_t^{\top}) \\
    \Rightarrow \:\:& W_{t+1} = W_t \left(I - \frac{1}{\lambda^2+\eta_t}  \vx_t \vx_t^{\top} \right) +  \frac{1}{\eta_t} \nabla_{y_t} \mathcal{L}(W_t,\vx_t) \vx_t^{\top} - \undermath{\frac{\lambda}{\lambda^2\eta_t+\eta^2_t} \nabla_{y_t} \mathcal{L}(W_t,\vx_t) \vx_t^{\top}}{\frac{1}{\lambda^2\eta_t+\eta^2_t} \nabla_{y_t} \mathcal{L}(W_t,\vx_t) \vx_t^{\top}\vx_t \vx_t^{\top}} \\
    \Rightarrow \:\:&W_{t+1} = W_t \left(I - \frac{1}{\lambda^2+\eta_t}  \vx_t \vx_t^{\top} \right)  - \left( \frac{\lambda}{\lambda^2\eta_t+\eta^2_t}  - \frac{1}{\eta_t} \right) \nabla_{y_t} \mathcal{L}(W_t,\vx_t) \vx_t^{\top} \\
    \Rightarrow \:\:&W_{t+1} = W_t \left(I - \alpha_t \vx_t \vx_t^{\top} \right)  - \beta \:\: \nabla_{y_t} \mathcal{L}(W_t,\vx_t) \vx_t^{\top}
\end{align}

\end{document}